\documentclass[examplefnt]{nowfnt}

\usepackage[utf8]{inputenc}

\definecolor{blue}{RGB}{0, 76, 153}
\definecolor{darkred}{RGB}{153, 0, 0}
\hypersetup{
  colorlinks,
  linkcolor={red!50!black},     
  citecolor=blue,    
  urlcolor=black     
}
\usepackage{amsmath}
\usepackage{multirow}
\usepackage{amssymb}
\usepackage{mathtools}
\usepackage{nicefrac}
\usepackage{enumerate}
\usepackage{graphicx}
\usepackage{setspace}
\usepackage{booktabs}
\usepackage{subcaption}
\usepackage{wrapfig}
\usepackage{cancel}
\let\olddiv\div
\usepackage{physics}
\usepackage{lscape}
\usepackage{longtable}

\allowdisplaybreaks

\newcommand\ci{\perp\!\!\!\perp}
\DeclareMathOperator*{\argmin}{arg\,min}

\usepackage{tikz}
\usetikzlibrary{bayesnet, calc, arrows.meta, bending}
\usetikzlibrary{decorations.pathreplacing,angles,quotes,shapes.multipart, calligraphy}
\usetikzlibrary{external}

\makeatletter
\newcommand{\customtag}[1]{%
  \refstepcounter{equation} 
  \tag*{(#1) (\theequation)}
}

\newcommand{\customlabel}[1]{%
  \edef\@currentlabel{\theequation}%
  \ltx@label{#1}
}
\makeatother

\title{Demystifying Variational\\Diffusion Models}

\maintitleauthorlist{
Fabio De Sousa Ribeiro \\
Imperial College London \\
f.de-sousa-ribeiro@imperial.ac.uk
\and
Ben Glocker \\
Imperial College London \\
b.glocker@imperial.ac.uk
}

\usepackage{mwe}

\author[1]{De Sousa Ribeiro, Fabio}
\author[2]{Glocker,Ben}

\affil[1]{Imperial College London; f.de-sousa-ribeiro@imperial.ac.uk}
\affil[2]{Imperial College London; b.glocker@imperial.ac.uk}

\begin{document}

\makeabstracttitle

\begin{abstract}
Despite the growing interest in diffusion models, gaining a deep understanding of the model class remains an elusive endeavour, particularly for the uninitiated in non-equilibrium statistical physics. Thanks to the rapid rate of progress in the field, most existing work on diffusion models focuses on either applications or theoretical contributions. Unfortunately, the theoretical material is often inaccessible to practitioners and new researchers, leading to a risk of superficial understanding in ongoing research. Given that diffusion models are now an indispensable tool, a clear and consolidating perspective on the model class is needed to properly contextualize recent advances in generative modelling and lower the barrier to entry for new researchers. To that end, we revisit predecessors to diffusion models like hierarchical latent variable models and synthesize a holistic perspective using only directed graphical modelling and variational inference principles. The resulting narrative is easier to follow as it imposes fewer prerequisites on the average reader relative to the view from non-equilibrium thermodynamics or stochastic differential equations. 
\end{abstract}

\chapter{Introduction}
\label{sec: introduction}
A generative model is a simulation of a data-generating process. Understanding the true generative process of data is valuable as it naturally reveals the causal relationships in the world. These causal relationships are advantageous as they tend to generalize more effectively to new situations than mere correlations, which may be spurious and unreliable. Generative modelling typically consists of using data from observations of $\mathbf{x}$ to estimate the marginal distribution $p(\mathbf{x})$. Knowing $p(\mathbf{x})$ facilitates many useful tasks, such as: (i) sample generation, (ii) density estimation, (iii) compression, (iv) data imputation, (v) model selection, etc. As $p(\mathbf{x})$ is typically unknown and/or intractable, we often have to approximate it with a model $p_{\boldsymbol{\theta}}(\mathbf{x}) \approx p(\mathbf{x})$, by optimizing some parameters $\boldsymbol{\theta}$. Although various generative modelling strategies exist, diffusion models~\citep{sohl2015deep,ho2020denoising} have emerged as the latest dominant paradigm. With that said, gaining a deep understanding of the model class remains an elusive endeavour, particularly for the uninitiated in non-equilibrium statistical physics.

Thanks to the rapid rate of progress in the field, existing work on diffusion models focuses on either applications or theoretical contributions. 
However, research material on diffusion is often inaccessible to practitioners and new researchers. Given that diffusion models are now an indispensable tool, we argue that a clear, consolidating perspective on the model class is needed to properly contextualize recent advances in generative modelling and lower the barrier to entry.
To that end, we revisit predecessors to diffusion models like hierarchical latent variable models (HLVMs)~\citep{salimans2015markov,valpola2015neural,sonderby2016ladder}, and synthesize a holistic perspective using \textit{only} directed graphical modelling and variational inference principles. The resulting narrative is easier to follow as it imposes fewer prerequisites on the reader relative to the view from non-equilibrium thermodynamics~\citep{sohl2015deep} or stochastic differential equations (SDEs)~\citep{song2021scorebased,song2021maximum}. Other variational perspectives on diffusion have been studied~\citep{huang2021variational,kingma2021variational,vahdat2021score}, but their expositions are optimized for technical and empirical contributions to the model class rather than accessibility. A notable exception is the technical review by~\citet{luo2022understanding}; however, our account is far more comprehensive, covers a lot more recent material, and is more mathematically consistent with the seminal works in the field~\citep{kingma2021variational,kingma2023understanding}.

We begin our exposition by revisiting deep latent variable models~\citep{kingma2013auto,rezende2014stochastic} and their hierarchical counterparts (\S\ref{sec: background}). We then highlight the difficulties with bottom-up inference procedures for even modestly deep hierarchies and present a compelling argument in favour of the \textit{top-down} hierarchical model using a concept called \textit{generative feedback} (\S\ref{subsec: Generative Feedback}). This contrasts with prior work~\citep{burda2015importance,luo2022understanding}, which offers an incomplete efficiency-based view. We then show that the top-down hierarchy is ubiquitous in both classical HLVMs~\citep{valpola2015neural,sonderby2016ladder,kingma2016improved} and diffusion models. We explain how both model classes share optimization objectives and offer an intuitive understanding of diffusion models as a specific instantiation of HLVMs with top-down inference. In \S\ref{subsec: main hole}, we reproduce the \textit{hole problem} in LVMs, explain how diffusion models overcome it by construction, and stress its importance for sample quality. In \S\ref{sec: variational diffusion models}, we provide a comprehensive account of modern diffusion models from the top-down hierarchy perspective, and in \S\ref{sec: discussion_outlook}, we conclude with a forward-looking discussion.
\chapter{Latent Variable Models}
\label{sec: background}
A latent variable model is a statistical model that connects observable variables to a set of unobserved (i.e., latent) variables. The core assumption is that the data are generated by a random process involving these unobserved variables, which capture the underlying structure of the data. This approach often simplifies analysis by reducing data dimensionality, uncovering hidden patterns, and enhancing interpretability. Latent variable models are widely used to identify and model latent factors that influence observed phenomena, thereby improving our ability to understand, predict, and draw inferences from complex data. 

A basic example of a latent variable model is the Gaussian Mixture Model (GMM), which posits that data are generated from a mixture of multiple Gaussian distributions, each representing a different latent component or cluster. The assumed data-generating process is simple: (i) a cluster (i.e., latent mixture component) is selected at random from a finite set; (ii) a data point is drawn from the corresponding Gaussian distribution. GMMs are useful density estimators as the Gaussianity assumption admits tractable closed-form solutions, enabling efficient computation of the marginal likelihood $p(\mathbf{x})$ and parameter estimation using methods like the Expectation-Maximization (EM) algorithm.

The advent of deep learning has given rise to \textit{deep} latent variable models, which leverage neural networks to capture more complex, non-linear relationships between observed and latent variables. These models, including Variational Autoencoders (VAEs) and diffusion models, have gained prominence for their ability to model high-dimensional data, generate realistic synthetic samples, and discover intricate underlying structures in ways that traditional models previously struggled to. In the following sections, we will cover the necessary fundamentals of VAEs and hierarchical latent variable models, which will form the foundation of our exposition on variational diffusion models.

\section{Variational Autoencoder}
Variational autoencoders~\citep{kingma2013auto} assume that data $\mathbf{x} \in \mathcal{X}^D$ are generated by some random process involving an unobserved random variable $\mathbf{z} \in \mathcal{Z}^K$. The generative process is straightforward: 
\begin{enumerate}[{(i)}]
    \item sample a latent variable from a prior distribution $\mathbf{z} \sim p(\mathbf{z})$;
    \item sample an observation from a conditional distribution $\mathbf{x}\sim p(\mathbf{x} \mid \mathbf{z})$.
\end{enumerate}
If we choose $\mathbf{z}$ to be a discrete random variable and $p(\mathbf{x} \mid \mathbf{z})$ to be a Gaussian distribution, then $p(\mathbf{x})$ is a Gaussian mixture. If we instead choose $\mathbf{z}$ to be a continuous random variable, then the marginal distribution of $\mathbf{x}$ represents an \textit{infinite} mixture of Gaussians, and is given by marginalizing out the latent variable: $p(\mathbf{x}) = \int p(\mathbf{x}, \mathbf{z})\mathop{\mathrm{d}\mathbf{z}}$. 

For complicated non-linear likelihood functions, where $p(\mathbf{x} \mid \mathbf{z})$ is parameterized by a deep neural network for example, integrating out the latent variable $\mathbf{z}$ to compute $p(\mathbf{x})$ has no analytic solution, so we must rely on approximations. A straightforward Monte Carlo approximation of $p(\mathbf{x})$ is certainly possible: 
\begin{align}
&p(\mathbf{x}) = \mathbb{E}_{\mathbf{z} \sim p(\mathbf{z})} \left[p(\mathbf{x} \mid \mathbf{z})\right] \approx
\frac{1}{N}\sum_{i=1}^N p(\mathbf{x} \mid \mathbf{z}_i), &\mathbf{z}_1,\dots,\mathbf{z}_N \mathop{\sim}\limits^{\mathrm{iid}} p(\mathbf{z}), &
\end{align}
but is subject to the \textit{curse of dimensionality}, since the number of samples needed to properly cover the latent space grows exponentially with the dimensionality of the latent variable $\mathbf{z}$.

\begin{figure}[!t]
    \centering
    \hfill
    \begin{subfigure}{.32\columnwidth}
        \centering
        \begin{tikzpicture}[thick,scale=1, every node/.style={scale=1}]
            \node[latent] (z1) {$\mathbf{z}$};
            \node[obs, below=15pt of z1] (x) {$\mathbf{x}$};
            \edge[-{Latex[scale=1.0]}]{z1}{x}
            \node[latent, draw=none, right=-2pt of z1, yshift=-16pt] (eq1) {$p(\mathbf{x} \mid \mathbf{z})$};
        \end{tikzpicture}
        \caption{Generative Model}
    \end{subfigure}
    \hfill
    \begin{subfigure}{.32\columnwidth}
        \centering
        \begin{tikzpicture}[thick,scale=1, every node/.style={scale=1}]
            \node[latent] (z1) {$\mathbf{z}$};
            \node[obs, below=15pt of z1] (x) {$\mathbf{x}$};
            \edge[-{Latex[scale=1.0]}]{x}{z1}
            \node[latent, draw=none, right=-2pt of z1, yshift=-18pt] (eq1) {$q(\mathbf{z} \mid \mathbf{x})$};
        \end{tikzpicture}
        \caption{Inference Model}
    \end{subfigure}
    \hfill
    \begin{subfigure}{.32\columnwidth}
        \centering
        \begin{tikzpicture}[thick,scale=1, every node/.style={scale=1}]
            \node[latent] (z1) {$\mathbf{z}$};
            \node[obs, below=15pt of z1] (x) {$\mathbf{x}$};
            \edge[-{Latex[scale=1.0]}]{z1}{x}
            \draw [dashed,-{Latex[scale=1.0]}] (x) to [out=135,in=235] (z1);
            \node[latent, draw=none, right=-2pt of z1, yshift=-16pt] (eq1) {$p(\mathbf{x} \mid \mathbf{z})$};
            \node[latent, draw=none, left=5pt of z1, yshift=-18pt] (eq1) {$q(\mathbf{z} \mid \mathbf{x})$};
        \end{tikzpicture}
        \caption{Combined Model}
    \end{subfigure}
    \hfill
    \caption{\textbf{Probabilistic graphical model of a latent variable model} (e.g. variational autoencoder). Directed arrows represent the assumed flow of conditional dependencies or causal influence between variables.}
    \label{fig: vanilla vae}
\end{figure}
Alternatively, we can turn to \textit{variational} methods, which pose probabilistic inference as an optimization problem~\citep{jordan1999introduction}. The first thing to note is that the intractability of $p(\mathbf{x})$ is related to the intractability of the true posterior over the latent variable $p(\mathbf{z} \mid \mathbf{x})$ through a basic identity:
\begin{align}
    &&p(\mathbf{x}) = \frac{p(\mathbf{x} \mid \mathbf{z})p(\mathbf{z})}{p(\mathbf{z} \mid \mathbf{x})},
    &&
    \mathrm{where}
    &&p(\mathbf{z} 
    \mid \mathbf{x}) = \frac{p(\mathbf{x} \mid \mathbf{z})p(\mathbf{z})}{\int p(\mathbf{x} \mid \mathbf{z})p(\mathbf{z})\mathop{\mathrm{d}\mathbf{z}}}.&&
\end{align}
Using a complicated neural network-based likelihood renders the integral on the right-hand side (RHS) intractable. To estimate $p(\mathbf{x})$ we can approximate the true posterior $p(\mathbf{z} \mid \mathbf{x})$ via a parametric inference model $q(\mathbf{z} \mid \mathbf{x})$ of our choice, such that $q(\mathbf{z} \mid \mathbf{x}) \approx p(\mathbf{z} \mid \mathbf{x})$. Learning a single function with shared variational parameters $\boldsymbol{\phi}$ to map each datapoint $\mathbf{x}$ to a posterior distribution $q(\mathbf{z} \mid \mathbf{x})$ is known as \textit{amortized inference}.
Alternatively, optimizing distinct variational parameters for each individual datapoint $\mathbf{x}$ is equally valid, more expressive, but much less efficient, as it typically involves a per-datapoint optimization loop known as \textit{coordinate-ascent} variational inference.

We may optionally write $q_{\boldsymbol{\phi}}(\mathbf{z} \mid \mathbf{x})$ and $p_{\boldsymbol{\theta}}(\mathbf{x} \mid \mathbf{z})$ to explicitly state that these are parametric distributions realized by an encoder-decoder setup with variational parameters $\boldsymbol{\phi}$ and model parameters $\boldsymbol{\theta}$. The typical VAE setup specifies a prior $p(\mathbf{z})$ with no learnable parameters, and it is often chosen to be standard Gaussian: $p(\mathbf{z}) = \mathcal{N}(\mathbf{z}; 0, \mathbf{I})$. It is important to note that unlike the latent variable(s) $\mathbf{z}$ which are \textit{local}, the parameters $\{\boldsymbol{\phi}, \boldsymbol{\theta}\}$ are \textit{global} since they are shared for all datapoints. 

To improve our approximation, we would like to minimize the Kullback-Leibler (KL) divergence $\argmin_{q(\mathbf{z} \mid \mathbf{x})} D_{\mathrm{KL}} \left(q(\mathbf{z} \mid \mathbf{x}) \parallel p(\mathbf{z} \mid \mathbf{x}) \right)$ (see \cite{rezendeblog} to learn why), but it is not possible do so directly as we do not have access to the true posterior $p(\mathbf{z} \mid \mathbf{x})$ for evaluation.

VAEs maximise the Variational Lower Bound (VLB) of $\log p(\mathbf{x})$:
\begin{align}
    D_{\mathrm{KL}} (q(\mathbf{z} &\mid \mathbf{x}) \parallel p(\mathbf{z} \mid \mathbf{x})) =  \int q(\mathbf{z} \mid \mathbf{x})  \log \frac{q(\mathbf{z} \mid \mathbf{x})}{p(\mathbf{z} \mid \mathbf{x})} \mathop{\mathrm{d}\mathbf{z}} 
    \\[5pt] &= \mathbb{E}_{q(\mathbf{z} \mid \mathbf{x})} \left[  \log q(\mathbf{z} \mid \mathbf{x}) - \log \frac{p(\mathbf{x}, \mathbf{z})}{p(\mathbf{x})}\right]
    \\[5pt] &= \mathbb{E}_{q(\mathbf{z} \mid \mathbf{x})} \left[  \log q(\mathbf{z} \mid \mathbf{x}) - \log p(\mathbf{x}, \mathbf{z}) \right] + \log p(\mathbf{x})
    \\[5pt] &= -\mathrm{VLB}(\mathbf{x}) + \log p(\mathbf{x}),
\end{align}
now adding $\mathrm{VLB}(\mathbf{x})$ to both sides reveals:
\begin{align}
    D_{\mathrm{KL}} \left(q(\mathbf{z} \mid \mathbf{x}) \parallel p(\mathbf{z} \mid \mathbf{x}) \right) + \mathrm{VLB}(\mathbf{x}) &= \log p(\mathbf{x}) 
    \\[5pt] \implies \mathrm{VLB}(\mathbf{x}) &\leq \log p(\mathbf{x}),
\end{align}
as $D_{\mathrm{KL}} \left(q(\mathbf{z} \mid \mathbf{x}) \parallel p(\mathbf{z} \mid \mathbf{x}) \right) \geq 0$ by Gibbs' inequality. Hence, maximizing the VLB implicitly minimizes the KL divergence of $q(\mathbf{z} \mid \mathbf{x})$ from the true posterior $p(\mathbf{z} \mid \mathbf{x})$ as desired. The VLB is also known as the Evidence Lower BOund (ELBO) since $p(\mathbf{x})$ is called the \textit{evidence}. The form of the VLB typically optimized by VAEs is:
\begin{align}
    \mathrm{VLB}(\mathbf{x}) &= \mathbb{E}_{q(\mathbf{z} \mid \mathbf{x})} \left[\log \frac{p(\mathbf{x}, \mathbf{z})}{q(\mathbf{z} \mid \mathbf{x})}\right]
    \\[5pt] &= \mathbb{E}_{q(\mathbf{z} \mid \mathbf{x})} \left[\log p(\mathbf{x} \mid \mathbf{z})\right]
    + \mathbb{E}_{q(\mathbf{z} \mid \mathbf{x})} \left[\log \frac{p(\mathbf{z})}{q(\mathbf{z} \mid \mathbf{x})} \right]
    \\[5pt] &= \mathbb{E}_{q(\mathbf{z} \mid \mathbf{x})} \left[\log p(\mathbf{x} \mid \mathbf{z})\right] - D_{\mathrm{KL}} \left(q(\mathbf{z} \mid \mathbf{x}) \parallel p(\mathbf{z}) \right),
\end{align}
which amounts to maximizing the expected likelihood $p(\mathbf{x} \mid \mathbf{z})$ of observing $\mathbf{x}$ under our approximate posterior $q(\mathbf{z} \mid \mathbf{x})$, regularized by the KL divergence of $q(\mathbf{z} \mid \mathbf{x})$ from the prior $p(\mathbf{z})$. 

If we let $\mathcal{D}$ be a dataset of i.i.d. data, then $\mathrm{VLB}(\mathcal{D}) = \sum_{\mathbf{x} \in \mathcal{D}}\mathrm{VLB}(\mathbf{x})$. Finally, we can use stochastic variational inference~\citep{hoffman2013stochastic} and the \textit{reparameterization trick}~\citep{kingma2013auto,rezende2014stochastic} to \textit{jointly} optimize the VLB w.r.t. the model parameters $\boldsymbol{\theta}$, and variational parameters $\boldsymbol{\phi}$. For more details on this procedure, the reader may refer to~\cite{kingma2019introduction} and \cite{blei2017variational}.

\begin{figure}[!t]
    \centering
    \hfill
    \begin{subfigure}{.32\columnwidth}
        \centering
        \begin{tikzpicture}[thick,scale=1, every node/.style={scale=1}]
            \node[latent] (zt) {$\mathbf{z}_T$};
            \draw node[draw=none, scale=0.75, below=6pt of zt] (z3) {\hspace{0.5pt}\rotatebox{90}{$\mathbf{\cdots}$}};
            \node[latent, below=13pt of z3] (z2) {$\mathbf{z}_2$};
            \node[latent, below=15pt of z2] (z1) {$\mathbf{z}_1$};
            \node[obs, below=15pt of z1] (x) {$\mathbf{x}$};
            \edge[-]{zt}{z3}
            \edge[-{Latex[scale=1.0]}]{z3}{z2}
            \edge[-{Latex[scale=1.0]}]{z2}{z1}
            \edge[-{Latex[scale=1.0]}]{z1}{x}
            \node[latent, draw=none, right=-2pt of z2, yshift=-16pt] (eq2) {$p(\mathbf{z}_1 \mid \mathbf{z}_2)$};
            \node[latent, draw=none, right=-2pt of z1, yshift=-16pt] (eq1) {$p(\mathbf{x} \mid \mathbf{z}_1)$};
            \node[latent, draw=none, right=-2pt of z3, xshift=6pt, yshift=-0pt] (eq3) {$p(\mathbf{z}_{t-1} \mid \mathbf{z}_t)$};
        \end{tikzpicture}
        \caption{Generative Model}
    \end{subfigure}
    \hfill
    \begin{subfigure}{.32\columnwidth}
        \centering
        \begin{tikzpicture}[thick,scale=1, every node/.style={scale=1}]
            \node[latent] (zt) {$\mathbf{z}_T$};
            \draw node[draw=none, scale=0.75, below=13pt of zt] (z3) {\hspace{0.5pt}\rotatebox{90}{$\mathbf{\cdots}$}};
            \node[latent, below=6pt of z3] (z2) {$\mathbf{z}_2$};
            \node[latent, below=15pt of z2] (z1) {$\mathbf{z}_1$};
            \node[obs, below=15pt of z1] (x) {$\mathbf{x}$};
            \edge[-{Latex[scale=1.0]}]{z3}{zt}
            \edge[-]{z2}{z3}
            \edge[-{Latex[scale=1.0]}]{z1}{z2}
            \edge[-{Latex[scale=1.0]}]{x}{z1}
            \node[latent, draw=none, right=-2pt of z2, yshift=-16pt] (eq2) {$q(\mathbf{z}_2 \mid \mathbf{z}_1)$};
            \node[latent, draw=none, right=-2pt of z1, yshift=-16pt] (eq1) {$q(\mathbf{z}_1 \mid \mathbf{x})$};
            \node[latent, draw=none, right=-2pt of z3, xshift=6pt, yshift=-0pt] (eq3) {$q(\mathbf{z}_{t} \mid \mathbf{z}_{t-1})$};
        \end{tikzpicture}
        \caption{Bottom-up Inference}
    \end{subfigure}
    \hfill
    \begin{subfigure}{.32\columnwidth}
        \centering
        \begin{tikzpicture}[thick,scale=1, every node/.style={scale=1}]
            \node[obs] (x) {$\mathbf{x}$};
            \node[latent, below=15pt of x] (zt) {$\mathbf{z}_T$};
            \draw node[draw=none, scale=0.75, below=6pt of zt] (z3) {\hspace{0.5pt}\rotatebox{90}{$\mathbf{\cdots}$}};
            \node[latent, below=13pt of z3] (z2) {$\mathbf{z}_2$};
            \node[latent, below=15pt of z2] (z1) {$\mathbf{z}_1$};
            \edge[-{Latex[scale=1.0]}]{x}{zt}
            \edge[-]{zt}{z3}
            \edge[-{Latex[scale=1.0]}]{z3}{z2}
            \edge[-{Latex[scale=1.0]}]{z2}{z1}
            \draw [-{Latex[scale=1.0]}] (x) to [out=225,in=135] (z2);
            \draw [-{Latex[scale=1.0]}] (x) to [out=225,in=135] (z1);
            \node[latent, draw=none, right=-2pt of x, yshift=-16pt] (eq1) {$q(\mathbf{z}_T \mid \mathbf{x})$};
            \node[latent, draw=none, right=-2pt of z2, yshift=-16pt] (eq2) {$q(\mathbf{z}_1 \mid \mathbf{z}_2, \mathbf{x})$};
            \node[latent, draw=none, right=-2pt of z3, xshift=6pt, yshift=-0pt] (eq3) {$q(\mathbf{z}_{t-1} \mid \mathbf{z}_{t}, \mathbf{x})$};
        \end{tikzpicture}
        \caption{Top-down Inference}
    \end{subfigure}
    \hfill
    \caption{\textbf{Hierarchical latent variable graphical models}. (a) The generative model $p(\mathbf{x}, \mathbf{z}_{1:T})$ of a hierarchical VAE with $T$ latent variables is a Markov chain. (b) The standard \textit{bottom-up} inference model $q(\mathbf{z}_{1:T} \mid \mathbf{x})$ of a hierarchical VAE is a Markov chain in the reverse direction. (c) The \textit{top-down} inference model follows the same topological ordering of the latent variables as the generative model. Notably, this top-down structure is also used to specify diffusion models. However, in diffusion models the posterior $q(\mathbf{z}_{1:T} \mid \mathbf{x})$ is tractable due to Gaussian conjugacy, which enables us to specify the \textit{generative} model transitions as $p(\mathbf{z}_{t-1} \mid \mathbf{z}_{t}) = q(\mathbf{z}_{t-1} \mid \mathbf{z}_{t}, \mathbf{x} = \hat{\mathbf{x}}_{\boldsymbol{\theta}}(\mathbf{z}_t; t))$,  where the data $\mathbf{x}$ is replaced by an image denoising model $\hat{\mathbf{x}}_{\boldsymbol{\theta}}(\mathbf{z}_t; t)$.
    }
    \label{fig: hvae}
\end{figure}
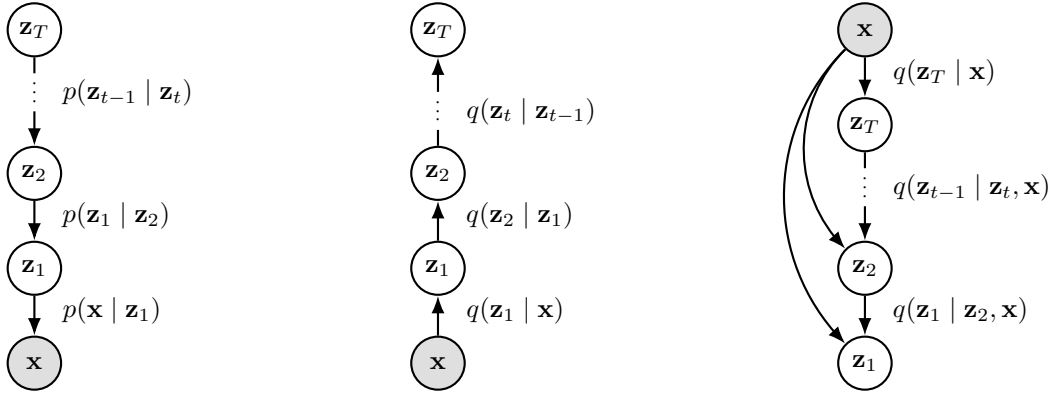

\section{Hierarchical Latent Variable Models}
\label{subsec: Hierarchical VAE}
A hierarchical VAE is a deep latent variable model comprised of a hierarchy of $T \geq 2$ latent variables $\mathbf{z}_1,\mathbf{z}_2, \dots, \mathbf{z}_T$. Introducing additional (auxiliary) latent variables significantly improves the flexibility and expressivity of both inference and generative models~\citep{salimans2015markov,ranganath2016hierarchical,maaloe2016auxiliary}.

The joint distribution $p(\mathbf{x}, \mathbf{z}_{1:T})$ specifying a generative model of $\mathbf{x}$ is a variational Markov chain:
\begin{align}
    \mathbf{z}_T \to \mathbf{z}_{T-1} \to \mathbf{z}_{T-2} \to \cdots \to \mathbf{z}_1 \to \mathbf{x}, 
\end{align}
defined as:
\begin{align}
    p(\mathbf{x}, \mathbf{z}_{1:T}) &= p(\mathbf{z}_T)p(\mathbf{z}_{T-1} \mid \mathbf{z}_T) \cdots p(\mathbf{z}_1 \mid \mathbf{z}_2) p(\mathbf{x} \mid \mathbf{z}_1)
    \\[5pt] & 
    = p(\mathbf{z}_T) \left[\prod_{t=2}^{T}p(\mathbf{z}_{t-1} \mid \mathbf{z}_t)\right]p(\mathbf{x} \mid \mathbf{z}_1). \label{eq: hvae_gen}
\end{align}
The approximate posterior $q(\mathbf{z}_{1:T} \mid \mathbf{x})$ is also a Markov chain but in the reverse (bottom-up) direction: 
\begin{align}
    \mathbf{z}_T \leftarrow \mathbf{z}_{T-1} \leftarrow \mathbf{z}_{T-2} \leftarrow \cdots \leftarrow \mathbf{z}_1 \leftarrow \mathbf{x},
\end{align}
which is defined as:
\begin{align}
    q(\mathbf{z}_{1:T} \mid \mathbf{x}) &= q(\mathbf{z}_1 \mid \mathbf{x})q(\mathbf{z}_2 \mid \mathbf{z}_1)q(\mathbf{z}_3 \mid \mathbf{z}_2) \cdots q(\mathbf{z}_T \mid \mathbf{z}_{T-1})
    \\[5pt] & 
    = q(\mathbf{z}_1 \mid \mathbf{x}) \prod_{t=2}^{T}q(\mathbf{z}_{t} \mid \mathbf{z}_{t-1}). \label{eq: hvae_inf}
\end{align}
The marginal likelihood $p(\mathbf{x})$ is obtained by marginalizing out the latent variables:
\begin{align}
    p(\mathbf{x}) 
    = \int p(\mathbf{x} \mid  \mathbf{z}_{1})p(\mathbf{z}_1) \mathop{\mathrm{d}\mathbf{z}_{1}},
\end{align}
where
\begin{align}
    &&p(\mathbf{z}_t) 
    = \int p(\mathbf{z}_t \mid  \mathbf{z}_{t+1})p(\mathbf{z}_{t+1}) \mathop{\mathrm{d}\mathbf{z}_{t+1}},&& \text{for} &&t = 1,\dots,T-1.&&
\end{align}
Like the standard non-hierarchical case, the model is fit by maximizing the VLB of $\log p(\mathbf{x})$, but now includes additional terms specifying the assumed factorization of the latent variables:
\begin{align}
    \log p(\mathbf{x}) \geq \mathrm{VLB}(\mathbf{x}) = \mathbb{E}_{q(\mathbf{z}_{1:T} \mid \mathbf{x})} \left[ \log \frac{p(\mathbf{x}, \mathbf{z}_{1:T})}{q(\mathbf{z}_{1:T} \mid \mathbf{x})} \right].
\end{align}
In the following sections, we explore both practical and theoretical challenges in using the aforementioned objective for training HLVMs -- particularly under bottom-up inference -- and introduce a new perspective using a concept we call \textit{generative feedback}, which augments the commonly cited but incomplete efficiency-based view.
\newpage
\section{Generative Feedback}
\label{subsec: Generative Feedback}
\cite{burda2015importance} and~\cite{sonderby2016ladder} both found that hierarchical latent variable models with purely bottom-up inference are typically not capable of utilizing more than two layers of latent variables. This often manifests as \textit{posterior collapse}, whereby the posterior distribution collapses to a standard Gaussian prior, failing to learn meaningful representations and effectively deactivating latent variables. Generally speaking, having a learnable inference network can destabilize training by creating asynchronous learning dynamics with the generator, often leading to posterior collapse due to a constantly shifting target.

To better understand why bottom-up inference is challenging for even modestly deep hierarchies, we start by noting the asymmetry between the associated generative and inference models in Equations~\ref{eq: hvae_gen} and~\ref{eq: hvae_inf} respectively.~\cite{burda2015importance,sohl2015deep} point this out as a source of difficulty in training the inference model efficiently, since there is no way to express each term in the VLB as an expectation under a distribution over a single latent variable.~\cite{luo2022understanding,bishop2023} present a similar efficiency-based argument against bottom-up inference in hierarchical latent variable models.

We claim that efficiency-based arguments paint an incomplete picture; the main reason one should avoid bottom-up inference is the lack of direct \textit{feedback} from the generative model.
To show why generative feedback is important, we stress that the purpose of the inference model is to perform \textit{Bayesian inference} at any given layer in the hierarchy. Concretely, its job is to compute the posterior distribution $q(\mathbf{z}_t \mid \mathbf{x})$ over each latent variable $\mathbf{z}_t$ by:
\begin{align}
    q(\mathbf{z}_t \mid \mathbf{x}) &= \frac{p(\mathbf{x} \mid \mathbf{z}_t)p(\mathbf{z}_t)}{p(\mathbf{x})}
    \\ & \propto p(\mathbf{x} \mid \mathbf{z}_t) \int p(\mathbf{z}_t \mid  \mathbf{z}_{t+1})p(\mathbf{z}_{t+1}) \mathop{\mathrm{d}\mathbf{z}_{t+1}}, \label{eq: qzx_bayes}
\end{align}
which clearly shows that each posterior $q(\mathbf{z}_t \mid \mathbf{x})$ is not only proportional to the current layer's prior $p(\mathbf{z}_t)$ but also depends on the layer above's $p(\mathbf{z}_{t+1})$, and so on, following the reverse of the generative Markov chain, as specified in Equation~\ref{eq: hvae_gen}.

Therefore, it stands to reason that interleaving feedback from each transition in the generative model into each respective transition in the inference model can align the inference network with the generative model and improve the inference procedure. To that end, we can take Equation~\ref{eq: qzx_bayes} and rewrite the posterior distribution over each $\mathbf{z}_t$ such that it contains a more explicit dependency on $\mathbf{z}_{t+1}$:
\begin{align}
    q(\mathbf{z}_t \mid \mathbf{z}_{t+1}, \mathbf{x}) & = \frac{p(\mathbf{x} \mid \mathbf{z}_{t}, \mathbf{z}_{t+1})p(\mathbf{z}_{t} \mid \mathbf{z}_{t+1})}{p(\mathbf{x} \mid \mathbf{z}_{t+1})}
    \\[5pt] & = \frac{p(\mathbf{x} \mid \mathbf{z}_{t})p(\mathbf{z}_{t} \mid \mathbf{z}_{t+1})}{p(\mathbf{x} \mid \mathbf{z}_{t+1})} \cdot \frac{p(\mathbf{z}_{t+1})}{p(\mathbf{z}_{t+1})},
\end{align}
where the likelihood term $p(\mathbf{x} \mid \mathbf{z}_{t}, \mathbf{z}_{t+1})$ simplifies to $p(\mathbf{x} \mid \mathbf{z}_{t})$ due to the conditional independence relation $\mathbf{x} \ci \mathbf{z}_{t+1} \mid \mathbf{z}_t$ specified by the generative Markov chain. Now simply re-anchoring the time indices from $t$ and $t+1$ to $t-1$ and $t$, respectively, reveals:
\begin{align}
    q(\mathbf{z}_{t-1} \mid \mathbf{z}_{t}, \mathbf{x}) & = \frac{p(\mathbf{x} \mid \mathbf{z}_{t-1})p(\mathbf{z}_{t-1} \mid \mathbf{z}_{t})p(\mathbf{z}_t)}{p(\mathbf{x}, \mathbf{z}_{t})}
    \\[5pt] & \propto p(\mathbf{x} \mid \mathbf{z}_{t-1})p(\mathbf{z}_{t-1} \mid \mathbf{z}_{t})p(\mathbf{z}_t).
\end{align}
The posterior $q(\mathbf{z}_{t-1} \mid \mathbf{z}_{t}, \mathbf{x})$ now follows the same topological ordering of the latent variables as the prior $p(\mathbf{z}_{t-1} \mid \mathbf{z}_t)$, and it coincides with the \textit{top-down} inference model in HVAEs~\citep{sonderby2016ladder,kingma2016improved}. Figure~\ref{fig: hvae} shows how this top-down structure compares to the bottom-up approach. An added benefit of the top-down approach is that the generative model can also receive data-dependent feedback from the inference procedure, which can be beneficial in practice. 

\begin{figure}[!t]
    \centering
    \begin{tikzpicture}[thick,scale=1, every node/.style={scale=1}]
        \node[latent] (zt) {$\hat{\mathbf{z}}_T$};
        \draw node[draw=none, scale=0.75, below=6pt of zt] (z3) {\hspace{0.5pt}\rotatebox{90}{$\mathbf{\cdots}$}};
        \node[latent, below=13pt of z3] (z2) {$\hat{\mathbf{z}}_2$};
        \node[latent, below=15pt of z2] (z1) {$\hat{\mathbf{z}}_1$};
        \edge[-]{zt}{z3}
        \edge[-{Latex[scale=1.0]}]{z3}{z2}
        \edge[-{Latex[scale=1.0]}]{z2}{z1}

        \node[latent, left=25pt of zt] (zzt) {$\mathbf{z}_T$};
        \draw node[draw=none, scale=0.75, below=13pt of zzt] (zz3) {\hspace{0.5pt}\rotatebox{90}{$\mathbf{\cdots}$}};
        \node[latent, left=25pt of z2] (zz2) {$\mathbf{z}_2$};
        \node[latent, left=25pt of z1] (zz1) {$\mathbf{z}_1$};
        \node[obs, below=15pt of zz1] (xx) {$\mathbf{x}$};

        \node[latent, draw=none, left=25pt of zzt] (st) {$\mathcal{N}(0,\sigma_T^2)$};
        \edge[-{Latex[scale=1.0]}]{st}{zzt}
        \node[latent, draw=none, left=25pt of zz2] (s2) {$\mathcal{N}(0,\sigma_2^2)$};
        \edge[-{Latex[scale=1.0]}]{s2}{zz2}
        \node[latent, draw=none, left=25pt of zz1] (s1) {$\mathcal{N}(0,\sigma_1^2)$};
        \edge[-{Latex[scale=1.0]}]{s1}{zz1}

        \node[latent, draw=none, right=25pt of zt] (dt) {$\mathbf{d}_T$};
        \draw node[draw=none, scale=0.75, below=13pt of dt] (d3) {\hspace{0.5pt}\rotatebox{90}{$\mathbf{\cdots}$}};
        \node[latent, draw=none, right=25pt of z2] (d2) {$\mathbf{d}_2$};
        \node[latent, draw=none, right=25pt of z1] (d1) {$\mathbf{d}_1$};
        \node[obs, below=15pt of d1] (x) {$\mathbf{x}$};
        
        \edge[blue, -{Latex[scale=1.0]}]{d3}{dt}
        \edge[blue, -]{d2}{d3}
        \edge[blue, -{Latex[scale=1.0]}]{d1}{d2}
        \edge[blue, -{Latex[scale=1.0]}]{x}{d1}
        \edge[densely dashed,-]{dt}{zt}
        \edge[densely dashed,-]{d2}{z2}
        \edge[densely dashed,-]{d1}{z1}
        \edge[blue,-{Latex[scale=1.0]}]{xx}{zz1}
        \edge[blue,-{Latex[scale=1.0]}]{zz1}{zz2}
        \edge[blue,-]{zz2}{zz3}
        \edge[blue,-{Latex[scale=1.0]}]{zz3}{zzt}
        \edge[-{Latex[scale=1.0]}]{zzt}{zt}
        \edge[-{Latex[scale=1.0]}]{zz2}{z2}
        \edge[-{Latex[scale=1.0]}]{zz1}{z1}
    \end{tikzpicture}
    \caption{\textbf{A Ladder Network}. The latent variables $\mathbf{z}_1, \mathbf{z}_2, \dots, \mathbf{z}_T$ are noisy representations of $\mathbf{x}$, and $\mathbf{d}_1, \mathbf{d}_2, \dots, \mathbf{d}_T$ are clean representations; both sets are produced by a shared encoder (blue arrows). The variables $\hat{\mathbf{z}}_1, \hat{\mathbf{z}}_2, \dots, \hat{\mathbf{z}}_T$ are outputs of denoising functions where $\hat{\mathbf{z}}_t = g_t(\mathbf{z}_t, \hat{\mathbf{z}}_{t+1})$. Notice how $g_t(\cdot)$ receives both bottom-up and top-down information. The dashed horizontal lines denote local cost functions used to minimize  $\|\hat{\mathbf{z}}_t - \mathbf{d}_t \|^2_2$. The main difference compared to denoising diffusion models is that here the denoising targets $\mathbf{d}_t$ are learned representations of $\mathbf{x}$ rather than fixed, increasingly noisier versions of $\mathbf{x}$.
    }
    \label{fig: ladder}
\end{figure}
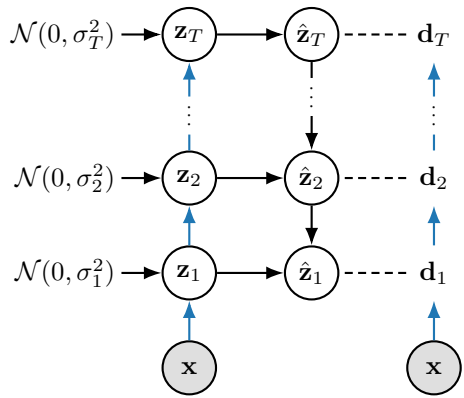
\paragraph{Ladder Networks.} \cite{valpola2015neural,rasmus2015semi} were, to the best of our knowledge, the first to introduce such lateral feedback connections between the inference and generative paths in hierarchical latent variable models. They called their denoising autoencoder a \textit{ladder network} (see Figure~\ref{fig: ladder}), which later inspired the ladder VAE~\citep{sonderby2016ladder}.~\cite{valpola2015neural} argue that incorporating lateral feedback connections enables the higher layers to learn abstract invariant representations as they no longer have to retain all the details about the input. Concretely, as depicted in Figure~\ref{fig: ladder}, each denoised variable $\hat{\mathbf{z}}_t \coloneqq g_t(\mathbf{z}_t, \hat{\mathbf{z}}_{t+1})$ is computed using a denoising function $g_t(\cdot)$ which receives bottom-up feedback from $\mathbf{z}_t$ and top-down feedback from $\hat{\mathbf{z}}_{t+1}$. For further discussion please refer to \S\ref{sec: variational diffusion models}.
\section{Top-down Inference}
\label{subsec: Top-down Inference Model}
The joint approximate posterior $q(\mathbf{z}_{1:T} \mid \mathbf{x})$ can be alternatively factorized into the \textit{top-down} inference model~\citep{sonderby2016ladder,kingma2016improved}. As mentioned in Section~\ref{subsec: Generative Feedback}, the top-down inference model follows the same topological ordering of the latent variables as the generative model, that is:
\begin{align}
    q(\mathbf{z}_{1:T} \mid \mathbf{x}) & = q(\mathbf{z}_T \mid \mathbf{x}) 
    q(\mathbf{z}_{T-1} \mid \mathbf{z}_{T}, \mathbf{x}) 
    \cdots
    q(\mathbf{z}_1 \mid \mathbf{z}_2, \mathbf{x})
    \\[5pt] & 
    = q(\mathbf{z}_T \mid \mathbf{x}) \prod_{t=2}^{T} q(\mathbf{z}_{t-1} \mid \mathbf{z}_t, \mathbf{x}).
\end{align}
Variants of the top-down inference model have featured in much deeper state-of-the-art HVAEs for sample generation~\citep{maaloe2019biva,vahdat2020nvae,child2020very,shu2022bit} and approximate counterfactual inference~\citep{pmlr-v202-de-sousa-ribeiro23a,monteiro2022measuring}. As we will show, the top-down hierarchical latent variable model serves as the basis for parameterizing diffusion models.

\paragraph{Variational Lower Bound.} For now, we derive the corresponding VLB to obtain a concrete optimization objective: $\mathrm{VLB}(\mathbf{x})$
\begin{align}
    \mathrm{VLB}(\mathbf{x})
    & = \mathbb{E}_{q(\mathbf{z}_{1:T} \mid \mathbf{x})} \left[ \log \frac{p(\mathbf{x}, \mathbf{z}_{1:T})}{q(\mathbf{z}_{1:T} \mid \mathbf{x})} \right]
    \\[5pt] & = \mathbb{E}_{q(\mathbf{z}_{1:T} \mid \mathbf{x})} \left[ \log \frac{p(\mathbf{z}_T) p(\mathbf{x} \mid \mathbf{z}_1) \prod_{t=2}^{T}p(\mathbf{z}_{t-1} \mid \mathbf{z}_t)}{q(\mathbf{z}_T \mid \mathbf{x}) \prod_{t=2}^{T} q(\mathbf{z}_{t-1} \mid \mathbf{z}_t, \mathbf{x})} \right]
    \\[5pt] & =
    \mathbb{E}_{q(\mathbf{z}_{1:T} \mid \mathbf{x})}\Bigg[\log p(\mathbf{x} \mid \mathbf{z}_{1}) + \log \frac{p(\mathbf{z}_T)}{ q(\mathbf{z}_T \mid \mathbf{x})} \nonumber
    \\[2pt] & \quad\qquad\qquad + \sum_{t=2}^T \log  \frac{p(\mathbf{z}_{t-1} \mid \mathbf{z}_t)}{q(\mathbf{z}_{t-1} \mid \mathbf{z}_t, \mathbf{x})}\Bigg]
    \\[5pt] & = \mathbb{E}_{q(\mathbf{z}_{1} \mid  \mathbf{x})}\left[\log p(\mathbf{x} \mid \mathbf{z}_{1})\right] + \mathbb{E}_{q(\mathbf{z}_{T} \mid \mathbf{x})}\left[\log \frac{p(\mathbf{z}_T)}{ q(\mathbf{z}_T \mid \mathbf{x})}\right] \nonumber 
    \\[2pt] & \quad\qquad\qquad + \sum_{t=2}^T \mathbb{E}_{q(\mathbf{z}_{t-1}, \mathbf{z}_{t} \mid \mathbf{x})} \left[\log \frac{p(\mathbf{z}_{t-1} \mid \mathbf{z}_t)}{q(\mathbf{z}_{t-1} \mid \mathbf{z}_t, \mathbf{x})} \right]
    \\[5pt] & = \mathbb{E}_{q(\mathbf{z}_{1} \mid  \mathbf{x})}\left[\log p(\mathbf{x} \mid \mathbf{z}_{1})\right] + \mathbb{E}_{q(\mathbf{z}_{T} \mid \mathbf{x})}\left[\log \frac{p(\mathbf{z}_T)}{ q(\mathbf{z}_T \mid \mathbf{x})}\right] \nonumber 
    \\[2pt] & \quad\qquad\qquad + \sum_{t=2}^T \mathbb{E}_{q(\mathbf{z}_{t} \mid \mathbf{x})} \left[\mathbb{E}_{q(\mathbf{z}_{t-1} \mid \mathbf{z}_{t}, \mathbf{x})} \left[ \log \frac{p(\mathbf{z}_{t-1} \mid \mathbf{z}_t)}{q(\mathbf{z}_{t-1} \mid \mathbf{z}_t, \mathbf{x})} \right]\right]
    \\[5pt] & = \mathbb{E}_{q(\mathbf{z}_{1} \mid \mathbf{x})}\left[\log p(\mathbf{x} \mid \mathbf{z}_{1})\right] - D_{\mathrm{KL}}(q(\mathbf{z}_{T} \mid \mathbf{x}) \parallel p(\mathbf{z}_{T})) \nonumber 
    \\[2pt] & \quad\qquad\qquad - \sum_{t=2}^T \mathbb{E}_{q(\mathbf{z}_{t} \mid \mathbf{x})} \left[D_{\mathrm{KL}}(q(\mathbf{z}_{t-1} \mid \mathbf{z}_{t}, \mathbf{x}) \parallel p(\mathbf{z}_{t-1} \mid \mathbf{z}_{t}))\right]. \label{eq: hvae_elbo}
\end{align}
It is well worth dedicating some time to understanding the details of the above derivation and the resulting expression, as it is the \textit{exact} objective optimized by diffusion models as well.

One thing to notice is that it comprises the familiar trade-off between minimizing input reconstruction error and keeping the hierarchical approximate posterior $q(\mathbf{z}_{1:T} \mid \mathbf{x})$ close to the hierarchical prior $p(\mathbf{z}_{1:T})$. In contrast to standard VAEs, the prior in HVAEs is typically learned from data rather than being fixed, as this affords greater flexibility and allows us to match the data distribution much more easily~\citep{kingma2016improved,hoffman2016elbo,tomczak2018vae}.
\begin{figure}[!t]
    \centering
    \begin{subfigure}{\columnwidth}
    \centering  \includegraphics[trim=0 50 0 0, clip, width=\textwidth]{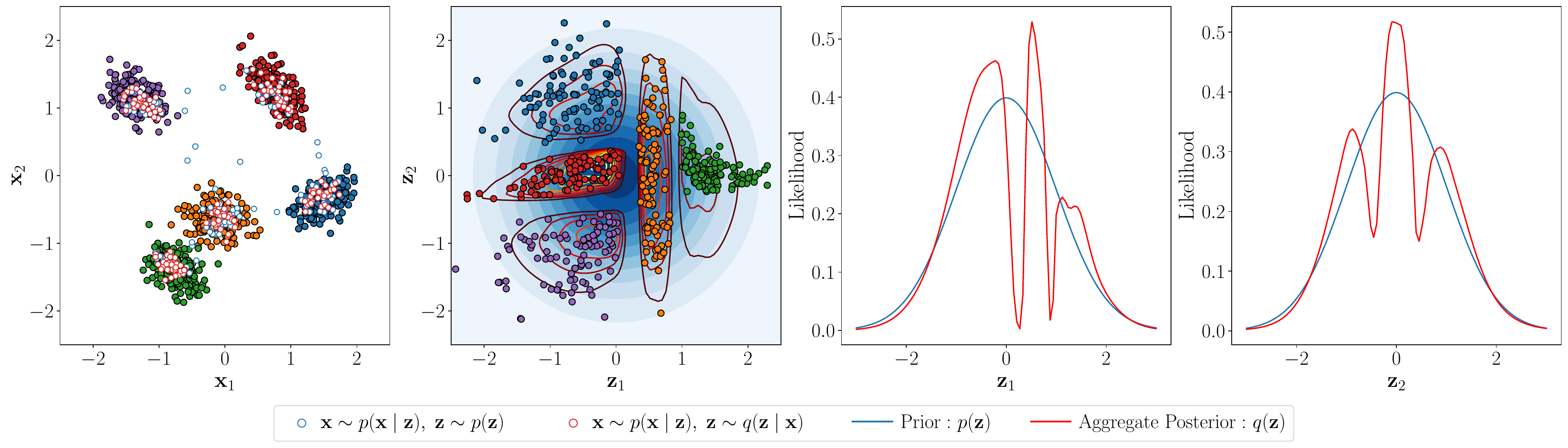}
    \end{subfigure}
    \begin{subfigure}{\columnwidth}
    \centering    \includegraphics[width=\textwidth]{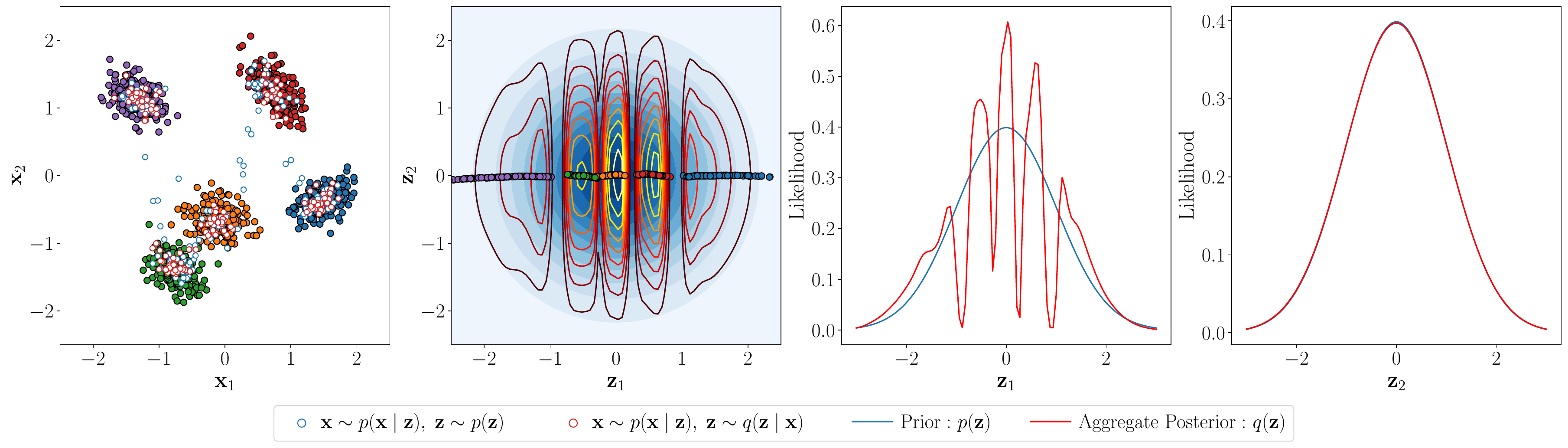}
    \end{subfigure}
    \caption{\textbf{Demonstration of the `hole problem'}. Results are from a single stochastic layer VAE trained on a 2D toy dataset with five clusters. The latent variable $\mathbf{z}$ is also 2-dimensional for illustration purposes.
    The leftmost column shows the dataset, overlaid with reconstructed datapoints (red border) and random samples from the generative model (blue border). The remaining columns show the assumed prior $p(\mathbf{z}) = \mathcal{N}(\mathbf{z};0, \mathbf{I})$ (blue contours) overlaid with the aggregate posterior $q(\mathbf{z}) = \sum_{i=1}^N q(\mathbf{z} \mid \mathbf{x}_i) / N$. As shown, there are regions with high density under the prior which are assigned low density under the aggregate posterior. This affects the quality of the random samples since we are likely to sample from regions in $p(\mathbf{z})$ not covered by the data. Further, the bottom row shows a common occurrence in VAEs where latent variable(s) are not activated/used at all by the model, in this case, $\mathbf{z}_2$ was not used.
    }
    \label{fig: hole}
\end{figure}
\newpage
\section{The W[H]ole Problem}
\label{subsec: main hole}
A primary issue with VAEs is the so-called \textit{hole problem}~\citep{rezende2018taming}. The hole problem refers to the often observed mismatch between the so-called aggregate posterior $q(\mathbf{z})$ and the prior $p(\mathbf{z})$ over the latent variables~\citep{makhzani2015adversarial,hoffman2016elbo}. 

The aggregate posterior is simply the average posterior distribution over the dataset $\mathcal{D} = \{\mathbf{x}_i\}_{i=1}^N$, that is:
\begin{align}
    &&q(\mathbf{z}) = \int q(\mathbf{z} \mid \mathbf{x})p_\mathcal{D}(\mathbf{x}) \mathop{\mathrm{d}\mathbf{x}} 
    , && p_{\mathcal{D}}(\mathbf{x}) = \frac{1}{N}\sum_{i=1}^N \delta(\mathbf{x} - \mathbf{x}_i),&&
\end{align}
where $p_\mathcal{D}(\mathbf{x})$ is the \textit{empirical distribution}, constructed by a Dirac delta function $\delta(\cdot)$ centered on each training datapoint $\mathbf{x}_i$. As shown in Figure~\ref{fig: hole}, there can be regions with \textit{high} probability density under the prior which have \textit{low} probability density under the aggregate posterior. As a result, the quality of generated samples can be affected when the decoder receives $\mathbf{z}$'s sampled from regions in $p(\mathbf{z})$ which are not well-covered by the data. 
This phenomenon can be interpreted as a type of distribution shift/drift which occurs when sampling from the model. 
One possible workaround is to use $q(\mathbf{z})$ as the prior instead, however, computing $q(\mathbf{z})$ is usually computationally prohibitive in practice as it requires marginalizing out a (possibly very large) dataset.

As we will show, diffusion models circumvent the hole problem by defining the aggregate posterior to be equal to the prior by construction. This has profound implications as we can avoid drift when sampling.
\chapter{Variational Diffusion Models}
\label{sec: variational diffusion models}
A diffusion probabilistic model~\citep{sohl2015deep} can be understood as a hierarchical VAE with a particular choice of inference and generative model. Like HVAEs, diffusion models are deep latent variable models that maximize the variational lower bound of the log-likelihood of the data (i.e. the ELBO). Diffusion models were largely inspired by ideas from statistical physics rather than variational Bayesian methods, so they come with a different set of modelling choices, advantages and nomenclature. The general idea behind diffusion models is to define a \textit{fixed} forward diffusion process that converts any complex data distribution into a tractable distribution, and then learn a generative model that reverses this diffusion process. Figure~\ref{fig: hvae2} compares diffusion models with (top-down inference) HVAEs and the following sections explore the details. Diffusion models have the following distinctive properties:
\begin{enumerate}[(i)]
    \item The joint posterior $q(\mathbf{z}_{1:T} \mid \mathbf{x})$ is fixed rather than learned from observed data;
    \item Each latent variable $\mathbf{z}_t$ has the same dimensionality as $\mathbf{x}$;
    \item The aggregate posterior $q(\mathbf{z}_T)$ is equal to the prior $p(\mathbf{z}_T)$ by construction;
    \item The functional form of the inference model is identical to that of the generative model. This corresponds exactly to the top-down inference model structure used in HVAEs;
    \item A single neural network is shared across all levels of the latent variable hierarchy, and each layer can be trained independently;
    \item They optimize a \textit{weighted} objective that suppresses modelling effort on imperceptible details and better aligns with human perception.
\end{enumerate}

Recent model innovations~\citep{ho2020denoising} -- along with insights from stochastic processes~\citep{anderson1982reverse} and score-based generative modelling~\citep{hyvarinen2005estimation,vincent2011connection,song2019generative,song2021scorebased} -- have yielded a myriad of impressive synthesis results~\citep{nichol2021improved,dhariwal2021diffusion,nichol2022glide,ho2022cascaded,rombach2022high,saharia2022photorealistic,hoogeboom2022equivariant}. 
\cite{kingma2021variational,kingma2023understanding} introduced a family of diffusion-based generative models they call Variational Diffusion Models (VDMs), and showed us that: 
\begin{enumerate}[(i)]
    \item The latent hierarchy can be made infinitely deep\footnote{This notion was concurrently explored by~\cite{song2021scorebased,huang2021variational,vahdat2021score}.} via a continuous-time model where $T \to \infty$; 
    \item The continuous-time VLB is invariant to the noise schedule\footnote{Except for the signal-to-noise ratio at its endpoints (see Section~\ref{subsubsec: Invariance to the Noise Schedule}).}, meaning we can learn/adapt our noise schedule such that it minimizes the variance of the resulting Monte Carlo estimator of the loss;
    \item Although \textit{weighted} diffusion objectives \textit{appear} markedly different from regular maximum likelihood training, they all implicitly optimize some instance of the ELBO;
    \item VDMs are capable of state-of-the-art image synthesis, showing that standard maximum likelihood-based training objectives (i.e. the ELBO) are not inherently at odds with perceptual quality.
\end{enumerate}
    \begin{landscape}
\begin{figure}[!t]
    \centering
    \hfill
    \begin{subfigure}{.24\columnwidth}
        \centering
        \begin{tikzpicture}[thick,scale=1, every node/.style={scale=1}]
            \node[latent] (x) {$\hat{\mathbf{x}}_{\boldsymbol{\theta}}$};
            \node[latent, below=15pt of x] (zt) {$\mathbf{z}_T$};
            \draw node[draw=none, scale=0.75, below=6pt of zt] (z3) {\hspace{0.5pt}\rotatebox{90}{$\mathbf{\cdots}$}};
            \node[latent, below=13pt of z3] (z2) {$\mathbf{z}_2$};
            \node[latent, below=15pt of z2] (z1) {$\mathbf{z}_1$};
            \edge[-]{zt}{z3}
            \edge[-{Latex[scale=1.0]}]{z3}{z2}
            \edge[-{Latex[scale=1.0]}]{z2}{z1}
            \draw [-{Latex[scale=1.0]}] (x) to [out=225,in=135] (z2);
            \draw [-{Latex[scale=1.0]}] (x) to [out=225,in=135] (z1);
            \node[latent, draw=none, right=-4pt of x, yshift=-16pt] (eq1) {$q(\mathbf{z}_T)$};
            \node[latent, draw=none, right=-4pt of z2, yshift=-16pt] (eq2) {$q(\mathbf{z}_1 \mid \mathbf{z}_2,  \hat{\mathbf{x}}_{\boldsymbol{\theta}})$};
            \node[latent, draw=none, right=-4pt of z3, xshift=6pt, yshift=-0pt] (eq3) {$q(\mathbf{z}_{t-1} \mid \mathbf{z}_{t}, \hat{\mathbf{x}}_{\boldsymbol{\theta}})$};
        \end{tikzpicture}
        \caption{Top-down Hierarchy}
    \end{subfigure}
    \hfill
    \begin{subfigure}{.2\columnwidth}
        \centering
        \begin{tikzpicture}[thick,scale=1, every node/.style={scale=1}]
            \node[obs] (x) {$\mathbf{x}$};
            \node[latent, below=15pt of x] (zt) {$\mathbf{z}_T$};
            \draw node[draw=none, scale=0.75, below=6pt of zt] (z3) {\hspace{0.5pt}\rotatebox{90}{$\mathbf{\cdots}$}};
            \node[latent, below=13pt of z3] (z2) {$\mathbf{z}_2$};
            \node[latent, below=15pt of z2] (z1) {$\mathbf{z}_1$};
            \edge[-]{zt}{z3}
            \edge[-{Latex[scale=1.0]}]{z3}{z2}
            \edge[-{Latex[scale=1.0]}]{z2}{z1}
            \draw [-{Latex[scale=1.0]}] (x) to [out=225,in=135] (z2);
            \draw [-{Latex[scale=1.0]}] (x) to [out=225,in=135] (z1);
            \draw [densely dashed, -{Latex[scale=1.0]}] (zt) to [out=45,in=-45] (x);
            \draw [densely dashed, -{Latex[scale=1.0]}] (z2) to [out=45,in=-45] (x);
            \draw [densely dashed, -{Latex[scale=1.0]}] (z1) to [out=45,in=-45] (x);
        \end{tikzpicture}
        \caption{Diffusion Model}
    \end{subfigure}
    \hfill
    \begin{subfigure}{.26\columnwidth}
        \centering       
        \begin{tikzpicture}[thick,scale=1, every node/.style={scale=1}]
            \node[latent] (zt) {$\mathbf{z}_T$};
            \draw node[draw=none, scale=0.75, below=6pt of zt] (z3) {\hspace{0.5pt}\rotatebox{90}{$\mathbf{\cdots}$}};
            \node[latent, below=13pt of z3] (z2) {$\mathbf{z}_2$};
            \node[latent, below=15pt of z2] (z1) {$\mathbf{z}_1$};
            \edge[-]{zt}{z3}
            \edge[-{Latex[scale=1.0]}]{z3}{z2}
            \edge[-{Latex[scale=1.0]}]{z2}{z1}
            \node[latent, draw=none, right=-4pt of z2, yshift=-16pt] (eq2) {$q(\mathbf{z}_1 \mid \mathbf{z}_2, \mathbf{d}_1)$};
            \node[latent, draw=none, right=-4pt of z3, xshift=6pt, yshift=-0pt] (eq3) {$q(\mathbf{z}_{t-1} \mid \mathbf{z}_t, \mathbf{d}_t)$};

            \node[latent, draw=none, left=15pt of zt] (dt) {$\mathbf{d}_T$};
            \draw node[draw=none, scale=0.75, below=13pt of dt] (d3) {\hspace{0.5pt}\rotatebox{90}{$\mathbf{\cdots}$}};
            \node[latent, draw=none, left=15pt of z2] (d2) {$\mathbf{d}_2$};
            \node[latent, draw=none, left=15pt of z1] (d1) {$\mathbf{d}_1$};
            \node[obs, below=15pt of d1] (x) {$\mathbf{x}$};
            
            \edge[-{Latex[scale=1.0]}]{d3}{dt}
            \edge[-]{d2}{d3}
            \edge[-{Latex[scale=1.0]}]{d1}{d2}
            \edge[-{Latex[scale=1.0]}]{x}{d1}
            \edge[-{Latex[scale=1.0]}]{dt}{zt}
            \edge[-{Latex[scale=1.0]}]{d2}{z2}
            \edge[-{Latex[scale=1.0]}]{d1}{z1}
        \end{tikzpicture}
        \caption{HVAE Inference Model}
    \end{subfigure}
    \hfill
    \begin{subfigure}{.26\columnwidth}
        \centering
        \begin{tikzpicture}[thick,scale=1, every node/.style={scale=1}]
            \node[latent, below] (zt) {$\mathbf{z}_T$};
            \draw node[draw=none, scale=0.75, below=6pt of zt] (z3) {\hspace{0.5pt}\rotatebox{90}{$\mathbf{\cdots}$}};
            \node[latent, below=13pt of z3] (z2) {$\mathbf{z}_2$};
            \node[latent, below=15pt of z2] (z1) {$\mathbf{z}_1$};
            \node[obs, below=15pt of z1] (x) {$\mathbf{x}$};
            \node[latent, draw=none, left=15pt of z1] (d1) {${\mathbf{d}}_1$};
            \node[latent, draw=none, left=15pt of z2] (d2) {${\mathbf{d}}_2$};

            \draw node[draw=none, scale=0.75, left=4pt of z3] (d3) {\hspace{0.5pt}\rotatebox{55}{$\mathbf{\cdots}$}};
            
            \edge[-]{zt}{z3}
            \edge[blue,-]{zt}{d3}
            \edge[blue,-{Latex[scale=1.0]}]{d3}{d2}
            \edge[blue, -{Latex[scale=1.0]}]{z2}{d1}
        
            \edge[-{Latex[scale=1.0]}]{z3}{z2}
            \edge[-{Latex[scale=1.0]}]{z2}{z1}
            \edge[-{Latex[scale=1.0]}]{z1}{x}
            \edge[-{Latex[scale=1.0]}]{d1}{z1}
            \edge[-{Latex[scale=1.0]}]{d2}{z2}

            \node[latent, draw=none, right=-4pt of z2, yshift=-16pt] (eq2) {$q(\mathbf{z}_1 \mid \mathbf{z}_2, \mathbf{d}_1)$};
            \node[latent, draw=none, right=-4pt of z3, xshift=6pt, yshift=-0pt] (eq3) {$q(\mathbf{z}_{t-1} \mid \mathbf{z}_{t}, \mathbf{d}_t)$};
        \end{tikzpicture}
        \caption{Reverse Process}
    \end{subfigure}
    \hfill
    \caption{\textbf{Probabilistic graphical models of HVAEs and diffusion models}. \textbf{(a)} The general top-down hierarchical latent variable model. \textbf{(b)} The top-down model used to specify diffusion models, where $q(\mathbf{z}_T \mid \mathbf{x}) = q(\mathbf{z}_T)$ by construction. Here the posterior $q(\mathbf{z}_{1:T} \mid \mathbf{x})$ is a fixed noising process, so the modelling task is bottom-up prediction of $\mathbf{x}$ from each $\mathbf{z}_t$, i.e. denoising (dashed lines). \textbf{(c)} The top-down model used for posterior inference in HVAEs. It consists of a deterministic bottom-up pass to compute $\mathbf{d}_1,\dots,\mathbf{d}_T$, followed a stochastic top-down pass to compute $\mathbf{z}_T,\dots,\mathbf{z}_1$. \textbf{(d)} The reverse process of a diffusion model, i.e. the generative model. The main differences compared to (c) are that here the deterministic variables $\mathbf{d}_{T-1},\dots,\mathbf{d}_1$ do not depend on $\mathbf{x}$ nor have their own hierarchical dependencies. Further, the {\color{blue}blue} lines represent a denoising model $\hat{\mathbf{x}}_{\boldsymbol{\theta}} :\mathbf{z}_t \to \mathbf{d}_t$ which is \textit{shared} across the hierarchy.
    }
    \label{fig: hvae2}
\end{figure}
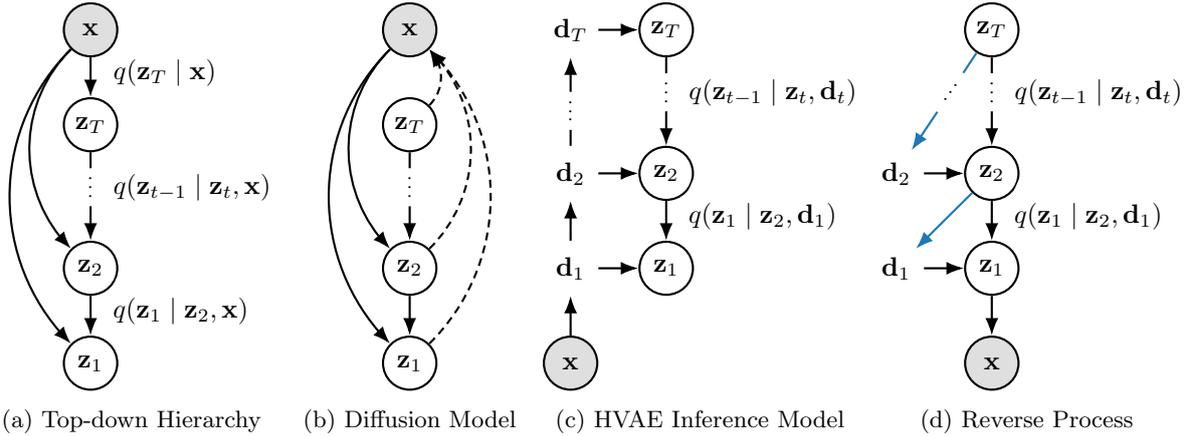
\end{landscape}
\subsection{On Representation Learning} 
One important distinction to make between hierarchical VAEs and diffusion models at this stage is that the role of the latent variables $\mathbf{z}_{1:T}$ is very different from a \textit{representation learning} perspective. In hierarchical VAEs, the posterior latents $\mathbf{z}_{1:T}$ are useful learned representations of $\mathbf{x}$, which \textit{increase} in semantic informativeness w.r.t. $\mathbf{x}$ as we go from $\mathbf{z}_1$ to $\mathbf{z}_T$. In diffusion models, the latent variables $\mathbf{z}_{1:T}$ generally have \textit{no} semantic meaning, and they \textit{decrease} in informativeness w.r.t. $\mathbf{x}$ as we go from $\mathbf{z}_1$ to $\mathbf{z}_T$. This is because each $\mathbf{z}_t$ in diffusion models is simply a noisy version of $\mathbf{x}$ following a Gaussian diffusion process.

The latent variables $ \mathbf{z}_1, \mathbf{z}_2, \dots, \mathbf{z}_T $ in ladder networks are noisy versions of the input $\mathbf{x}$, and the noise is additive isotropic Gaussian, like in diffusion models. However, the main difference is that each $ \mathbf{z}_t $ is a (noisy) learned representation of the input $ \mathbf{x} $ rather than simply a linear function of it, i.e.: $ \mathbf{z}_t = f_t(\mathbf{z}_{t-1}) + \sigma_t \boldsymbol{\epsilon} $, where $ f_t(\cdot) $ is a learned non-linear function and $ \sigma_t $ specifies the level of additive Gaussian noise corruption. In contrast, the diffusion process can be parameterized directly (linearly) in terms of $ \mathbf{x} $ as: $ \mathbf{z}_t = \alpha_t \mathbf{x} + \sigma_t \boldsymbol{\epsilon} $. Moreover, the denoising targets also differ in that diffusion models learn to denoise the input $ \mathbf{x} $ directly or linear functions of it (i.e. via noise $ \boldsymbol{\epsilon} = (\mathbf{z}_t - \alpha_t \mathbf{x}) / \sigma_t $ or velocity $ \mathbf{v} = \alpha_t \boldsymbol{\epsilon} - \sigma_t \mathbf{x}$ prediction), whereas ladder networks learn to denoise learned representations of the input $ \mathbf{d}_t $ given by the same non-linear functions $ f_t(\cdot) $ used to define the latent variables. 

The sum of the local cost functions in ladder networks is analogous to the diffusion loss -- in both cases, the goal is to reconstruct the input from different noisy versions of it. However, unlike diffusion models, ladder networks focus on hierarchical representation learning rather than sample generation and it is generally not possible to evaluate each local cost function independently of the others. Nonetheless, we argue that ladder networks can serve as valuable inspiration for designing diffusion-based representation learning methods in the future.

\begin{figure}[!t]
    \begin{subfigure}{\textwidth}
        \centering 
        \includegraphics[trim=0 0 0 0,clip,width=\textwidth]{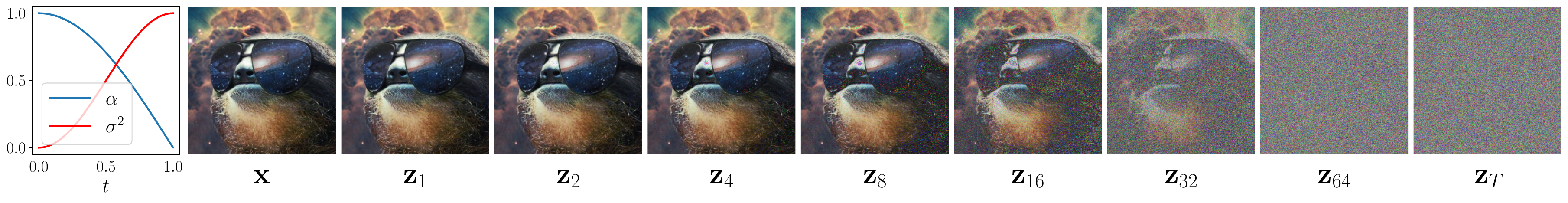}
        \label{fig:cosine_sloth}
    \end{subfigure}
    \\[-8pt]
    \begin{subfigure}{\textwidth}
        \centering
        \includegraphics[trim=0 0 0 0,clip,width=\textwidth]{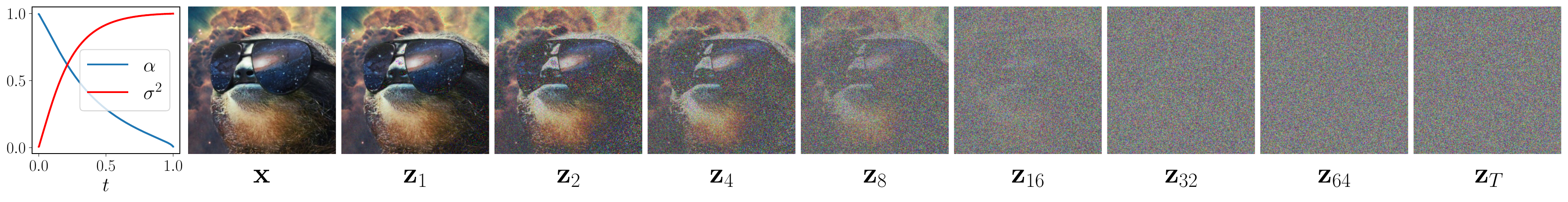}
        \label{fig:edm_sloth}
    \end{subfigure}
    \vspace{-25pt}
    \caption{\textbf{Example of a Gaussian diffusion process} ($T=100$). Showing two popular noise schedules in terms of $\{\alpha, \sigma^2\}$ as per Section~\ref{subsec: Gaussian Diffusion Process: Forward Time}. (\textit{Top}) cosine~\citep{nichol2021improved}; (\textit{Bottom}) EDM~\citep{karras2022elucidating}.}
    \label{fig: diff_sloth}
\end{figure}
\section{Forward Process: Gaussian Diffusion}
\label{subsec: Gaussian Diffusion Process: Forward Time}
A \textit{Gaussian diffusion process} gradually transforms data $\mathbf{x}$ into random noise over time, by adding increasing amounts of Gaussian noise at each timestep $t=0,\dots,1$ resulting in a set of latent variables $\mathbf{z}_0,\dots,\mathbf{z}_1$\footnote{Without loss of generality, and for consistency with the continuous-time case where $T \to \infty$, we now denote the latent variables as $\mathbf{z}_{0:1}$ rather than $\mathbf{z}_{1:T}$.}.
Each latent variable $\mathbf{z}_t$ is simply a noisy version of $\mathbf{x}$, and its distribution conditional on $\mathbf{x}$ is given by:
\begin{align}
    &&q(\mathbf{z}_t \mid \mathbf{x}) = \mathcal{N}\left(\mathbf{z}_t;\alpha_{t} \mathbf{x}, \sigma^2_{t} \mathbf{I}\right), && \mathbf{z}_t = \alpha_{t} \mathbf{x} + \sigma_{t}\boldsymbol{\epsilon}_t, &&
\end{align}
where $\boldsymbol{\epsilon}_t \sim \mathcal{N}\left(\boldsymbol{\epsilon}_t; 0, \mathbf{I}\right)$, $\alpha_{t} \in (0,1)$ and $\sigma^2_{t} \in (0,1)$ are chosen scalar valued functions of time $t \in [0,1]$. See Figs.~\ref{fig: diff_sloth} \&~\ref{fig: gd} for examples.

The key idea is to define the forward diffusion process such that the noisiest latent variable $\mathbf{z}_1$ at time $t=1$ is standard Gaussian distributed: $q(\mathbf{z}_1 \mid \mathbf{x}) = \mathcal{N}(\mathbf{z}_1; 0, \mathbf{I})$, thus $q(\mathbf{z}_1 \mid \mathbf{x}) = q(\mathbf{z}_1)$. To that end, the scaling coefficients $\alpha_{0} > \ldots > \alpha_{1}$ \textit{decrease} w.r.t. time $t$, whereas the noise variances $\sigma_{0}^2 < \ldots < \sigma_{1}^2$ \textit{increase} w.r.t. $t$. As we will show, this enables us to learn a generative Markov chain which starts from $\mathbf{z}_1 \sim q(\mathbf{z}_1)$ and reverses the forward diffusion process to obtain samples from the data distribution. The implications of this are profound; the \textit{aggregate} posterior $q(\mathbf{z}_1)$ is equal to the prior $p(\mathbf{z}_1)$ by construction, which circumvents the \textit{hole problem} in VAEs (see Figure~\ref{fig: hole}).~\cite{hoffman2016elbo} showed that the optimal prior is the aggregate posterior, as long as our posterior approximation is good enough.

\subsection{Variance Preserving Process} 
A \textit{variance-preserving} (VP) diffusion process is achieved by solving for the value of $\alpha_t$ such that the variance of the respective latent variable $\mathbb{V}[\mathbf{z}_t]$ is equal to the variance of the input data $\mathbb{V}[\mathbf{x}]$. This can be important from a modelling perspective, as adding increasing amounts of noise to the input alters its statistics and can affect learning.

We can begin by first applying some basic properties of \textit{variance} to simplify $\mathbb{V}[\mathbf{z}_t]$ as follows:
\begin{align}
    \mathbb{V}[\mathbf{z}_t] & = \mathbb{V}[\alpha_{t} \mathbf{x} + \sigma_{t}\boldsymbol{\epsilon}_t] 
    \\[5pt] & 
    = \mathbb{V}[\alpha_{t} \mathbf{x}] + \mathbb{V}[\sigma_{t}\boldsymbol{\epsilon}_t] 
    \\[5pt] & 
    = \alpha_{t}^2\mathbb{V}[\mathbf{x}] + \sigma_{t}^2\mathbb{V}[\boldsymbol{\epsilon}_t] 
    \\[5pt] & = \alpha_{t}^2\mathbb{V}[\mathbf{x}] + \sigma_{t}^2,
\end{align}
since $\mathbb{V}[\boldsymbol{\epsilon}_t] = 1$ by definition. Taking the result and solving for $\alpha_t$ yields
\begin{align}
    \alpha_{t}^2\mathbb{V}[\mathbf{x}] + \sigma_{t}^2 & = \mathbb{V}[\mathbf{x}] 
    \\[5pt] \alpha_{t}^2 & = \frac{\mathbb{V}[\mathbf{x}] - \sigma_{t}^2}{\mathbb{V}[\mathbf{x}]}
    \\[5pt] \implies \mathbb{V}[\mathbf{z}_t] & = \mathbb{V}[\mathbf{x}] \iff \alpha_{t}^2 = 1 - \frac{\sigma_{t}^2}{\mathbb{V}[\mathbf{x}]},
\end{align}
which further simplifies to $\alpha_t^2 = 1 - \sigma_t^2$ as long as our input data is standardized (i.e. $\mathbb{V}[\mathbf{x}] = 1$).

Choosing diffusion processes is a fairly deep research topic with significant practical implications. We have decided to cover VP processes only, as they are the most widely used in practice as of the time of this writing and can be relatively straightforward to understand. Furthermore, all other types of forward processes can be converted to a VP process without loss of generality. For more information on variance-exploding processes, for example, please refer to~\cite{song2021scorebased}. For details on learning the noise schedule from data see Appendix~\ref{subsubsec: Noise Schedule}.
\begin{landscape}
\begin{figure}[!t]
    \hfill
    \begin{subfigure}{.32\columnwidth}
        \centering
        \begin{tikzpicture}[thick,scale=1, every node/.style={scale=1}]
            \node[obs] (x) {$\mathbf{x}$};
            \node[latent, above=20pt of x] (z1) {$\mathbf{z}_1$};
            \node[latent, above=20pt of z1] (z2) {$\mathbf{z}_{2}$};
            \draw node[draw=none, scale=0.8, above=9pt of z2] (z3) {\hspace{0.5pt}\rotatebox{90}{$\mathbf{\cdots}$}};
            \draw [-{Latex[scale=1.0]}] (x) to [out=45,in=-45] (z2);
            \node[latent, above=9pt of z3] (zt) {$\mathbf{z}_T$};
            \edge[-{Latex[scale=1.0]}]{x}{z1}
            \node[rectangle, draw=none, fill=white, scale=1, above=2pt of x, xshift=-12pt] (eq1) {$\alpha_1$};
            \draw [-{Latex[scale=1.0]}] (x) to [out=45,in=-45] (zt);
            \node[rectangle, draw=none, fill=white, scale=1, right=22pt of z2, yshift=-13pt] (eq2) {$\alpha_T$};
            \node[rectangle, draw=none, scale=1, left=15pt of zt] (nt) {$\mathcal{N}\big(0, \sigma^2_T\mathbf{I}\big)$};
            \node[rectangle, draw=none, scale=1, left=15pt of z2] (n2) {$\mathcal{N}\big(0, \sigma^2_2\mathbf{I}\big)$};
            \node[rectangle, draw=none, scale=1, left=15pt of z1] (n1) {$\mathcal{N}\big(0, \sigma^2_1\mathbf{I}\big)$};
            \edge[-{Latex[scale=1.0]}]{nt}{zt}
            \edge[-{Latex[scale=1.0]}]{n2}{z2}
            \edge[-{Latex[scale=1.0]}]{n1}{z1}
            \node[rectangle, draw=none, fill=white, scale=1, right=11pt of z1, yshift=-1pt] (eq2) {$\alpha_2$};
        \end{tikzpicture}
        \caption{Gaussian Diffusion}
        \label{fig: gd}
    \end{subfigure}
    \hfill
    \begin{subfigure}{.32\columnwidth}
        \centering
        \begin{tikzpicture}[thick,scale=1, every node/.style={scale=1}]
            \node[obs] (x) {$\mathbf{x}$};
            \node[latent, above=20pt of x] (z1) {$\mathbf{z}_1$};
            \node[latent, above=20pt of z1] (z2) {$\mathbf{z}_{2}$};
            \draw node[draw=none, scale=0.8, above=6pt of z2] (z3) {\hspace{0.5pt}\rotatebox{90}{$\mathbf{\cdots}$}};
            \node[latent, above=12pt of z3] (zt) {$\mathbf{z}_T$};
            \edge[-{Latex[scale=1.0]}]{x}{z1}
            \node[rectangle, draw=none, fill=white, scale=1, above=1.5pt of x, xshift=18pt] (eq1) {$\alpha_{1|0}$};
            \node[rectangle, draw=none, fill=white, scale=1, above=1.5pt of z1, xshift=18pt] (eq2) {$\alpha_{2|1}$};
            \node[rectangle, draw=none, fill=white, scale=1, right=-0.4pt of z3] (eq3) {$\alpha_{t|t-1}$};
            
            \edge[-{Latex[scale=1.0]}]{z1}{z2}
            \node[rectangle, draw=none, scale=1, left=15pt of zt] (nt) {$\mathcal{N}\big(0, \sigma^2_{T|T-1}\mathbf{I}\big)$};
            \node[rectangle, draw=none, scale=1, left=15pt of z2] (n2) {$\mathcal{N}\big(0, \sigma^2_{2|1}\mathbf{I}\big)$};
            \node[rectangle, draw=none, scale=1, left=15pt of z1] (n1) {$\mathcal{N}\big(0, \sigma^2_{1|0}\mathbf{I}\big)$};
            \edge[-{Latex[scale=1.0]}]{nt}{zt}
            \edge[-{Latex[scale=1.0]}]{n2}{z2}
            \edge[-{Latex[scale=1.0]}]{n1}{z1}

            \edge[-]{z2}{z3}
            \edge[-{Latex[scale=1.0]}]{z3}{zt}
        \end{tikzpicture}
        \caption{Markovian Transitions}
        \label{fig: mt}
    \end{subfigure}
    \hfill
    \begin{subfigure}{.34\columnwidth}
        \centering
        \begin{tikzpicture}[thick,scale=1, every node/.style={scale=1}]
            \node[obs] (x) {$\mathbf{x}$};
            \node[latent, below=20pt of x] (zt) {$\mathbf{z}_T$};
            \draw node[draw=none, scale=0.8, below=6pt of zt] (z3) {\hspace{0.5pt}\rotatebox{90}{$\mathbf{\cdots}$}};
            \node[latent, below=12pt of z3] (z2) {$\mathbf{z}_2$};
            \node[latent, below=20pt of z2] (z1) {$\mathbf{z}_1$};
            \edge[-]{zt}{z3}
            \edge[-{Latex[scale=1.0]}]{z3}{z2}
            \edge[-{Latex[scale=1.0]}]{z2}{z1}
            \node[rectangle, draw=none, fill=white, scale=0.9, right=21pt of z2, yshift=30pt] (eq5) {$\alpha_{1}\sigma^2_{2|1}/\sigma^{2}_{2}$};
            \draw [-{Latex[scale=1.0]}] (x) to [out=-45,in=45] (z1);
            \node[rectangle, draw=none, fill=white, scale=0.9, right=12pt of zt] (eq6) {$\alpha_{2}\sigma^2_{3|2}/\sigma^{2}_{3}$};
            \draw [-{Latex[scale=1.0]}] (x) to [out=-45,in=45] (z2);
            \node[rectangle, draw=none, scale=1, left=15pt of zt] (nt) {$\mathcal{N}\big(0, \mathbf{I}\big)$};
            \node[rectangle, draw=none, scale=1, left=15pt of z2] (n2) {$\mathcal{N}\big(0, \sigma^2_{2|3}\mathbf{I}\big)$};
            \node[rectangle, draw=none, scale=1, left=15pt of z1] (n1) {$\mathcal{N}\big(0, \sigma^2_{1|2}\mathbf{I}\big)$};
            \edge[-{Latex[scale=1.0]}]{nt}{zt}
            \edge[-{Latex[scale=1.0]}]{n2}{z2}
            \edge[-{Latex[scale=1.0]}]{n1}{z1}
            \node[rectangle, draw=none, fill=white, scale=0.9, left=-6pt of z1, yshift=22pt] (eq3) {$\alpha_{2|1}\sigma^2_{1}/\sigma^{2}_{2}$};
            \node[rectangle, draw=none, fill=white, scale=0.9, left=-6pt of z2, yshift=31pt] (eq4) {$\alpha_{t|t-1}\sigma^2_{t-1}/\sigma^{2}_{t}$};
        \end{tikzpicture}
        \caption{Top-down Posterior}
        \label{fig: tdp}
    \end{subfigure}
    \caption{\textbf{Graphical model(s) describing a discrete-time Gaussian diffusion process} ($T$ timesteps in total). \textbf{(a)} Parameterization of the forward process in terms of the conditionals $q(\mathbf{z}_t \mid \mathbf{x})$ (ref. Section~\ref{subsec: Gaussian Diffusion Process: Forward Time}). Each latent variable $\mathbf{z}_t$ is a noisy version of $\mathbf{x}$ given by: $\mathbf{z}_t = \alpha_t \mathbf{x} + \sigma_t \boldsymbol{\epsilon}_t$, and $\boldsymbol{\epsilon}_t \sim \mathcal{N}(0, \mathbf{I})$. \textbf{(b)} Markov chain formed by a sequence of transition distributions $q(\mathbf{z}_t \mid \mathbf{z}_{t-1})$ (ref. Section~\ref{subsubsec: lgt}). Each latent variable is given by: $\mathbf{z}_t = \alpha_{t|t-1} \mathbf{z}_{t-1} + \sigma_{t|t-1} \boldsymbol{\epsilon}_t$, with parameters $\alpha_{t|t-1} \coloneqq \alpha_{t} / \alpha_{t-1}$ and $\sigma^2_{t|t-1} \coloneqq \sigma^2_{t} - \alpha^2_{t|t-1}\sigma^2_{t-1}$. \textbf{(c)} The top-down posterior is tractable due to Gaussian conjugacy: $q(\mathbf{z}_{t-1} \mid \mathbf{z}_t, \mathbf{x}) \propto q(\mathbf{z}_{t} \mid \mathbf{z}_{t-1})q(\mathbf{z}_{t-1} \mid \mathbf{x})$ (ref. Section~\ref{subsubsec: qzs}), where $q(\mathbf{z}_{t-1} \mid \mathbf{x})$ acts as a Gaussian prior and $q(\mathbf{z}_{t} \mid \mathbf{z}_{t-1})$ as a Gaussian likelihood.  
    This top-down posterior is used to specify the \textit{generative} model transitions as $p(\mathbf{z}_{t-1} \mid \mathbf{z}_{t}) = q(\mathbf{z}_{t-1} \mid \mathbf{z}_{t}, \mathbf{x} = \hat{\mathbf{x}}_{\boldsymbol{\theta}}(\mathbf{z}_t; t))$, where the data $\mathbf{x}$ is replaced by a learnable denoising model $\hat{\mathbf{x}}_{\boldsymbol{\theta}}(\mathbf{z}_t; t)$. 
    }
    \label{fig: diffusion_process}
    \hfill
\end{figure}
\end{landscape}
\newpage
\section{Linear Gaussian Transitions}
\label{subsubsec: lgt}
The conditional distribution of $\mathbf{z}_t$ given a preceding latent variable $\mathbf{z}_s$, for any timestep $s < t$, is given by:
\begin{align}
    && q(\mathbf{z}_t \mid \mathbf{z}_{s}) = \mathcal{N}\left(\mathbf{z}_t;\alpha_{t|s}\mathbf{z}_s, \sigma_{t|s}^2 \mathbf{I} \right), && \mathbf{z}_t = \alpha_{t|s} \mathbf{z}_s + \sigma_{t|s}\boldsymbol{\epsilon}_t, &&
\end{align}
where $\boldsymbol{\epsilon}_t \sim \mathcal{N}\left(\boldsymbol{\epsilon}_t; 0, \mathbf{I}\right)$, thereby forming a Markov chain:
\begin{align}
    \mathbf{z}_1 \leftarrow \mathbf{z}_{(T-1)/T} \leftarrow \mathbf{z}_{(T-2)/T} \leftarrow \cdots \leftarrow \mathbf{z}_0 \leftarrow \mathbf{x}.
\end{align}
Figure~\ref{fig: mt} provides an example. In the continuous-time case where $T \to \infty$, each transition is w.r.t. an infinitesimal change in time $\mathrm{d}t$.

The transition distribution $q(\mathbf{z}_t \mid \mathbf{z}_{s})$ is useful for computing closed-form expressions for the parameters of the posterior $q(\mathbf{z}_s \mid \mathbf{z}_t, \mathbf{x})$, which defines our \textit{reverse-process}, i.e. the generative model (c.f. Section~\ref{subsubsec: qzs}). Let's focus on deriving $\alpha_{t|s}$ first. By construction, we know that each $\mathbf{z}_t$ is given by:
\begin{align}
    \mathbf{z}_t = \alpha_{t} \mathbf{x} + \sigma_{t}\boldsymbol{\epsilon}_t = \alpha_{t} \left(\frac{\mathbf{z}_s - \sigma_s \boldsymbol{\epsilon}_s}{\alpha_s} \right) + \sigma_{t}\boldsymbol{\epsilon}_t,
\end{align}
since $\mathbf{x} = (\mathbf{z}_s - \sigma_s \boldsymbol{\epsilon}_s) / \alpha_s$ for any $s<t$. The conditional mean of $q(\mathbf{z}_t \mid \mathbf{z}_{s})$ is then readily given by:
\begin{align}
    \mathbb{E} \left[ \mathbf{z}_t \mid \mathbf{z}_s \right] &= \alpha_{t} \left(\frac{\mathbf{z}_s - \sigma_s \mathbb{E} \left[\boldsymbol{\epsilon}_s\right]}{\alpha_s} \right) + \sigma_{t}\mathbb{E} \left[\boldsymbol{\epsilon}_t\right] 
    \\[5pt] &
    = \frac{\alpha_t}{\alpha_s}\mathbf{z}_s \customtag{since $\mathbb{E}[\boldsymbol{\epsilon}_t] = 0, \ \forall t$}
    \\[5pt] & 
    \eqqcolon \alpha_{t|s}\mathbf{z}_s. \label{eq: cond_alpha}
\end{align}
To compute a closed-form expression for the variance $\sigma^2_{t|s}$ of the transition distribution $q(\mathbf{z}_t \mid \mathbf{z}_{s})$, we can start by rewriting the equation for $\mathbf{z}_t$ in terms of the preceding latent $\mathbf{z}_s$ as follows:
\begin{align}
    \mathbf{z}_t &= \alpha_{t|s} \mathbf{z}_s + \sigma_{t|s}\boldsymbol{\epsilon}_{t} 
    \\[5pt] &= \frac{\alpha_{t}}{\alpha_{s}} \left( \alpha_s \mathbf{x} + \sigma_s \boldsymbol{\epsilon}_s\right) + \sigma_{t|s}\boldsymbol{\epsilon}_{t} \customtag{substitute $\alpha_{t|s}$, $\mathbf{z}_s$}
    \\[5pt] & = \alpha_t \mathbf{x} + \frac{\alpha_t}{\alpha_s} \sigma_s \boldsymbol{\epsilon}_s + \sigma_{t|s}\boldsymbol{\epsilon}_{t}    
    \\[5pt] \implies \sigma_t \boldsymbol{\epsilon}_{t} & = \frac{\alpha_t}{\alpha_s} \sigma_s \boldsymbol{\epsilon}_s + \sigma_{t|s}\boldsymbol{\epsilon}_{t}. \customtag{since $\mathbf{z}_t = \alpha_t \mathbf{x} + \sigma_t \boldsymbol{\epsilon}_t$}
\end{align}

The above implication allows us to compute the variance $\sigma_{t|s}^2$ straightforwardly. Firstly, recall that variance is invariant to changes in a location parameter, therefore: $\mathbb{V}\left[ cX \right] = c^2\mathbb{V}\left[ X \right]$ for some constant $c$ and random variable $X$. Secondly, the variance of a sum of $n$ independent random variables is simply the sum of their variances: $\mathbb{V}\left[\sum_{i=1}^n X_n\right] = \sum_{i=1}^n\mathbb{V}\left[ X_i\right]$. Using these two properties we show that:
\begin{align}
    \mathbb{V}\left[\sigma_t \boldsymbol{\epsilon}_t\right] & = \mathbb{V}\left[\frac{\alpha_t}{\alpha_s} \sigma_s \boldsymbol{\epsilon}_s + \sigma_{t|s}\boldsymbol{\epsilon}_{t}\right]
    \\[5pt] \sigma^2_t\mathbb{V}\left[ \boldsymbol{\epsilon}_t\right] & = \left(\frac{\alpha_t}{\alpha_s}\right)^2 \sigma_s^2\mathbb{V}\left[ \boldsymbol{\epsilon}_s\right] + \sigma_{t|s}^2 \mathbb{V}\left[\boldsymbol{\epsilon}_{t}\right]
    \\[5pt] \sigma^2_t & = \left(\frac{\alpha_t}{\alpha_s}\right)^2 \sigma_s^2 + \sigma_{t|s}^2
    \\[5pt] \sigma_{t|s}^2 & = \sigma^2_t - \alpha_{t|s}^2 \sigma_s^2. \label{eq: post_var}
\end{align}
\section{The Top-down Posterior}
\label{subsubsec: qzs}
Since the forward process is a Markov chain, the joint distribution of any two latent variables $\mathbf{z}_t$ and $\mathbf{z}_s$ where $t > s$ factorizes as: $q(\mathbf{z}_s, \mathbf{z}_t \mid \mathbf{x}) =
q(\mathbf{z}_t \mid \mathbf{z}_s)q(\mathbf{z}_s \mid \mathbf{x})$. Using Bayes' theorem, it is then possible to derive closed-form expressions for the parameters of the posterior distribution $q(\mathbf{z}_s \mid \mathbf{z}_{t}, \mathbf{x})$, which is itself Gaussian due to conjugacy, where $q(\mathbf{z}_s \mid \mathbf{x})$ acts as a Gaussian prior and $q(\mathbf{z}_t \mid \mathbf{z}_s)$ a Gaussian likelihood:
\begin{align}
    q(\mathbf{z}_s \mid \mathbf{z}_{t}, \mathbf{x}) &= \mathcal{N} \left(\mathbf{z}_s;\boldsymbol{\mu}_Q(\mathbf{z}_t, \mathbf{x}; s, t), \sigma^2_Q(s,t) \mathbf{I}\right), \\[5pt] \mathbf{z}_s &= \boldsymbol{\mu}_Q(\mathbf{z}_t, \mathbf{x}; s, t) + \sigma_{Q}(s,t)\boldsymbol{\epsilon}_t,
\end{align}
with $\boldsymbol{\epsilon}_t \sim \mathcal{N}\left(\boldsymbol{\epsilon}_t; 0, \mathbf{I}\right)$. In the following, we will derive closed-form expressions for the posterior parameters $\boldsymbol{\mu}_Q(\mathbf{z}_t, \mathbf{x}; s, t)$ and $\sigma^2_Q(s,t)$ in detail. For a graphical model of the posterior see Figure~\ref{fig: tdp}.

Before proceeding, we note that this posterior distribution will be instrumental in defining our generative model (i.e. the reverse process) as explained later on in Section~\ref{subsec: Discrete-time Generative Model}. Furthermore, notice that the posterior $q(\mathbf{z}_s \mid \mathbf{z}_t, \mathbf{x})$ coincides with the \textit{top-down} inference model specification of a hierarchical VAE. 

For simplicity, let $D$ denote the dimensionality of $\mathbf{z}_t$, satisfying $\mathrm{dim}(\mathbf{z}_t) = \mathrm{dim}(\mathbf{x}), \forall t$. Furthermore, recall that our covariance matrix of choice is isotropic/spherical: $\sigma^2_Q \mathbf{I}$. The posterior is then given by
\begin{align}
    &q(\mathbf{z}_s \mid \mathbf{z}_t, \mathbf{x}) 
    = \frac{q(\mathbf{z}_t \mid \mathbf{z}_s)q(\mathbf{z}_s \mid \mathbf{x})}{q(\mathbf{z}_t \mid \mathbf{x})} 
    \\[5pt] & 
    \propto q(\mathbf{z}_t \mid \mathbf{z}_s)q(\mathbf{z}_s \mid \mathbf{x})
    \\[5pt] & = \mathcal{N}\left(\mathbf{z}_t; \alpha_{t|s} \mathbf{z}_s, \sigma^2_{t|s}\mathbf{I}\right)\cdot \mathcal{N}\left(\mathbf{z}_s;\alpha_s\mathbf{x}, \sigma^2_s\mathbf{I} \right)
    \\[5pt] & = \prod_{i=1}^D \frac{1}{\sigma_{t|s}\sqrt{2\pi}}\exp\left\{-\frac{1}{2\sigma_{t|s}^2}\left(\mathbf{z}_{t,i} - \alpha_{t|s}\mathbf{z}_{s,i}\right)^2 \right\} 
    \\[5pt] & \hspace{57pt} \cdot \prod_{i=1}^D \frac{1}{\sigma_{s}\sqrt{2\pi}}\exp\left\{-\frac{1}{2\sigma^2_{s}}\left(\mathbf{z}_{s,i} - \alpha_{s}\mathbf{x}_i\right)^2 \right\} 
    \\[5pt] & \propto \prod_{i=1}^D \exp\left\{-\frac{1}{2\sigma_{t|s}^2}\left(\mathbf{z}_{t,i} - \alpha_{t|s}\mathbf{z}_{s,i}\right)^2 \right\} \cdot \prod_{i=1}^D \exp\left\{-\frac{1}{2\sigma^2_{s}}\left(\mathbf{z}_{s,i} - \alpha_{s}\mathbf{x}_i\right)^2 \right\}
    \\[5pt] & = \prod_{i=1}^D \exp\Bigg\{-\frac{1}{2\sigma^2_{t|s}}\left(\mathbf{z}_{t,i}^2 - 2\mathbf{z}_{t,i}\alpha_{t|s}\mathbf{z}_{s,i} + \alpha_{t|s}^2\mathbf{z}_{s,i}^2\right) 
    \\[5pt] & \hspace{57pt} -\frac{1}{2\sigma^2_{s}}\left(\mathbf{z}_{s,i}^2 - 2\mathbf{z}_{s,i}\alpha_{s}\mathbf{x}_i + \alpha_{s}^2\mathbf{x}_i^2\right) \Bigg\}
    \\[5pt] & = \prod_{i=1}^D \exp\Bigg\{-\frac{1}{2}\Bigg[\frac{\mathbf{z}_{t,i}^2 - 2\mathbf{z}_{t,i}\alpha_{t|s}\mathbf{z}_{s,i} + \alpha_{t|s}^2\mathbf{z}_{s,i}^2}{\sigma^2_{t|s}} 
    \\[5pt] & \hspace{57pt} + \frac{\mathbf{z}_{s,i}^2  -2\mathbf{z}_{s,i}\alpha_{s}\mathbf{x}_i + \alpha_{s}^2\mathbf{x}_i^2}{\sigma^2_{s}} \Bigg]\Bigg\}
    \\[5pt] & = \prod_{i=1}^D \exp\Bigg\{-\frac{1}{2}\Bigg[
    \mathbf{z}_{s,i}^2\left(\frac{\alpha_{t|s}^2}{\sigma_{t|s}^2} + \frac{1}{\sigma_s^2}\right) -2\mathbf{z}_{s,i}\left( \frac{\alpha_{t|s}\mathbf{z}_{t,i}}{\sigma_{t|s}^2}+\frac{\alpha_s\mathbf{x}_i}{\sigma_s^2}\right)
    \\[5pt] & \hspace{57pt} + \frac{\mathbf{z}_{t,i}^2}{\sigma_{t|s}^2} 
    + \frac{\alpha_s^2\mathbf{x}_i^2}{\sigma_{s}^2}
    \Bigg]\Bigg\}. 
    \label{eq:mm}
\end{align}

The next step is to `match the moments' from Equation~\eqref{eq:mm} with what we expect to see in a Gaussian distribution, i.e. something of the form: 
\begin{align}
\mathcal{N}\left(x;\mu, \sigma^2 \right)\propto \exp\left\{- \frac{x^2}{2\sigma^2} + \frac{\mu x}{\sigma^2} - \frac{\mu^2}{2\sigma^2}
 \right\}    .
\end{align}
This exercise yields closed-form expressions for the parameters of the posterior distribution as desired. Without loss of generality, consider the $D=1$ dimensional case for brevity.
  
Matching the first term in Eq.~\eqref{eq:mm} with $-\frac{x^2}{2\sigma^2}$ we can see that:
\begin{align}
    -\frac{\mathbf{z}_s^2}{2}\left(\frac{\alpha_{t|s}^2}{\sigma_{t|s}^2} + \frac{1}{\sigma_s^2}\right) \implies \frac{1}{\sigma^2_Q} = \frac{\alpha_{t|s}^2}{\sigma_{t|s}^2} + \frac{1}{\sigma_s^2},
\end{align}
where $\sigma^2_Q$ is the variance of the posterior $q(\mathbf{z}_s \mid \mathbf{z}_t, \mathbf{x})$. Matching the second term in Eq.~\eqref{eq:mm} with $\frac{\mu x}{\sigma^2}$ we get:
\begin{align}
    \mathbf{z}_s\left(\frac{\alpha_{t|s}\mathbf{z}_t}{\sigma_{t|s}^2} + \frac{\alpha_s\mathbf{x}}{\sigma_s^2}\right) &\implies \frac{\boldsymbol{\mu}_Q}{\sigma^2_Q} = \frac{\alpha_{t|s}\mathbf{z}_t}{\sigma_{t|s}^2} + \frac{\alpha_s\mathbf{x}}{\sigma_s^2} \\[5pt]&\implies \boldsymbol{\mu}_Q = {\sigma^2_Q} \left(\frac{\alpha_{t|s}\mathbf{z}_t}{\sigma_{t|s}^2} + \frac{\alpha_s\mathbf{x}}{\sigma_s^2}\right),
\end{align}
where $\boldsymbol{\mu}_Q$ is the mean of the posterior $q(\mathbf{z}_s \mid \mathbf{z}_t, \mathbf{x})$. 

\begin{table}[!t]
    \centering
    \begin{tabular}{lll}
        \toprule
        Distribution & Mean & Covariance \\
        \midrule
        $q(\mathbf{z}_t \mid \mathbf{x})$ (\S\ref{subsec: Gaussian Diffusion Process: Forward Time})&  $\alpha_t \mathbf{x}$ &  $\sigma_t^2\mathbf{I}$ \\[2pt]
        $q(\mathbf{z}_t \mid \mathbf{z}_s)$ (\S\ref{subsubsec: lgt}) & $\alpha_{t|s} \mathbf{z}_s$ & $\sigma_{t|s}^2\mathbf{I}$ \\[2pt]
        $q(\mathbf{z}_s \mid \mathbf{z}_t, \mathbf{x})$ (\S\ref{subsubsec: qzs}) & $\boldsymbol{\mu}_Q(\mathbf{z}_t, \mathbf{x}; s, t)$ &  $\sigma^2_Q(s,t)\mathbf{I}$ \\[2pt]
        \bottomrule
    \end{tabular}
    \\[5pt]
    \begin{tabular}{ll}
        \toprule
        Parameter & Expression \\
        \midrule
        $\alpha_{t|s}$ & $\alpha_t / \alpha_s$ \\[5pt]
        $\sigma^2_{t|s}$ & $\sigma^2_t - \alpha_{t|s}^2 \sigma_s^2$ \\[5pt]
        $\boldsymbol{\mu}_Q(\mathbf{z}_t, \mathbf{x}; s, t)$ & $\displaystyle\frac{\alpha_{t|s}\sigma_s^2}{\sigma^2_{t}}\mathbf{z}_t + \frac{\alpha_s \sigma^2_{t|s}}{\sigma_{t}^2}\mathbf{x}$ \\[5pt]
        $\sigma^2_Q(s,t)$ & $\displaystyle \frac{\sigma_{t|s}^2\sigma_s^2}{\alpha_{t|s}^2 \sigma_s^2 + \sigma_{t|s}^2}$ \\[5pt]
        \bottomrule
    \end{tabular}
    \caption{Breakdown of the distributions involved in defining a typical Gaussian diffusion (LHS), along with closed-form expressions for their respective parameters (RHS). Note that $s$ denotes a preceding timestep relative to timestep $t$, i.e. $s < t$. The top-down posterior distribution $q(\mathbf{z}_s \mid \mathbf{z}_t, \mathbf{x})$ is tractable due to Gaussian conjugacy: $q(\mathbf{z}_s \mid \mathbf{z}_t, \mathbf{x}) \propto q(\mathbf{z}_t \mid \mathbf{z}_s) q(\mathbf{z}_s \mid \mathbf{x}) $, where $q(\mathbf{z}_s \mid \mathbf{x})$ plays the role of a conjugate (Gaussian) prior and $q(\mathbf{z}_t \mid \mathbf{z}_s)$ the plays the role of a Gaussian likelihood.}
    \label{tab: dist_params}
\end{table}
The closed-form expressions for $\boldsymbol{\mu}_Q$, $\sigma^2_Q$ simplify quite significantly:
\begin{align}
    \frac{1}{\sigma^2_Q} &= \frac{\sigma^2_s}{\sigma^2_s}\cdot
    \frac{\alpha_{t|s}^2}{\sigma_{t|s}^2} + \frac{\sigma^2_{t|s}}{\sigma^2_{t|s}} \cdot \frac{1}{\sigma_s^2} \\[5pt] & = \frac{\alpha_{t|s}^2 \sigma_s^2 + \sigma_{t|s}^2}{\sigma_{t|s}^2\sigma_s^2} \implies \sigma_Q^2 = \frac{\sigma_{t|s}^2\sigma_s^2}{\alpha_{t|s}^2 \sigma_s^2 + \sigma_{t|s}^2},
\end{align}
and for the posterior mean we then have:
\begin{align}
    \boldsymbol{\mu}_Q & = 
    {\sigma^2_Q} \left(\frac{\alpha_{t|s}\mathbf{z}_t}{\sigma_{t|s}^2} + \frac{\alpha_s\mathbf{x}}{\sigma_s^2}\right)
    \\[5pt] &= \frac{\sigma_{t|s}^2\sigma_s^2}{\alpha_{t|s}^2 \sigma_s^2 + \sigma_{t|s}^2} \cdot \frac{\sigma_s^2\alpha_{t|s}\mathbf{z}_t + \sigma_{t|s}^2\alpha_s\mathbf{x}}{\sigma_{t|s}^2\sigma_s^2} 
    \\[5pt] &
    = \frac{\sigma_s^2\alpha_{t|s}\mathbf{z}_t + \sigma_{t|s}^2\alpha_s\mathbf{x}}{\alpha_{t|s}^2 \sigma_s^2 + \sigma_{t|s}^2}
    \\[5pt] &= \frac{\alpha_{t|s}\sigma_s^2}{\alpha_{t|s}^2 \sigma_s^2 + \sigma_{t|s}^2}\mathbf{z}_t + \frac{\alpha_s \sigma^2_{t|s}}{\alpha_{t|s}^2 \sigma_s^2 + \sigma_{t|s}^2}\mathbf{x}.
\end{align}
Using the fact that $\sigma_{t|s}^2 = \sigma^2_t - \alpha_{t|s}^2 \sigma_s^2$ as in Equation~\ref{eq: post_var}, we get the final expression:
\begin{align}
    \boldsymbol{\mu}_Q(\mathbf{z}_t, \mathbf{x};s,t) = \frac{\alpha_{t|s}\sigma_s^2}{\sigma^2_{t}}\mathbf{z}_t + \frac{\alpha_s \sigma^2_{t|s}}{\sigma_{t}^2}\mathbf{x}, \label{eq: post_mu}
\end{align}
revealing that the posterior mean $\boldsymbol{\mu}_Q$, equivalently denoted as $\boldsymbol{\mu}_Q(\mathbf{z}_t, \mathbf{x};s,t)$ by~\cite{kingma2021variational}, is essentially a weighted average of the conditioning set $\{\mathbf{z}_t, \mathbf{x}\}$ of the posterior distribution $q(\mathbf{z}_s \mid \mathbf{z}_t, \mathbf{x})$.

In summary, the top-down posterior distribution is given by:
\begin{align}
    q(\mathbf{z}_s \mid \mathbf{z}_{t}, \mathbf{x}) & = \mathcal{N} \left(\mathbf{z}_s;\frac{\alpha_{t|s}\sigma_s^2}{\sigma^2_{t}}\mathbf{z}_t + \frac{\alpha_s \sigma^2_{t|s}}{\sigma_{t}^2}\mathbf{x}, \frac{\sigma_{t|s}^2\sigma_s^2}{\alpha_{t|s}^2 \sigma_s^2 + \sigma_{t|s}^2} \mathbf{I}\right)
    \\[5pt] & = \mathcal{N} \left(\mathbf{z}_s;\boldsymbol{\mu}_Q(\mathbf{z}_t, \mathbf{x}; s, t), \sigma^2_Q(s,t) \mathbf{I}\right).
\end{align}
To conclude, Table~\ref{tab: dist_params} provides a concise breakdown of all the distributions involved in defining a Gaussian diffusion, along with the respective closed-form expressions of their parameters.
\section{Reverse Process: Discrete-Time Generative Model}
\label{subsec: Discrete-time Generative Model}
The generative model in diffusion models inverts the Gaussian diffusion process outlined in Section~\ref{subsec: Gaussian Diffusion Process: Forward Time}. In other words, it estimates the \textit{reverse-time} variational Markov Chain relative to a corresponding \textit{forward-time} diffusion process. An interesting aspect of VDMs is that they admit continuous-time generative models ($T \to \infty$) in a principled manner, and these correspond to the infinitely deep limit of a hierarchical VAE with a fixed encoder. We first describe the discrete-time model (finite $T$)~\citep{sohl2015deep,ho2020denoising}, as it closely relates to the material already covered, then describe the continuous-time version. 

\paragraph{Notation.} To unify the notation for both the discrete and continuous-time model versions, \cite{kingma2021variational} uniformly discretize time into $T$ segments of width $\tau = 1/T$. Each time segment corresponds to a level/step in the hierarchy of latent variables defined as follows: 
\begin{align}
    && t(i) = \frac{i}{T}, && s(i) = \frac{i-1}{T}, &&
\end{align}
where $s(i)$ precedes $t(i)$ in the timestep hierarchy, for an index $i$. For simplicity, we may sometimes use $s$ and $t$ as shorthand notation for $s(i)$ and $t(i)$ when our intentions are clear from context. 

As previously mentioned, the discrete-time generative model of a variational diffusion model is identical to the hierarchical VAE's generative model described in Section~\ref{subsec: Hierarchical VAE}. Using the new index notation defined above, we can re-express the discrete-time generative model as:
\begin{align}
    p(\mathbf{x}, \mathbf{z}_{0:1}) & = p(\mathbf{z}_1)p(\mathbf{z}_{(T-1)/T} \mid \mathbf{z}_T)
    \cdots p(\mathbf{z}_0 \mid \mathbf{z}_{1/T})p(\mathbf{x} \mid \mathbf{z}_0) \\[5pt]& = \underbrace{p(\mathbf{z}_1)}_{\mathrm{prior}} \underbrace{p(\mathbf{x} \mid \mathbf{z}_0)}_{\mathrm{likelihood}} \prod_{i=1}^T \underbrace{p(\mathbf{z}_{s(i)} \mid \mathbf{z}_{t(i)})}_{\mathrm{transitions}}.
\end{align}
This corresponds to a Markov chain:
\begin{align}
\mathbf{z}_1 \to \mathbf{z}_{(T-1)/T} \to \mathbf{z}_{(T-2)/T} \to \cdots \to \mathbf{z}_0 \to \mathbf{x},    
\end{align}
which is equivalent in principle to the hierarchical VAE's Markov chain: $\mathbf{z}_T \to \mathbf{z}_{T-1} \to \cdots \to \mathbf{z}_1 \to \mathbf{x}$, for equal $T$.

\noindent Each component of the discrete-time generative model is defined as:
\begin{enumerate}[(i)]
    \item The \textbf{prior} term can be safely set to $p(\mathbf{z}_1) = \mathcal{N}\left(0, \mathbf{I}\right)$ in a variance preserving diffusion process since -- for small enough $\mathrm{SNR}(t=1)$ -- the noisiest latent $\mathbf{z}_1$ holds almost no information about the input $\mathbf{x}$. In other words, this means that $q(\mathbf{z}_1 \mid \mathbf{x}) \approx \mathcal{N}\left(\mathbf{z}_1; 0, \mathbf{I}\right)$ by construction, and as such there exists a distribution $p(\mathbf{z}_1)$ such that $D_{\mathrm{KL}}\left(q(\mathbf{z}_1 \mid \mathbf{x}) \parallel p(\mathbf{z}_1) \right) \approx 0$.
    \item The \textbf{likelihood} term $p(\mathbf{x} \mid \mathbf{z}_0)$ factorizes over the number of elements $D$ (e.g. pixels) in $\mathbf{x}$, $\mathbf{z}_0$ as:
    \begin{align}
        p(\mathbf{x} \mid \mathbf{z}_0) = \prod_{i=1}^D p({x}^{(i)} \mid {z}_0^{(i)}),
    \end{align}
    such as a product of (potentially discretized) Gaussian distributions. This distribution could conceivably be modelled autoregressively, but there is little advantage in doing so, as $\mathbf{z}_0$ (the least noisy latent) is almost identical to $\mathbf{x}$ by construction. This means that $p(\mathbf{x} \mid \mathbf{z}_0) \approx q(\mathbf{x} \mid \mathbf{z}_0)$ for sufficiently large $\mathrm{SNR}(t=0)$. Intuitively, since $\mathbf{z}_0$ is almost equal to $\mathbf{x}$ by construction, modelling $p(\mathbf{z}_0)$ is practically equivalent to modelling $p(\mathbf{x})$, so the likelihood term $p(\mathbf{x} \mid \mathbf{z}_0)$ is typically omitted, as learning $p(\mathbf{z}_0 \mid \mathbf{z}_{1/T})$ has proven to be sufficient in practice. 
    \item The \textbf{transition} conditional distributions $p(\mathbf{z}_s \mid \mathbf{z}_t)$ are defined to be the same as the top-down posteriors $q(\mathbf{z}_s \mid \mathbf{z}_t, \mathbf{x})$ presented in Section~\ref{subsubsec: qzs}, but with the observed data $\mathbf{x}$ replaced by the output of a time-dependent \textit{denoising} model $\hat{\mathbf{x}}_{\boldsymbol{\theta}}(\mathbf{z}_t;t)$, that is:
    \begin{align}
        p(\mathbf{z}_s \mid \mathbf{z}_t) = q(\mathbf{z}_s \mid \mathbf{z}_t, \mathbf{x} = \hat{\mathbf{x}}_{\boldsymbol{\theta}}(\mathbf{z}_t;t)).    
    \end{align}
    The role of the denoising model is to predict $\mathbf{x}$ from each of its noisy versions $\mathbf{z}_t$ in turn. There are three different interpretations of this component of the generative model, as we describe next.
\end{enumerate}
\section{Generative Transitions}
\label{subsubsec: Deriving p}
The conditional distributions of the generative model are given by:
\begin{align}
    p(\mathbf{z}_s \mid \mathbf{z}_t) = 
    \mathcal{N}\left(\mathbf{z}_s; \boldsymbol{\mu}_{\boldsymbol{\theta}}(\mathbf{z}_t; s, t), \sigma_Q^2(s,t)\mathbf{I} \right)
\end{align}
where $\sigma_Q^2(s,t)$ is the posterior variance we derived in Equation~\ref{eq: post_var}, and $\boldsymbol{\mu}_{\boldsymbol{\theta}}(\mathbf{z}_t; s, t)$ is analogous to the posterior mean we derived in Equation~\ref{eq: post_mu}, that is: 
\begin{align}
    q(\mathbf{z}_s \mid \mathbf{z}_t, \mathbf{x}) = \mathcal{N} \left(\mathbf{z}_s;\boldsymbol{\mu}_Q(\mathbf{z}_t, \mathbf{x};s,t), \sigma^2_Q(s,t) \mathbf{I}\right),
\end{align}
where the posterior mean is given by
\begin{align}
    \boldsymbol{\mu}_Q(\mathbf{z}_t, \mathbf{x};s,t) = \frac{\alpha_{t|s}\sigma_s^2}{\sigma^2_{t}}\mathbf{z}_t + \frac{\alpha_s \sigma^2_{t|s}}{\sigma_{t}^2}\mathbf{x}.
\end{align}
The crucial difference between $\boldsymbol{\mu}_Q(\mathbf{z}_t, \mathbf{x};s,t)$ and $\boldsymbol{\mu}_{\boldsymbol{\theta}}(\mathbf{z}_t; s, t)$ is that, in the latter, the observed data $\mathbf{x}$ is replaced by our image prediction model $\hat{\mathbf{x}}_{\boldsymbol{\theta}}(\mathbf{z}_t;t)$ with parameters $\boldsymbol{\theta}$:
\begin{align}
    \boldsymbol{\mu}_{\boldsymbol{\theta}}(\mathbf{z}_t;s,t) = \frac{\alpha_{t|s}\sigma_s^2}{\sigma^2_{t}}\mathbf{z}_t + \frac{\alpha_s \sigma^2_{t|s}}{\sigma_{t}^2}\hat{\mathbf{x}}_{\boldsymbol{\theta}}(\mathbf{z}_t;t).
\end{align}
It is worth noting at this stage that there are multiple equivalent parameterizations beyond image-prediction $\hat{\mathbf{x}}_{\boldsymbol{\theta}}(\mathbf{z}_t;t)$, e.g. noise and velocity prediction, all of which can be reparameterized to compute the posterior mean estimate $\boldsymbol{\mu}_{\boldsymbol{\theta}}(\mathbf{z}_t;s,t)$. For consistency, we will continue to use the image-prediction version for now and provide a detailed derivation of the alternatives in Section~\ref{subsec: Understanding Diffusion Objectives}.
\section{Variational Lower Bound}
\label{subsubsec: Variational Lower Bound: Top-down HVAE}
The optimization objective of a discrete-time variational diffusion model is the ELBO in Equation~\ref{eq: hvae_elbo}, i.e. the same as a hierarchical VAE's with a \textit{top-down} inference model. For consistency, we re-express the VLB here using the discrete-time index notation: $s(i) = (i-1)/T$,  $t(i) = i/T$: $-\log p(\mathbf{x})$
\begin{align}
    &\leq -\mathbb{E}_{q(\mathbf{z}_0 \mid \mathbf{x})}\left[\log p(\mathbf{x} \mid \mathbf{z}_0) \right] + D_{\mathrm{KL}}\left( q(\mathbf{z}_1 \mid \mathbf{x}) \parallel p(\mathbf{z}_1) \right) + {\mathcal{L}_T(\mathbf{x})}
    \\[5pt] &= -\mathrm{VLB}(\mathbf{x}), \customtag{free energy}
\end{align}
where the so-called diffusion loss $\mathcal{L}_T(\mathbf{x})$ term is given by:
\begin{align}
    \mathcal{L}_T(\mathbf{x}) = \sum_{i=1}^T \mathbb{E}_{q(\mathbf{z}_{t(i)} \mid \mathbf{x})} \left[D_{\mathrm{KL}}(q(\mathbf{z}_{s(i)} \mid \mathbf{z}_{t(i)}, \mathbf{x}) \parallel p(\mathbf{z}_{s(i)} \mid \mathbf{z}_{t(i)}))\right]. 
\end{align}
The remaining terms are the familiar expected reconstruction loss and KL of the posterior from the prior. For reasons explained in detail in Section~\ref{subsec: Discrete-time Generative Model} and Appendix~\ref{app: Dealing with Edge Effects}, under a well-specified diffusion process, these terms can be safely omitted in practice as they do not provide meaningful contributions to the diffusion loss.
\subsection{Deriving the KL Divergence Terms}
\label{subsubsec: deriving dkl}
Minimizing the diffusion loss $\mathcal{L}_T(\mathbf{x})$ involves computing the (expected) KL divergence of the posterior from the prior, at each noise level. \citet{kingma2021variational} provide a relatively detailed derivation of $D_{\mathrm{KL}}(q(\mathbf{z}_{s(i)} \mid \mathbf{z}_{t(i)}, \mathbf{x}) \parallel p(\mathbf{z}_{s(i)} \mid \mathbf{z}_{t(i)}))$; we re-derive it here for completeness, whilst adding some additional instructive details to aid in understanding.

Using $s$ and $t$ as shorthand notation for $s(i)$ and $t(i)$, recall that the posterior is given by:
\begin{align}
    q(\mathbf{z}_s \mid \mathbf{z}_t, \mathbf{x}) & = \mathcal{N}\left(\mathbf{z}_s; \boldsymbol{\mu}_Q(\mathbf{z}_t, \mathbf{x};s,t), \sigma^2_Q(s,t) \mathbf{I}\right),
     \\[5pt]\boldsymbol{\mu}_Q(\mathbf{z}_t,\mathbf{x};s,t) &= \frac{\alpha_{t|s}\sigma_s^2}{\sigma^2_{t}}\mathbf{z}_t + \frac{\alpha_s \sigma^2_{t|s}}{\sigma_{t}^2}\mathbf{x},
\end{align}
and since we have defined our generative model as $p(\mathbf{z}_s \mid \mathbf{z}_t) = q(\mathbf{z}_s \mid \mathbf{z}_t, \mathbf{x} = \hat{\mathbf{x}}_{\boldsymbol{\theta}}(\mathbf{z}_t;t))$ we have
\begin{align}
    p(\mathbf{z}_s \mid \mathbf{z}_t) &= \mathcal{N}\left(\mathbf{z}_s; \boldsymbol{\mu}_{\boldsymbol{\theta}}(\mathbf{z}_t;s,t), \sigma^2_Q(s,t)\mathbf{I}\right), \\[5pt]\boldsymbol{\mu}_{\boldsymbol{\theta}}(\mathbf{z}_t;s,t) &= \frac{\alpha_{t|s}\sigma_s^2}{\sigma^2_{t}}\mathbf{z}_t + \frac{\alpha_s \sigma^2_{t|s}}{\sigma_{t}^2}\hat{\mathbf{x}}_{\boldsymbol{\theta}}(\mathbf{z}_t;t).
\end{align}

Of particular importance is the fact that the variances of both $q(\mathbf{z}_s \mid \mathbf{z}_t, \mathbf{x})$ and $p(\mathbf{z}_s \mid \mathbf{z}_t)$ are equal:
\begin{align}
    \sigma^2_Q(s,t) = \frac{\sigma_{t|s}^2\sigma_s^2}{\alpha_{t|s}^2 \sigma_s^2 + \sigma_{t|s}^2} = \frac{\sigma_{t|s}^2\sigma_s^2}{\sigma_t^2},
\end{align}
where the result in Equation~\ref{eq: post_var}, $\sigma_{t|s}^2 = \sigma^2_t - \alpha_{t|s}^2 \sigma_s^2$, simplifies the denominator. Furthermore, both distributions have identical isotropic/spherical covariances: $\sigma_Q^2(s,t)\mathbf{I}$, which we denote as $\sigma_Q^2\mathbf{I}$ for short. 

The KL divergence between $D$-dimensional Gaussian distributions is available in closed form, thus:
\begin{align}
    &D_{\mathrm{KL}}(q(\mathbf{z}_{s} \mid \mathbf{z}_{t},   \mathbf{x}) \parallel p(\mathbf{z}_{s} \mid \mathbf{z}_{t})) \\[5pt] &=
  \frac{1}{2}\Bigg[ \operatorname{Tr}\left(\frac{1}{\sigma_Q^2}\mathbf{I}\sigma_Q^2\mathbf{I}\right)  - D +
    \left(\boldsymbol{\mu}_{\boldsymbol{\theta}} - \boldsymbol{\mu}_Q\right)^\top\frac{1}{\sigma_Q^2}\mathbf{I}\left(\boldsymbol{\mu}_{\boldsymbol{\theta}} - \boldsymbol{\mu}_Q\right) 
    \\[5pt] & \qquad \ \ + \log \frac{\det\left(\sigma_Q^2\mathbf{I}\right)}{\det\left( \sigma_Q^2\mathbf{I}\right)}\Bigg]
    \\[5pt]  & =
    \frac{1}{2}\left[D - D +
    \frac{1}{\sigma_Q^2}\left(\boldsymbol{\mu}_{\boldsymbol{\theta}} - \boldsymbol{\mu}_Q\right)^\top\left(\boldsymbol{\mu}_{\boldsymbol{\theta}} - \boldsymbol{\mu}_Q\right) + 0 \right]
    \\[5pt] & = \frac{1}{2\sigma_{Q}^{2}}\sum_{i=1}^D\left(  \boldsymbol{\mu}_{Q,i} - \boldsymbol{\mu}_{\boldsymbol{\theta},i}\right)^2
    \\[5pt] & = \frac{1}{2 \sigma^2_{Q}(s,t)} \left\| \boldsymbol{\mu}_{Q}(\mathbf{z}_t, \mathbf{x}; s, t) - \boldsymbol{\mu}_{\boldsymbol{\theta}}(\mathbf{z}_t; s, t) \right\|^2_2. \label{eq: klqp}
\end{align}
It is possible to simplify the above equation quite significantly, resulting in a short expression involving the signal-to-noise ratio of the diffused data. To that end, expressing Equation~\ref{eq: klqp} in terms of the denoising model $\hat{\mathbf{x}}_{\boldsymbol{\theta}}(\mathbf{z}_t;t)$ we get:
\begin{align}
    &D_{\mathrm{KL}}(q(\mathbf{z}_{s} \mid \mathbf{z}_{t},   \mathbf{x}) \parallel p(\mathbf{z}_{s} \mid \mathbf{z}_{t})) 
    \\[5pt] &= \frac{1}{2 \sigma^2_{Q}(s,t)} \left\| \boldsymbol{\mu}_{Q}(\mathbf{z}_t, \mathbf{x}; s, t) - \boldsymbol{\mu}_{\boldsymbol{\theta}}(\mathbf{z}_t; s, t) \right\|^2_2 
    \\[5pt] & = \frac{1}{2 \sigma^2_{Q}(s,t)} \left\| \frac{\alpha_{t|s}\sigma_s^2}{\sigma^2_{t}}\mathbf{z}_t + \frac{\alpha_s \sigma^2_{t|s}}{\sigma_{t}^2}\mathbf{x} - \left( \frac{\alpha_{t|s}\sigma_s^2}{\sigma^2_{t}}\mathbf{z}_t + \frac{\alpha_s \sigma^2_{t|s}}{\sigma_{t}^2}\hat{\mathbf{x}}_{\boldsymbol{\theta}}(\mathbf{z}_t;t) \right)\right\|^2_2
    \\[5pt] & = \frac{1}{2 \sigma^2_{Q}(s,t)} \left\| \frac{\alpha_s \sigma^2_{t|s}}{\sigma_{t}^2}\mathbf{x} - \frac{\alpha_s \sigma^2_{t|s}}{\sigma_{t}^2}\hat{\mathbf{x}}_{\boldsymbol{\theta}}(\mathbf{z}_t;t) \right\|^2_2
    \\[5pt] & = \frac{1}{2 \sigma^2_{Q}(s,t)} \left(\frac{\alpha_s \sigma^2_{t|s}}{\sigma_{t}^2}\right)^2 \left\| \mathbf{x} - \hat{\mathbf{x}}_{\boldsymbol{\theta}}(\mathbf{z}_t;t) \right\|^2_2  
    \\[5pt] & = \frac{\sigma_t^2}{2 \sigma^2_{t|s}\sigma^2_{s}} \frac{\alpha_s^2 \sigma^4_{t|s}}{\sigma_{t}^4} \left\| \mathbf{x} - \hat{\mathbf{x}}_{\boldsymbol{\theta}}(\mathbf{z}_t;t) \right\|^2_2 
    \customtag{recall $\sigma_Q^2(s,t)= (\sigma_{t|s}^2\sigma_s^2)/ \sigma_t^2$}
    \\[5pt] & = \frac{1}{2 \sigma^2_{s}} \frac{\alpha_s^2 \sigma^2_{t|s}}{\sigma_{t}^2} \left\| \mathbf{x} - \hat{\mathbf{x}}_{\boldsymbol{\theta}}(\mathbf{z}_t;t) \right\|^2_2
    \customtag{exponents cancel}
    \\[5pt] & = \frac{1}{2 \sigma^2_{s}} \frac{\alpha_s^2 (\sigma^2_{t} - \alpha^2_{t|s}\sigma_s^2)}{\sigma_{t}^2} \left\| \mathbf{x} - \hat{\mathbf{x}}_{\boldsymbol{\theta}}(\mathbf{z}_t;t) \right\|^2_2 
    \customtag{recall $\sigma_{t|s}^2 = \sigma_{t}^2 - \alpha^2_{t|s}\sigma_s^2$}
    \\[5pt] & = \frac{1}{2} \frac{\sigma^{-2}_{s}\left(\alpha_s^2 \sigma^2_{t} - \alpha^2_{s}\frac{\alpha^2_{t}}{\alpha^2_{s}}\sigma_s^2\right)}{\sigma_{t}^2} \left\| \mathbf{x} - \hat{\mathbf{x}}_{\boldsymbol{\theta}}(\mathbf{z}_t;t) \right\|^2_2 
    \\[5pt] & = \frac{1}{2} \frac{\alpha_s^2 \sigma^2_{t}\sigma^{-2}_{s} - \alpha^2_{t}}{\sigma_{t}^2} \left\| \mathbf{x} - \hat{\mathbf{x}}_{\boldsymbol{\theta}}(\mathbf{z}_t;t) \right\|^2_2 
    \\[5pt] & = \frac{1}{2} \left(\frac{\alpha_s^2 \sigma^2_{t}}{\sigma_{s}^2}\frac{1}{\sigma_t^2} - \frac{\alpha^2_{t}}{\sigma_{t}^2} \right) \left\| \mathbf{x} - \hat{\mathbf{x}}_{\boldsymbol{\theta}}(\mathbf{z}_t;t) \right\|^2_2 
    \\[5pt] & = \frac{1}{2} \left(\frac{\alpha_s^2}{\sigma_{s}^2} - \frac{\alpha^2_{t}}{\sigma_{t}^2}\right)\left\| \mathbf{x} - \hat{\mathbf{x}}_{\boldsymbol{\theta}}(\mathbf{z}_t;t) \right\|^2_2 
    \\[5pt] & = \frac{1}{2} \left(\mathrm{SNR}(s) - \mathrm{SNR}(t)\right)\left\| \mathbf{x} - \hat{\mathbf{x}}_{\boldsymbol{\theta}}(\mathbf{z}_t;t) \right\|^2_2. \label{eq: kl_denoising}
\end{align}
In words, the final expression shows that the diffusion loss, at timestep $t$, consists of a squared error term involving the data $\mathbf{x}$ and the model $\hat{\mathbf{x}}_{\boldsymbol{\theta}}(\mathbf{z}_t;t)$, weighted by a difference in signal-to-noise ratio at $s$ and $t$.

\section{Estimator of the Discrete-Time Diffusion Loss}
For model training, we can compute the diffusion loss $\mathcal{L}_T(\mathbf{x})$ with an unbiased Monte Carlo estimator by:
\begin{enumerate}[(i)]
    \item Using the \textit{reparameterisation gradient estimator}~\citep{kingma2013auto,rezende2014stochastic} to sample $\mathbf{z}_t \sim q(\mathbf{z}_t \mid \mathbf{x})$:
    \begin{align}
        &&\mathbf{z}_t = \alpha_t \mathbf{x} + \sigma_t \boldsymbol{\epsilon} \coloneqq g_{\alpha_t,\sigma_t}(\boldsymbol{\epsilon}, \mathbf{x}),
        &&\boldsymbol{\epsilon} \sim p(\boldsymbol{\epsilon}) = \mathcal{N}(0, \mathbf{I}).&&
    \end{align}
    \item Avoid computing all $T$ loss terms by selecting a single timestep, sampled uniformly at random from $i \sim U\{1,T\}$, at each iteration.
\end{enumerate}
Under the above setup, the estimator of the diffusion loss is given by:
\begin{align}
    &\mathcal{L}_T(\mathbf{x}) = \sum_{i=1}^T \mathbb{E}_{q(\mathbf{z}_{t(i)} \mid \mathbf{x})} \left[D_{\mathrm{KL}}(q(\mathbf{z}_{s(i)} \mid \mathbf{z}_{t(i)}, \mathbf{x}) \parallel p(\mathbf{z}_{s(i)} \mid \mathbf{z}_{t(i)}))\right]
    \\[5pt] & = \sum_{i=1}^T \mathbb{E}_{q(\mathbf{z}_{t} \mid \mathbf{x})} \left[D_{\mathrm{KL}}(q(\mathbf{z}_{s} \mid \mathbf{z}_{t}, \mathbf{x}) \parallel p(\mathbf{z}_{s} \mid \mathbf{z}_{t}))\right] 
    \customtag{using shorthand notation $s$, $t$}
    \\[5pt] & = \sum_{i=1}^T\int \left(
    \frac{1}{2}\left(\mathrm{SNR}(s) - \mathrm{SNR}(t)\right)\left\| \mathbf{x} - \hat{\mathbf{x}}_{\boldsymbol{\theta}}\left( \mathbf{z}_t;t\right) \right\|^2_2 \right)    
    q(\mathbf{z}_t \mid \mathbf{x})\mathop{\mathrm{d}\mathbf{z}_t} 
    \customtag{from Equation~\ref{eq: kl_denoising}}
    \\[5pt] & = \frac{1}{2}\int  \left(
    \sum_{i=1}^T\left(\mathrm{SNR}(s) - \mathrm{SNR}(t)\right)\left\| \mathbf{x} - \hat{\mathbf{x}}_{\boldsymbol{\theta}}\left( g_{\alpha_t, \sigma_t}(\boldsymbol{\epsilon},\mathbf{x});t\right) \right\|^2_2 \right)    
    p(\boldsymbol{\epsilon}) \mathop{\mathrm{d}\boldsymbol{\epsilon}} 
    \customtag{as $\mathbf{z}_t = \alpha_t \mathbf{x} + \sigma_t \boldsymbol{\epsilon}$}
    \\[5pt] & = \frac{1}{2}\mathbb{E}_{\boldsymbol{\epsilon} \sim \mathcal{N}(0,\mathbf{I})}
    \left[T\cdot\mathbb{E}_{i \sim U{\{1,T\}}}\left[\left(\mathrm{SNR}(s) - \mathrm{SNR}(t)\right)\left\| \mathbf{x} - \hat{\mathbf{x}}_{\boldsymbol{\theta}}\left( \mathbf{z}_t;t\right) \right\|^2_2 \right]\right] 
    \customtag{Monte Carlo estimate} \customlabel{eq: mc_estm}
    \\[5pt] & = \frac{T}{2}\mathbb{E}_{\boldsymbol{\epsilon} \sim \mathcal{N}(0,\mathbf{I}),i \sim U{\{1,T\}}}\left[\left(\mathrm{SNR}(s) - \mathrm{SNR}(t)\right)\left\| \mathbf{x} - \hat{\mathbf{x}}_{\boldsymbol{\theta}}\left( \mathbf{z}_t;t\right) \right\|^2_2 \right]. \label{eq: final_mc}
\end{align}
For clarity, above we used Monte Carlo estimation and a basic identity to arrive at Equation~\ref{eq: mc_estm}:
\begin{align}
    \mathbb{E}_q\left[f(x)\right] \approx \frac{1}{T}\sum_{i=1}^T f(x_i) \implies T \cdot \mathbb{E}_q\left[f(x)\right] \approx \sum_{i=1}^T f(x_i),
\end{align}
where $x_i \sim q$ are random samples from a distribution $q$, which is representative of $U\{1,T\}$ in our case.

Equation~\ref{eq: final_mc} can be rewritten in terms of the more commonly used noise-prediction model $\hat{\boldsymbol{\epsilon}}_{\boldsymbol{\theta}}(\mathbf{z}_t;t)$~\citep{ho2020denoising} as follows:
\begin{align}
    &\mathcal{L}_T(\mathbf{x})
    = \frac{T}{2}\mathbb{E}_{\boldsymbol{\epsilon} \sim \mathcal{N}(0,\mathbf{I}),i \sim U{\{1,T\}}}\Bigg[\left(\mathrm{SNR}(s) - \mathrm{SNR}(t)\right) \nonumber
    \\[5pt] & \qquad\qquad\qquad\qquad\qquad\qquad\cdot \bigg\| \frac{\mathbf{z}_t - \sigma_t \boldsymbol{\epsilon}}{\alpha_t} - \frac{\mathbf{z}_t - \sigma_t \hat{\boldsymbol{\epsilon}}_{\boldsymbol{\theta}}(\mathbf{z}_t;t)}{\alpha_t} \bigg\|^2_2 \Bigg] \customtag{since $\mathbf{x} = (\mathbf{z}_t - \sigma_t \boldsymbol{\epsilon}_t) / \alpha_t$}
    \\[5pt] & = \frac{T}{2}\mathbb{E}_{\boldsymbol{\epsilon} \sim \mathcal{N}(0,\mathbf{I}),i \sim U{\{1,T\}}}\left[\left(\mathrm{SNR}(s) - \mathrm{SNR}(t)\right)\left\| \frac{\sigma_t}{\alpha_t}\left(\hat{\boldsymbol{\epsilon}}_{\boldsymbol{\theta}}(\mathbf{z}_t;t) - \boldsymbol{\epsilon} \right)\right\|^2_2 \right]
    \\[5pt] & = \frac{T}{2}\mathbb{E}_{\boldsymbol{\epsilon} \sim \mathcal{N}(0,\mathbf{I}),i \sim U{\{1,T\}}}\left[\frac{\sigma_t^2}{\alpha_t^2}\left(\mathrm{SNR}(s) - \mathrm{SNR}(t)\right)\left\| \boldsymbol{\epsilon} - \hat{\boldsymbol{\epsilon}}_{\boldsymbol{\theta}}(\mathbf{z}_t;t) \right\|^2_2 \right]
    \\[5pt] & = \frac{T}{2}\mathbb{E}_{\boldsymbol{\epsilon} \sim \mathcal{N}(0,\mathbf{I}),i \sim U{\{1,T\}}}\Big[\mathrm{SNR}(t)^{-1}\left(\mathrm{SNR}(s) - \mathrm{SNR}(t)\right) \nonumber
    \\[5pt] & \qquad\qquad\qquad\qquad\qquad\qquad\cdot \left\| \boldsymbol{\epsilon} - \hat{\boldsymbol{\epsilon}}_{\boldsymbol{\theta}}(\mathbf{z}_t;t) \right\|^2_2 \Big]
    \\[5pt] & = \frac{T}{2}\mathbb{E}_{\boldsymbol{\epsilon} \sim \mathcal{N}(0,\mathbf{I}),i \sim U{\{1,T\}}}\left[
    \left(\frac{\mathrm{SNR}(s)}{\mathrm{SNR}(t)} - 1\right) \left\| \boldsymbol{\epsilon} - \hat{\boldsymbol{\epsilon}}_{\boldsymbol{\theta}}(\mathbf{z}_t;t) \right\|^2_2 \right].
\end{align}
For more details on the noise prediction parameterization please refer to Section~\ref{subsec: Understanding Diffusion Objectives}. The above estimator of the loss can be made more stable in practice by re-expressing the constant term inside the expectation using numerically stable primitives. For details, please refer to Appendix~\ref{app: Numerically stable Estimator}.
\newpage
\subsection{Ancestral Sampling} 
Once trained, to randomly sample from our generative diffusion model:
\begin{align}
    p(\mathbf{x} \mid \mathbf{z}_0)\prod_{i=1}^T p(\mathbf{z}_{s(i)} \mid \mathbf{z}_{t(i)}),    
\end{align}
we can perform what's known as \textit{ancestral sampling}. That is, we start with noise $\mathbf{z}_1 \sim \mathcal{N}\left(0, \mathbf{I}\right)$ and follow the estimated reverse Markov Chain: 
\begin{align}
    \mathbf{z}_1 \to \mathbf{z}_{(T-1)/T} \to \mathbf{z}_{(T-2)/T} \to \cdots \to \mathbf{z}_0 \to \mathbf{x}.
\end{align}
Since the forward process transitions are Markovian and linear Gaussian, the top-down posterior is tractable due to Gaussian conjugacy. Furthermore, our generative model is defined to be equal to the top-down posterior, that is, for time $s<t$: 
\begin{align}
    p(\mathbf{z}_s \mid \mathbf{z}_t) = q(\mathbf{z}_s \mid \mathbf{z}_t, \mathbf{x} = \hat{\mathbf{x}}_{\boldsymbol{\theta}}(\mathbf{z}_t;t)),    
\end{align}
with a denoising model $\hat{\mathbf{x}}_{\boldsymbol{\theta}}(\mathbf{z}_t;t)$ in place of $\mathbf{x}$. Thus, we can use our estimate of the posterior mean $\boldsymbol{\mu}_{\boldsymbol{\theta}}(\mathbf{z}_t;s,t)$ to sample from $q$ in reverse order, according to the following update rule, for any $s < t$:
\begin{align}
    \mathbf{z}_s & = \boldsymbol{\mu}_{\boldsymbol{\theta}}(\mathbf{z}_t;s,t) + \sigma_{Q}(s,t) \boldsymbol{\epsilon}
    \\[5pt] &= \frac{\alpha_s}{\alpha_t} \left( \mathbf{z}_t - \sigma_t c \hat{\boldsymbol{\epsilon}}_{\boldsymbol{\theta}}(\mathbf{z}_t;t)\right) + \sqrt{1-\alpha_s^2 c} \boldsymbol{\epsilon},
\end{align}
where $c = -\mathrm{expm1}\left(\gamma_{\boldsymbol{\eta}}(s) -\gamma_{\boldsymbol{\eta}}(t)\right)$, $\boldsymbol{\epsilon} \sim \mathcal{N}\left(0, \mathbf{I}\right)$, and we used the fact that $\sigma_s = \sqrt{1-\alpha^2_s}$ by definition in a variance-preserving diffusion process. 
For details on related derivations please see Appendix~\ref{app: Simplified Expressions for the Generative Transitions}. 
\section{Reverse Process: Continuous-Time Generative Model}
\label{subsec: Continuous-time Generative Model}
A continuous-time variational diffusion model ($T \to \infty$) corresponds to the infinitely deep limit of a hierarchical VAE, when the diffusion process (noise schedule) is learned rather than fixed. As previously alluded to, the extension of diffusion models to continuous-time has been proven to be advantageous by various authors~\citep{song2021scorebased,kingma2021variational,huang2021variational,vahdat2021score}.

In this section, we explain why using a continuous-time VLB is strictly preferable over a discrete-time version and provide detailed derivations of its estimator in terms of a denoising and noise-prediction model. Note that, due to the shared notation between discrete and continuous-time models introduced in Section~\ref{subsec: Discrete-time Generative Model}, the various derivations and results therein (e.g. for $p(\mathbf{z}_s \mid \mathbf{z}_t)$) are equally applicable for the continuous-time version presented in this section.
\section{On Infinite Depth}
\label{subsubsec: on infinite depth}
\cite{kingma2021variational} showed that doubling the number of timesteps $T$ always improves the diffusion loss, which suggests we should optimize a continuous-time VLB, with $T \rightarrow \infty$. This finding is straightforward to verify; we start by recalling that the discrete-time diffusion loss using $T$ steps is given by: $\mathcal{L}_T(\mathbf{x})$
\begin{align}
    = \frac{1}{2}\mathbb{E}_{\boldsymbol{\epsilon} \sim \mathcal{N}(0,\mathbf{I})}\left[ \sum_{i=1}^T\left( \mathrm{SNR}(s(i)) - \mathrm{SNR}(t(i))\right)
    \left\| \mathbf{x} - \hat{\mathbf{x}}_{\boldsymbol{\theta}}(\mathbf{z}_{t(i)};t(i)) \right\|^2_2 \right], \label{eq: diff_lz}
\end{align}
where $s(i) = (i - 1) / T$ and $t(i) = i / T$. To double the number of timesteps $T$, we can introduce a new symbol $t'(i)$ to represent an interpolation between $s(i)$ and $t(i)$, defined as:
\begin{align}
    t'(i) = \frac{s(i) + t(i)}{2} = \frac{1}{2} \left(\frac{i-1}{T} + \frac{i}{T}\right) = \frac{i-0.5}{T} = t(i) - \frac{0.5}{T}.
\end{align}
Using shorthand notation $s$, $t$ and $t'$ for $s(i)$, $t(i)$ and $t'(i)$; the diffusion loss with $T$ timesteps can be written equivalently to Equation~\ref{eq: diff_lz} as:
\begin{align}
    \mathcal{L}_T(\mathbf{x}) = \frac{1}{2}\mathbb{E}_{\boldsymbol{\epsilon} \sim \mathcal{N}(0,\mathbf{I})}\Bigg[ &\sum_{i=1}^T\big( \mathrm{SNR}(s) - \mathrm{SNR}(t') \nonumber
    \\[5pt] & + \mathrm{SNR}(t') - \mathrm{SNR}(t)\big)
    \left\| \mathbf{x} - \hat{\mathbf{x}}_{\boldsymbol{\theta}}(\mathbf{z}_{t};t) \right\|^2_2 \Bigg],
\end{align}
whereas the new diffusion loss with $2T$ timesteps is given by:
\begin{align}
    &\mathcal{L}_{2T}(\mathbf{x}) = \frac{1}{2}\mathbb{E}_{\boldsymbol{\epsilon} \sim \mathcal{N}(0,\mathbf{I})}\Bigg[ \sum_{i=1}^T \left(\mathrm{SNR}(s) - \mathrm{SNR}(t')\right) \nonumber
    \\[5pt] & 
    \cdot\left\| \mathbf{x} - \hat{\mathbf{x}}_{\boldsymbol{\theta}}(\mathbf{z}_{t'};t') \right\|^2_2 + \left( \mathrm{SNR}(t') - \mathrm{SNR}(t)\right)
    \left\| \mathbf{x} - \hat{\mathbf{x}}_{\boldsymbol{\theta}}(\mathbf{z}_{t};t) \right\|^2_2 
    \Bigg].
\end{align}
If we then subtract the two losses and cancel out common terms we get the following:
\begin{align}
    \mathcal{L}_{2T}(\mathbf{x}) &- \mathcal{L}_{T}(\mathbf{x}) = \frac{1}{2}\mathbb{E}_{\boldsymbol{\epsilon} \sim \mathcal{N}(0,\mathbf{I})}\Bigg[ \nonumber
    \\[0pt] &\sum_{i=1}^{T} \Big\{\mathrm{SNR}(s) \left\| \mathbf{x} - 
    \hat{\mathbf{x}}_{\boldsymbol{\theta}}(\mathbf{z}_{t'};t') \right\|^2_2 - \mathrm{SNR}(t') \left\| \mathbf{x} - \hat{\mathbf{x}}_{\boldsymbol{\theta}}(\mathbf{z}_{t'};t') \right\|^2_2 \nonumber
    \\[5pt] & + \cancel{\mathrm{SNR}(t') \left\| \mathbf{x} - \hat{\mathbf{x}}_{\boldsymbol{\theta}}(\mathbf{z}_{t};t) \right\|^2_2} - \cancel{\mathrm{SNR}(t) \left\| \mathbf{x} - \hat{\mathbf{x}}_{\boldsymbol{\theta}}(\mathbf{z}_{t};t) \right\|^2_2 \Big\}} \nonumber
    \\[5pt] & \hspace{-20pt} - \Bigg( 
    \sum_{i=1}^{T} \mathrm{SNR}(s) \left\| \mathbf{x} - \hat{\mathbf{x}}_{\boldsymbol{\theta}}(\mathbf{z}_{t};t) \right\|^2_2 - \mathrm{SNR}(t') \left\| \mathbf{x} - \hat{\mathbf{x}}_{\boldsymbol{\theta}}(\mathbf{z}_{t};t) \right\|^2_2 \nonumber
    \\[5pt] & + \cancel{\mathrm{SNR}(t') \left\| \mathbf{x} - \hat{\mathbf{x}}_{\boldsymbol{\theta}}(\mathbf{z}_{t};t) \right\|^2_2} - \cancel{\mathrm{SNR}(t) \left\| \mathbf{x} - \hat{\mathbf{x}}_{\boldsymbol{\theta}}(\mathbf{z}_{t};t) \right\|^2_2}
    \Bigg) \Bigg]
    \\[0pt] & \hspace{-30pt} = \frac{1}{2}\mathbb{E}_{\boldsymbol{\epsilon} \sim \mathcal{N}(0,\mathbf{I})}\Bigg[
    \sum_{i=1}^T \left( \mathrm{SNR}(s) - \mathrm{SNR}(t') \right) \nonumber
    \\[5pt] & \qquad \qquad \cdot\left(\left\| \mathbf{x} - \hat{\mathbf{x}}_{\boldsymbol{\theta}}(\mathbf{z}_{t'};t') \right\|^2_2 - \left\| \mathbf{x} - \hat{\mathbf{x}}_{\boldsymbol{\theta}}(\mathbf{z}_{t};t) \right\|^2_2 \right) \Bigg]. \label{eq: 2t_loss}
\end{align}
We can use Equation~\ref{eq: 2t_loss} to justify optimizing a continuous-time objective. Since $t' < t$, the prediction error term with $\mathbf{z}_{t'}$ will be lower than the one with $\mathbf{z}_{t}$, as $\mathbf{z}_{t'}$ is a less noisy version of $\mathbf{x}$ from earlier on in the diffusion process. In other words, it is always easier to predict $\mathbf{x}$ from $\mathbf{z}_{t'}$ than from $\mathbf{z}_{t}$, given an adequately trained model. Formally, doubling the number of timesteps $T$ always improves the VLB:
\begin{align}
    \mathcal{L}_{2T}(\mathbf{x}) - \mathcal{L}_{T}(\mathbf{x}) < 0 \implies \mathrm{VLB}_{2T}(\mathbf{x}) > \mathrm{VLB}_{T}(\mathbf{x}), \ \forall T \in \mathbb{N}^{+}.
\end{align}
Thus it is strictly advantageous to optimize a continuous-time VLB, where $T \rightarrow \infty$ and time $t$ is treated as continuous rather than discrete.
\section{Estimator of the Continuous-Time Diffusion Loss}
\label{subsubsec: Monte Carlo Estimator of linfty}
To arrive at an unbiased Monte Carlo estimator of the continuous-time diffusion loss $\mathcal{L}_\infty(\mathbf{x})$, we can first take the discrete-time version and substitute in the time segment width $\tau = 1/T$ to reveal:  
\begin{align}
    &\mathcal{L}_T(\mathbf{x}) =\nonumber
    \\[5pt] & = \frac{T}{2}\mathbb{E}_{\boldsymbol{\epsilon} \sim \mathcal{N}(0,\mathbf{I}),i \sim U{\{1,T\}}}\left[\left(\mathrm{SNR}(s) - \mathrm{SNR}(t)\right)\left\| \mathbf{x} - \hat{\mathbf{x}}_{\boldsymbol{\theta}}( \mathbf{z}_t;t) \right\|^2_2 \right]
    \\[5pt] & = \frac{1}{2}\mathbb{E}_{\boldsymbol{\epsilon} \sim \mathcal{N}(0,\mathbf{I}),i \sim U{\{1,T\}}}\bigg[
    \\[2pt] & \quad \qquad 
    T \left(\mathrm{SNR}\left(t - \frac{1}{T}\right) - \mathrm{SNR}(t)\right)\left\| \mathbf{x} - \hat{\mathbf{x}}_{\boldsymbol{\theta}}( \mathbf{z}_t;t) \right\|^2_2 \bigg] 
    \customtag{since $s = (i - 1)/T$}
    \\[5pt] & = \frac{1} {2}\mathbb{E}_{\boldsymbol{\epsilon} \sim \mathcal{N}(0,\mathbf{I}),i \sim U{\{1,T\}}}\left[\frac{\mathrm{SNR}(t - \tau) - \mathrm{SNR}(t)}{\tau}\left\| \mathbf{x} - \hat{\mathbf{x}}_{\boldsymbol{\theta}}( \mathbf{z}_t;t) \right\|^2_2 \right], 
    \customtag{substitute $\tau = 1/T$} 
    \customlabel{eq: bckwd}
\end{align}
again using the shorthand notation $s$ and $t$ for $s(i) = (i-1)/T$ and $t(i) = i/T$, respectively. 

The constant inside the expectation in Equation~\ref{eq: bckwd} is readily recognized as the (negative) \textit{backward difference} numerical approximation to the derivative of $\mathrm{SNR}(t)$ w.r.t $t$, since:
\begin{align}
    \dv{\mathop{\mathrm{SNR}(t)}}{t} & = \lim_{\tau \to 0} \frac{\mathrm{SNR}(t + \tau) - \mathrm{SNR}(t)}{\tau} \customtag{forward diff.}
    \\[5pt] & = \lim_{\tau \to 0} \frac{\mathrm{SNR}(t) - \mathrm{SNR}(t - \tau)}{\tau}, \customtag{backward diff.} 
\end{align}
and therefore
\begin{align}
     \lim_{\tau \to 0} \frac{\mathrm{SNR}(t - \tau) - \mathrm{SNR}(t)}{\tau} & = \lim_{\tau \to 0} -\frac{\mathrm{SNR}(t) - \mathrm{SNR}(t - \tau)}{\tau} \\[5pt] & = -\dv{\mathop{\mathrm{SNR}(t)}}{t}.
\end{align}
\newpage 
Thus taking the limit as $T \to \infty$ of the discrete-time diffusion loss:
\begin{align}
    &\mathcal{L}_\infty(\mathbf{x}) = \lim_{T \to \infty}\frac{1} {2}\mathbb{E}_{\boldsymbol{\epsilon} \sim \mathcal{N}(0,\mathbf{I})}\left[\sum_{i=1}^T\left(\mathrm{SNR}(s) - \mathrm{SNR}(t) \right)\left\| \mathbf{x} - \hat{\mathbf{x}}_{\boldsymbol{\theta}}( \mathbf{z}_{t};t) \right\|^2_2 \right]
    \\[5pt] & = \lim_{T \to \infty}\frac{1} {2}\mathbb{E}_{\boldsymbol{\epsilon} \sim \mathcal{N}(0,\mathbf{I}), i \sim U\{1,T\}}\bigg[
    \\[2pt] 
    & \qquad\qquad\qquad\qquad \frac{\mathrm{SNR}(t - \tau) - \mathrm{SNR}(t)}{\tau}\left\| \mathbf{x} - \hat{\mathbf{x}}_{\boldsymbol{\theta}}( \mathbf{z}_{t};t) \right\|^2_2 \bigg]
    \\[5pt] & = \frac{1} {2}\mathbb{E}_{\boldsymbol{\epsilon} \sim \mathcal{N}(0,\mathbf{I})}\left[\int_{0}^1-\dv{\mathop{\mathrm{SNR}(t)}}{t}\left\| \mathbf{x} - \hat{\mathbf{x}}_{\boldsymbol{\theta}}( \mathbf{z}_t;t) \right\|^2_2 \mathop{\mathrm{d} t} \right] \label{eq: t_int}
    \\[5pt] & = -\frac{1} {2}\mathbb{E}_{\boldsymbol{\epsilon} \sim \mathcal{N}(0,\mathbf{I}),t \sim \mathcal{U}(0,1)}\left[\mathrm{SNR}'(t)\left\| \mathbf{x} - \hat{\mathbf{x}}_{\boldsymbol{\theta}}( \mathbf{z}_t;t) \right\|^2_2 \right]. \label{eq: dv_loss}
\end{align}
We can express the above in terms of the noise-prediction model $\hat{\boldsymbol{\epsilon}}_{\boldsymbol{\theta}}(\mathbf{z}_t; t)$ as follows:
\begin{align}
    &\mathrm{SNR}'(t)\left\| \mathbf{x} - \hat{\mathbf{x}}_{\boldsymbol{\theta}}( \mathbf{z}_t;t) \right\|^2_2 
    \\[5pt] &= \mathrm{SNR}'(t)\left\| \frac{\mathbf{z}_t - \sigma_t \boldsymbol{\epsilon}}{\alpha_t} - \frac{\mathbf{z}_t - \sigma_t \hat{\boldsymbol{\epsilon}}_{\boldsymbol{\theta}}(\mathbf{z}_t;t)}{\alpha_t} \right\|^2_2
    \\[5pt] & = \mathrm{SNR}'(t) \left\| \frac{\sigma_t}{\alpha_t} \left(\boldsymbol{\epsilon} - \hat{\boldsymbol{\epsilon}}_{\boldsymbol{\theta}}(\mathbf{z}_t;t)\right) \right\|^2_2 \customtag{cancel $\mathbf{z}_t$ terms and factor}
    \\[5pt] & = \mathrm{SNR}'(t) \cdot \frac{\sigma_t^2}{\alpha_t^2}\left\| \boldsymbol{\epsilon} - \hat{\boldsymbol{\epsilon}}_{\boldsymbol{\theta}}(\mathbf{z}_t;t) \right\|^2_2
    \\[5pt] & = \mathrm{SNR}'(t) \cdot \mathrm{SNR}(t)^{-1} \left\| \boldsymbol{\epsilon} - \hat{\boldsymbol{\epsilon}}_{\boldsymbol{\theta}}(\mathbf{z}_t;t) \right\|^2_2 \customtag{$\mathrm{SNR}(t) = \alpha_t^2 / \sigma_t^2$}
    \\[5pt] & = \mathrm{SNR}(t)^{-1} \cdot \dv{t}e^{-\gamma_{\boldsymbol{\eta}}(t)} \left\| \boldsymbol{\epsilon} - \hat{\boldsymbol{\epsilon}}_{\boldsymbol{\theta}}(\mathbf{z}_t;t) \right\|^2_2 
    \\[5pt] & = \mathrm{SNR}(t)^{-1} \cdot e^{-\gamma_{\boldsymbol{\eta}}(t)} \cdot -\dv{t}\gamma_{\boldsymbol{\eta}}(t) \left\| \boldsymbol{\epsilon} - \hat{\boldsymbol{\epsilon}}_{\boldsymbol{\theta}}(\mathbf{z}_t;t) \right\|^2_2 \customtag{chain rule}
    \\[5pt] & = \frac{1}{e^{-\gamma_{\boldsymbol{\eta}}(t)}} \cdot e^{-\gamma_{\boldsymbol{\eta}}(t)} \cdot -\dv{t}\gamma_{\boldsymbol{\eta}}(t) \left\| \boldsymbol{\epsilon} - \hat{\boldsymbol{\epsilon}}_{\boldsymbol{\theta}}(\mathbf{z}_t;t) \right\|^2_2 \customtag{$\mathrm{SNR}(t) = e^{-\gamma_{\boldsymbol{\eta}}(t)}$}
    \\[5pt] & = -\gamma'_{\boldsymbol{\eta}}(t) \left\| \boldsymbol{\epsilon} - \hat{\boldsymbol{\epsilon}}_{\boldsymbol{\theta}}(\mathbf{z}_t;t) \right\|^2_2, \label{eq: negamma}
\end{align}
where the simplified form of $\mathrm{SNR}(t) = \exp(-\gamma_{\boldsymbol{\eta}}(t))$ derived in Equation~\ref{eq: snrt} was used to arrive at the final result. Plugging the final expression back into the expected loss in Equation~\ref{eq: dv_loss} we get
\begin{align}
    &\mathcal{L}_\infty(\mathbf{x}) = \frac{1} {2}\mathbb{E}_{\boldsymbol{\epsilon} \sim \mathcal{N}(0,\mathbf{I}),t \sim \mathcal{U}(0,1)}\left[\gamma'_{\boldsymbol{\eta}}(t) \left\| \boldsymbol{\epsilon} - \hat{\boldsymbol{\epsilon}}_{\boldsymbol{\theta}}(\mathbf{z}_t;t) \right\|^2_2 \right] \label{eq: ct_diff_loss}
    \\[5pt] & = \mathbb{E}_{q(\mathbf{z}_0 \mid \mathbf{x})}\left[\log p(\mathbf{x} \mid \mathbf{z}_0) \right] - D_{\mathrm{KL}}\left( q(\mathbf{z}_1 \mid \mathbf{x}) \parallel p(\mathbf{z}_1) \right) - \mathrm{VLB}(\mathbf{x})
    \\[5pt] & = - \mathrm{VLB}(\mathbf{x}) + c, 
\end{align}
where $c \approx 0 $ is constant with respect to the model parameters of $\hat{\boldsymbol{\epsilon}}_{\boldsymbol{\theta}}(\mathbf{z}_t;t)$. For more details on related aspects please refer to Appendix~\ref{app: Dealing with Edge Effects}.

In the next section, we explain why the continuous-time VLB derive in this section is invariant to the noise schedule of the forward diffusion process, except for at its endpoints. In other words, the VLB is unaffected by the shape of the signal-to-noise ratio function $\mathrm{SNR}(t)$ between $t=0$ and $t=1$. We also explain how this invariance holds for models that optimize a \textit{weighted} diffusion loss rather than the standard VLB, and connect the many weighted diffusion objectives used in practice to maximizing this principled variational lower bound.

\chapter{Understanding Diffusion Objectives}
\label{subsec: Understanding Diffusion Objectives}
In this section, we provide a deeper understanding of diffusion models. 
\newline We start by covering many of the different diffusion losses and network output parameterizations commonly used in literature, namely: 
\begin{enumerate}[(i)]
    \item Image Prediction~\citep{sohl2015deep,kingma2021variational};
    \item Noise Prediction~\citep{ho2020denoising};
    \item Score-based~\citep{song2021scorebased,song2021maximum};
    \item Energy-based~\citep{salimans2021should,du2023reduce};
    \item Velocity Prediction~\citep{salimans2022progressive};
    \item Flow-based~\citep{lipman2023flow,albergo2023building,liu2023flow}.
\end{enumerate}
We show in detail how all these parameterizations are equivalently valid, as each one can be expressed in terms of the others using a simple linear function. This implies that one can, for example, choose a parameterization for training and then reparameterize during inference. A common use case is to train a noise prediction model, then reparameterize it back to image space at inference time to clip the predicted image to a valid range (e.g. $[-1, 1]$) at each denoising step. This has been shown to mitigate the accumulation of sampling errors in practice, thereby improving sample quality.

We then explain the close connection between \textit{weighted} diffusion losses used in practice and maximizing the variational lower bound (i.e. the ELBO). Our exposition is designed to be instructive and consistent with VDM\texttt{++}~\citep{kingma2023understanding}, without departing too far from the material already covered and the notation already used. In Sections~\ref{subsec: Discrete-time Generative Model} and~\ref{subsec: Continuous-time Generative Model} we established that diffusion-based objectives correspond directly to the ELBO when the weighting function is uniform. However, the relationship between the non-uniform weighted diffusion objectives and the ELBO is less well understood, as they appear to optimize different things on the face of it. This has led to the widely held belief that the ELBO (i.e. maximum likelihood) may not be the correct objective for obtaining high-quality samples. 

Although weighted diffusion model objectives \textit{appear} markedly different from the ELBO, all commonly used diffusion objectives optimize a weighted integral of ELBOs over different noise levels~\citep{kingma2023understanding}. Furthermore, if the weighting function is monotonic, then the diffusion objective equates to the ELBO under simple Gaussian noise-based data augmentation. Lastly, we will show how different diffusion objectives imply specific weighting functions $w(\cdot)$ of the noise schedule. To avoid unnecessary repetition, we refer the reader to~\cite{kingma2023understanding} for a detailed breakdown of the most commonly used diffusion loss functions in the literature and the respective derivations of their implied weighting functions.
\begin{table}[!t]
    \small
    \centering
    \begin{tabular}{cr}
        \toprule
        \multirow{2}{*}{Model} & Posterior Mean \\
        & $\boldsymbol{\mu}_{\boldsymbol{\theta}}
        (\mathbf{z}_t;s,t)$ 
        \\
        \midrule
        \\[-8pt]
        Image Denoising & \multirow{2}{*}{$\displaystyle\frac{\alpha_{t|s}\sigma_s^2}{\sigma^2_{t}}\mathbf{z}_t + \frac{\alpha_s \sigma^2_{t|s}}{\sigma_{t}^2}\hat{\mathbf{x}}_{\boldsymbol{\theta}}(\mathbf{z}_t;t)$} \\ 
        $\hat{\mathbf{x}}_{\boldsymbol{\theta}}(\mathbf{z}_t;t)$ & 
        \\[10pt]
        Noise Prediction & \multirow{2}{*}{$\displaystyle\frac{1}{\alpha_{t|s}}\mathbf{z}_t - \frac{\sigma^2_{t|s} }{\alpha_{t|s}\sigma_{t}}\hat{\boldsymbol{\epsilon}}_{\boldsymbol{\theta}}(\mathbf{z}_t;t)$} \\
        $\hat{\boldsymbol{\epsilon}}_{\boldsymbol{\theta}}(\mathbf{z}_t;t)$ & 
        \\[10pt]
        Score-based & \multirow{2}{*}{$\displaystyle\frac{1}{\alpha_{t|s}}\mathbf{z}_t + \frac{\sigma^2_{t|s}}{\alpha_{t|s}}\mathbf{s}_{\boldsymbol{\theta}}(\mathbf{z}_t;t)$}\\
        $\mathbf{s}_{\boldsymbol{\theta}}(\mathbf{z}_t;t)$ & 
        \\[10pt]
        Energy-based & \multirow{2}{*}{$\displaystyle\frac{1}{\alpha_{t|s}}\mathbf{z}_t - \frac{\sigma^2_{t|s}}{\alpha_{t|s}}\nabla_{\mathbf{z}_t} E_{\boldsymbol{\theta}}(\mathbf{z}_t;t)$} \\
        $E_{\boldsymbol{\theta}}(\mathbf{z}_t;t)$ & 
        \\[10pt]
        Velocity Prediction & \multirow{2}{*}{$\displaystyle\frac{1 - \sigma^2_{t|s}}{\alpha_{t|s}}\mathbf{z}_t - \frac{\sigma^2_{t|s}\alpha_s}{\sigma_t} \hat{\mathbf{v}}_{\boldsymbol{\theta}}(\mathbf{z}_t;t)$} \\
        $\hat{\mathbf{v}}_{\boldsymbol{\theta}}(\mathbf{z}_t;t)$ & 
        \\[10pt]
        Flow-based & \multirow{2}{*}{$\displaystyle\frac{\sigma_t - \sigma^2_{t|s}}{\alpha_{t|s}\sigma_t}\mathbf{z}_t - \frac{\sigma^2_{t|s}\alpha_s}{\sigma_t} \hat{\mathbf{u}}_{\boldsymbol{\theta}}(\mathbf{z}_t;t)$} \\
        $\hat{\mathbf{u}}_{\boldsymbol{\theta}}(\mathbf{z}_t;t)$ & 
        \\[5pt]
        \bottomrule
    \end{tabular}
    \caption{\textbf{Equivalent ways of parameterizing a variational diffusion model}. Here $\boldsymbol{\mu}_{\boldsymbol{\theta}}(\mathbf{z}_t;s,t)$ is our model estimate of the true mean $\boldsymbol{\mu}_Q(\mathbf{z}_t, \mathbf{x};s,t)$ of the top-down posterior distribution $q(\mathbf{z}_s \mid \mathbf{z}_t, \mathbf{x})$, at any time $s < t$ (c.f. Section~\ref{subsubsec: qzs}).}
    \label{tab: equiv_param2}
\end{table}
\newpage
\section{Model Parameterizations}
As previously mentioned, there are multiple ways of operationalizing a diffusion model in terms of what the neural network outputs in practice. In the following, we present the most commonly used parameterizations, show that they are equivalently valid, and derive the associated expression for the posterior mean estimate of a variational diffusion model $\boldsymbol{\mu}_{\boldsymbol{\theta}}(\mathbf{z}_t;s,t)$. The results are summarized in Tables~\ref{tab: equiv_param2} and \ref{tab: equiv_param}.
\paragraph{Image Denoising {\normalfont model} {\mathversion{normal}$\hat{\mathbf{x}}_{\boldsymbol{\theta}}(\mathbf{z}_t;t)$}{\normalfont:}}
\begin{align}
    \boldsymbol{\mu}_{\boldsymbol{\theta}}(\mathbf{z}_t;s,t) = \frac{\alpha_{t|s}\sigma_s^2}{\sigma^2_{t}}\mathbf{z}_t + \frac{\alpha_s \sigma^2_{t|s}}{\sigma_{t}^2}\hat{\mathbf{x}}_{\boldsymbol{\theta}}(\mathbf{z}_t;t),
\end{align}
which as mentioned earlier, simply predicts $\mathbf{x}$ from its noisy versions $\mathbf{z}_t$, i.e. performs image \textit{denoising}. For a derivation, see Section~\ref{subsubsec: Deriving p}.
\paragraph{Noise Prediction {\normalfont model} {\mathversion{normal}$\hat{\boldsymbol{\epsilon}}_{\boldsymbol{\theta}}(\mathbf{z}_t;t)$}{\normalfont:}}
\begin{align}
    \boldsymbol{\mu}_{\boldsymbol{\theta}}(\mathbf{z}_t;s,t) = \frac{1}{\alpha_{t|s}}\mathbf{z}_t - \frac{\sigma^2_{t|s} }{\alpha_{t|s}\sigma_{t}}\hat{\boldsymbol{\epsilon}}_{\boldsymbol{\theta}}(\mathbf{z}_t;t),
    \label{eq: noise_net_out}
\end{align}
which we can derive in detail starting from the denoising model:
\begin{align}
\boldsymbol{\mu}_{\boldsymbol{\theta}}(\mathbf{z}_t;s,t) &= \frac{\alpha_{t|s}\sigma_s^2}{\sigma^2_{t}}\mathbf{z}_t + \frac{\alpha_s \sigma^2_{t|s}}{\sigma_{t}^2}\hat{\mathbf{x}}_{\boldsymbol{\theta}}(\mathbf{z}_t;t) \qquad\qquad\qquad\qquad
    \\[5pt] &= \frac{\alpha_{t|s}\sigma_s^2\mathbf{z}_t}{\sigma_t^2} + \frac{\alpha_s \sigma^2_{t|s}\left(\frac{\mathbf{z}_t - \sigma_t \hat{\boldsymbol{\epsilon}}_{\boldsymbol{\theta}}(\mathbf{z}_t;t)}{\alpha_t}\right)}{\sigma_{t}^2} 
    \customtag{since $\mathbf{x} = (\mathbf{z}_t - \sigma_t \boldsymbol{\epsilon}_t) / \alpha_t$}
    \\[5pt] &= \frac{\alpha_{t|s}}{\alpha_{t|s}} \cdot \frac{\alpha_{t|s}\sigma_s^2\mathbf{z}_t + \frac{\alpha_s \sigma^2_{t|s}\mathbf{z}_t}{\alpha_t}-\frac{\alpha_s \sigma^2_{t|s}\sigma_t \hat{\boldsymbol{\epsilon}}_{\boldsymbol{\theta}}(\mathbf{z}_t;t)}{\alpha_t}}{\sigma_{t}^2} 
    \customtag{recall that $\alpha_{t|s} = \frac{\alpha_t}{\alpha_s}$}
    \\[5pt] &= \frac{\frac{\alpha_t}{\alpha_s}\Big(\alpha_{t|s}\sigma_s^2\mathbf{z}_t + \frac{\alpha_s \sigma^2_{t|s}\mathbf{z}_t}{\alpha_t}-\frac{\alpha_s \sigma^2_{t|s}\sigma_t \hat{\boldsymbol{\epsilon}}_{\boldsymbol{\theta}}(\mathbf{z}_t;t)}{\alpha_t}\Big)}{\alpha_{t|s}\sigma_{t}^2} 
    \customtag{cancel common factors}
    \\[5pt] &= \frac{\alpha_{t|s}^2\sigma_s^2\mathbf{z}_t + \sigma^2_{t|s}\mathbf{z}_t - \sigma^2_{t|s}\sigma_t \hat{\boldsymbol{\epsilon}}_{\boldsymbol{\theta}}(\mathbf{z}_t;t)}{\alpha_{t|s}\sigma_{t}^2}
    \\[5pt] &= \frac{\mathbf{z}_t\left(\sigma_{t}^2 - \alpha_{t|s}^2\sigma_s^2 + \alpha_{t|s}^2\sigma_s^2\right)}{\alpha_{t|s}\sigma_{t}^2} - \frac{\sigma^2_{t|s}\sigma_t \hat{\boldsymbol{\epsilon}}_{\boldsymbol{\theta}}(\mathbf{z}_t;t)}{\alpha_{t|s}\sigma_{t}^2}
    \customtag{recall that $\sigma_{t|s}^2 = \sigma_t^2 - \alpha_{t|s}^2\sigma_s^2$}
    \\[5pt] &= \frac{\mathbf{z}_t\sigma_{t}^2}{\alpha_{t|s}\sigma_{t}^2} - \frac{\sigma^2_{t|s} \hat{\boldsymbol{\epsilon}}_{\boldsymbol{\theta}}(\mathbf{z}_t;t)}{\alpha_{t|s}\sigma_{t}} 
    \\[5pt] &= \frac{1}{\alpha_{t|s}}\mathbf{z}_t - \frac{\sigma^2_{t|s}}{\alpha_{t|s}\sigma_{t}}\hat{\boldsymbol{\epsilon}}_{\boldsymbol{\theta}}(\mathbf{z}_t;t).
\end{align}
\cite{ho2020denoising} first introduced noise prediction, which remains widely used for its practical effectiveness and close connection to score-based models~\citep{song2021scorebased}, as described next.
\begin{landscape}
    \begin{table}[!t]
    \small
    \centering
    \begin{tabular}{c|ccccc}
        \toprule
        \multirow{2}{*}{Model} & \multicolumn{5}{c}{{Equivalent Reparameterization}} 
        \\[2pt]
         & \multirow{1}{*}{$\hat{\mathbf{x}}_{\boldsymbol{\theta}}(\mathbf{z}_t;t)$} & \multirow{1}{*}{$\hat{\boldsymbol{\epsilon}}_{\boldsymbol{\theta}}(\mathbf{z}_t;t)$} & \multirow{1}{*}{${\mathbf{s}}_{\boldsymbol{\theta}}(\mathbf{z}_t;t)$} & \multirow{1}{*}{$\hat{\mathbf{v}}_{\boldsymbol{\theta}}(\mathbf{z}_t;t)$} & \multirow{1}{*}{$\hat{\mathbf{u}}_{\boldsymbol{\theta}}(\mathbf{z}_t;t)$}
        \\
        \midrule 
        \\[-8pt]
        Image Denoising & \multirow{2}{*}{-} & \multirow{2}{*}{$\displaystyle \frac{\mathbf{z}_t - \sigma_t\hat{\boldsymbol{\epsilon}}_{\boldsymbol{\theta}}}{\alpha_t}$} & \multirow{2}{*}{$\displaystyle\frac{\mathbf{z}_t + \sigma^2_t\mathbf{s}_{\boldsymbol{\theta}}}{\alpha_t}$} & \multirow{2}{*}{$\alpha_t \mathbf{z}_t - \sigma_t \hat{\mathbf{v}}_{\boldsymbol{\theta}}$} & \multirow{2}{*}{$\mathbf{z}_t - \sigma_t\hat{\mathbf{u}}_{\boldsymbol{\theta}}$}
        \\
        $\hat{\mathbf{x}}_{\boldsymbol{\theta}}(\mathbf{z}_t;t)$ & & & & &
        \\[10pt]
        Noise Prediction & \multirow{2}{*}{$\displaystyle\frac{\mathbf{z}_t - \alpha_t\hat{\mathbf{x}}_{\boldsymbol{\theta}}}{\sigma_t}$} & \multirow{2}{*}{-} & \multirow{2}{*}{$-\sigma_t{\mathbf{s}}_{\boldsymbol{\theta}}$} & \multirow{2}{*}{$\sigma_t \mathbf{z}_t + \alpha_t \hat{\mathbf{v}}_{\boldsymbol{\theta}}$} & \multirow{2}{*}{$\alpha_t\hat{\mathbf{u}}_{\boldsymbol{\theta}} + \mathbf{z}_t$}
        \\
        $\hat{\boldsymbol{\epsilon}}_{\boldsymbol{\theta}}(\mathbf{z}_t;t)$ & & & & &
        \\[10pt]
        Score-based & \multirow{2}{*}{$\displaystyle\frac{\alpha_t\hat{\mathbf{x}}_{\boldsymbol{\theta}} - \mathbf{z}_t)}{\sigma^2_t}$} & \multirow{2}{*}{$\displaystyle\frac{-\hat{\boldsymbol{\epsilon}}_{\boldsymbol{\theta}}}{ \sigma_t}$} & \multirow{2}{*}{-} & \multirow{2}{*}{$\displaystyle\frac{-\sigma_t\mathbf{z}_t + \alpha_t\hat{\mathbf{v}}_{\boldsymbol{\theta}}}{\sigma_t}$} & \multirow{2}{*}{$\displaystyle\frac{-\alpha_t\hat{\mathbf{u}}_{\boldsymbol{\theta}} + \mathbf{z}_t}{\sigma_t}$}
        \\
        ${\mathbf{s}}_{\boldsymbol{\theta}}(\mathbf{z}_t;t)$ & & & & &
        \\[10pt]
        Velocity Prediction & \multirow{2}{*}{$\displaystyle\frac{\alpha_t\mathbf{z}_t - \hat{\mathbf{x}}_{\boldsymbol{\theta}}}{\sigma_t}$} & \multirow{2}{*}{$\displaystyle\frac{\hat{\boldsymbol{\epsilon}}_{\boldsymbol{\theta}} - \sigma_t\mathbf{z}_t)}{\alpha_t}$} & \multirow{2}{*}{$\displaystyle\frac{-\sigma_t(\mathbf{z}_t + \mathbf{s}_{\boldsymbol{\theta}})}{\alpha_t}$} & \multirow{2}{*}{-} & \multirow{2}{*}{$\hat{\mathbf{u}}_{\boldsymbol{\theta}} - \mathbf{z}_t$}
        \\
        $\hat{\mathbf{v}}_{\boldsymbol{\theta}}(\mathbf{z}_t;t)$ & & & & &
        \\[10pt]
        Flow-Based & \multirow{2}{*}{$\displaystyle\frac{\mathbf{z}_t - \hat{\mathbf{x}}_{\boldsymbol{\theta}}}{\sigma_t}$} & \multirow{2}{*}{$\displaystyle\frac{\hat{\boldsymbol{\epsilon}}_{\boldsymbol{\theta}}(\alpha_t - \sigma_t) - \mathbf{z}_t}{\alpha_t}$} & \multirow{2}{*}{$\displaystyle\frac{\sigma_t\mathbf{s}_{\boldsymbol{\theta}}(\sigma_t - \alpha_t) - \mathbf{z}_t}{\alpha_t}$} & \multirow{2}{*}{$\mathbf{z}_t + \hat{\mathbf{v}}_{\boldsymbol{\theta}}$} & \multirow{2}{*}{-}
        \\
        $\hat{\mathbf{u}}_{\boldsymbol{\theta}}(\mathbf{z}_t;t)$ & & & & &
        \\[5pt]
        \bottomrule
    \end{tabular}
    \caption{\textbf{Equivalent reparameterizations of a variational diffusion model}. For example, image prediction can be expressed in terms of velocity prediction as $\hat{\mathbf{x}}_{\boldsymbol{\theta}}(\mathbf{z}_t;t) = \alpha_t\mathbf{z}_t - \sigma_t \hat{\mathbf{v}}_{\boldsymbol{\theta}}(\mathbf{z}_t;t)$. As shown, all the translation operations are simple linear functions of each other because $\mathbf{z}_t = \alpha_t\mathbf{x} + \sigma_t\boldsymbol{\epsilon}_t$ by the definition of the forward diffusion process. This implies that one can choose to parameterize the diffusion model using any of the mentioned methods during training and then re-parameterize it during inference. A common use case is to train a noise prediction model, then reparameterize back to image space at inference time to clip the predicted image to a valid range (e.g. $[-1, 1]$) at each denoising step $t$. Finally, noise and velocity prediction have sometimes demonstrated superior performance and numerical stability in practice.}
    \label{tab: equiv_param}
\end{table}
\end{landscape}
\paragraph{Score-based {\normalfont model} {\mathversion{normal}$\mathbf{s}_{\boldsymbol{\theta}}(\mathbf{z}_t;t)$}{\normalfont:}}
\begin{align}
    \boldsymbol{\mu}_{\boldsymbol{\theta}}(\mathbf{z}_t;s,t) = \frac{1}{\alpha_{t|s}}\mathbf{z}_t + \frac{\sigma^2_{t|s}}{\alpha_{t|s}}\mathbf{s}_{\boldsymbol{\theta}}(\mathbf{z}_t;t),
    \label{eq: score_net_mu}
\end{align}
which approximates $\nabla_{\mathbf{z}_t} \log q(\mathbf{z}_t)$, and is closely related to noise-prediction in the following way:
\begin{align}
    &\mathbf{s}_{\boldsymbol{\theta}}(\mathbf{z}_t;t) \approx \nabla_{\mathbf{z}_t} \log q(\mathbf{z}_t) \\[5pt] & = \mathbb{E}_{q(\mathbf{x})}\left[\nabla_{\mathbf{z}_t} \log q(\mathbf{z}_t \mid \mathbf{x})\right] 
    \customtag{marginal of the data $q(\mathbf{x})$}
    \\[5pt] & = \mathbb{E}_{q(\mathbf{x})}\left[\nabla_{\mathbf{z}_t} \log \mathcal{N}\left(\mathbf{z}_t;\alpha_{t} \mathbf{x}, \sigma^2_{t} \mathbf{I}\right)\right]
    \\[5pt] & = \mathbb{E}_{q(\mathbf{x})}\left[\nabla_{\mathbf{z}_t} \log \left( \prod_{i=1}^D \frac{1}{\sigma_t \sqrt{2\pi} } \exp\left\{-\frac{1}{2\sigma_t^2}\left(\mathbf{z}_{t,i} - \alpha_t \mathbf{x}_i\right)^2\right\} \right)\right] 
    \customtag{isotropic covariance}
    \\[5pt] & = \mathbb{E}_{q(\mathbf{x})}\left[\nabla_{\mathbf{z}_t} \left(-\frac{D}{2}\log \left(2\pi\sigma^2_t\right) -\frac{1}{2\sigma_t^2}\sum_{i=1}^D\left(\mathbf{z}_{t,i} - \alpha_t \mathbf{x}_i\right)^2 \right)\right]
    \\[5pt] & = \mathbb{E}_{q(\mathbf{x})}\left[-\frac{1}{\sigma_t^2} \left(\mathbf{z}_{t} - \alpha_t \mathbf{x}\right)\right] 
    \customtag{expected gradient}
    \\[5pt] & = \mathbb{E}_{q(\mathbf{x})}\left[-\frac{1}{\sigma_t} \hat{\boldsymbol{\epsilon}}_{\boldsymbol{\theta}}(\mathbf{z}_t;t)\right] 
    \customtag{due to $\boldsymbol{\epsilon} = (\mathbf{z}_t - \alpha_t \mathbf{x}) / \sigma_t$}
    \\[5pt] & = -\frac{1}{\sigma_t} \hat{\boldsymbol{\epsilon}}_{\boldsymbol{\theta}}(\mathbf{z}_t;t).
\end{align}
The optimal score model (with parameters $\boldsymbol{\theta}^*$) is equal to the gradient of the log-probability density w.r.t. the data at each noise scale: $\mathbf{s}_{\boldsymbol{\theta}^{*}}(\mathbf{z}_t;t) = \nabla_{\mathbf{z}_t}\log q(\mathbf{z}_t)$, for any $t$. This stems from a Score Matching with Langevin Dynamics (SMLD) perspective on generative modelling~\citep{song2019generative,song2021scorebased}. SMLD is closely related to probabilistic diffusion models~\citep{ho2020denoising}. For continuous state spaces, diffusion models implicitly compute the score at each noise scale, so the two approaches can be categorized jointly as \textit{score-based models}. For more details on score-based models, please refer to~\cite{song2021scorebased}.
\paragraph{Energy-based {\normalfont model} {\mathversion{normal}$E_{\boldsymbol{\theta}}(\mathbf{z}_t;t)$}{\normalfont:}}
\begin{align}
    \boldsymbol{\mu}_{\boldsymbol{\theta}}(\mathbf{z}_t;s,t) = \frac{1}{\alpha_{t|s}}\mathbf{z}_t - \frac{\sigma^2_{t|s}}{\alpha_{t|s}}\nabla_{\mathbf{z}_t} E_{\boldsymbol{\theta}}(\mathbf{z}_t;t),
    \label{eq: energy_net_mu}
\end{align}
since the score model can be parameterized with the gradient of an energy-based model:
\begin{align}
    \mathbf{s}_{\boldsymbol{\theta}}(\mathbf{z}_t;t) & \approx \nabla_{\mathbf{z}_t} \log q(\mathbf{z}_t) 
    \\[5pt] & = \nabla_{\mathbf{z}_t} \log \left(\frac{1}{Z}\exp\left(-E_{\boldsymbol{\theta}}(\mathbf{z}_t;t)\right) \right) 
    \customtag{Boltzmann dist.}
    \\[5pt] & = \nabla_{\mathbf{z}_t} \Big(-E_{\boldsymbol{\theta}}(\mathbf{z}_t;t) - \log Z \Big) \customtag{$\nabla_{\mathbf{z}_t} \log Z = 0$}
    \\[5pt] & = -\nabla_{\mathbf{z}_t} E_{\boldsymbol{\theta}}(\mathbf{z}_t;t),
\end{align}
which we can use to substitute $\mathbf{s}_{\boldsymbol{\theta}}(\mathbf{z}_t;t)$ in Equation~\ref{eq: score_net_mu} to get the new expression in Equation~\ref{eq: energy_net_mu}. \cite{du2023reduce} provide compelling arguments in favour of an energy-based parameterization as it enables the use of more sophisticated sampling schemes and forms of composition. 

With that said, the energy-based parameterization of diffusion models remains comparatively underexplored to date. For a detailed review of energy-based models and their relationship with score-based models refer to~\cite{song2021train} and \cite{salimans2021should}.
\paragraph{Velocity Prediction {\normalfont model} \mathversion{normal}{$\hat{\mathbf{v}}_{\boldsymbol{\theta}}(\mathbf{z}_t;t)$}{\normalfont:}}
\begin{align}
    \boldsymbol{\mu}_{\boldsymbol{\theta}}(\mathbf{z}_t;s,t) = \frac{1 - \sigma^2_{t|s}}{\alpha_{t|s}}\mathbf{z}_t - \frac{\sigma^2_{t|s}\alpha_s}{\sigma_t} \hat{\mathbf{v}}_{\boldsymbol{\theta}}(\mathbf{z}_t;t),
    \label{eq: velocity_net_mu}
\end{align}
which represents the tangential velocity resulting from an angular parameterization $\phi_t = \mathrm{arctan}(\sigma_t / \alpha_t)$ of the latent variable $\mathbf{z}_t$ specified as follows~\citep{salimans2022progressive}:
\begin{align}
    \mathbf{z}_t &= \cos(\phi_t) \mathbf{x} + \sin(\phi_t) \boldsymbol{\epsilon}, 
    \\[5pt]\mathbf{v} &\coloneqq \dv{ \mathbf{z}_t}{\phi_t} = \cos(\phi_t) \boldsymbol{\epsilon} - \sin(\phi_t)\mathbf{x}.
\end{align}
The velocity $\mathbf{v}$ represents how the latent variable $\mathbf{z}_t$ changes w.r.t. $\phi_t$, and is analogous to the tangential velocity of a point moving on a circle. This may be alternatively described as $\mathbf{v} = \| \mathbf{z}_t\| \dv{\phi_t}{t}$, where $\| \mathbf{z}_t\|$ is the radius of the circular path and $\dv{\phi_t}{t}$ is the angular velocity.

\begin{figure}[t!]
    \centering
    \begin{tikzpicture}[declare function={angle=45;},bullet/.style={inner sep=1pt,fill,draw,circle,solid}, scale=3]
        \draw[-Latex,black, thick] (0,0)--(1.12,0) node[below] {$\mathbf{x}$};
        \draw[-Latex,black, thick] (0,0)--(0,1.12) node[left] {$\boldsymbol{\epsilon}$};
        \path (0,0) coordinate (O);
        \draw[] (1,0) arc[start angle=0, end angle=90, radius=1];
        \draw[red!50!black, thick,-Latex] (O) -- (angle:1) node[midway,xshift=-2pt,above=0pt]{\color{black}$\mathbf{z}$};
        \draw[] (angle:1) -- (0:{cos(angle)});
        \draw[blue,thick,-Latex] (angle:1) -- ++({sin(angle)},{-cos(angle)}) node[midway,xshift=1pt,above=1pt]{\color{black}$\mathbf{v}$};
        \draw[-{[bend]}] (angle/1.7:0.23) node{$\phi$} (0:.15) arc(0:angle:0.15);
        \draw [decoration={brace, mirror}, decorate] 
        (0,-0.05) -- ({cos(angle)},-0.05) node[midway,below=0.5ex]{$\alpha$};
        \draw [decoration={brace}, decorate] 
        (-0.05,0) -- (-0.05, {sin(angle)}) node[midway,left=0.5ex]{$\sigma$};
        \draw (angle:.9) -- ++({0.1*sin(angle)},{-0.1*cos(angle)}) -- ++({0.1*cos(angle)},{0.1*sin(angle)});
    \end{tikzpicture}
    \caption{Illustrating the relationships between $\mathbf{v}$-prediction (i.e. the tangential velocity) and the noise schedule quantities involved in defining a diffusion process. Figure inspired by~\cite{salimans2022progressive}.}
    \label{fig: v_prediction_diagram}
\end{figure}
To derive our result in Equation~\ref{eq: velocity_net_mu}, we can start by expressing $\boldsymbol{\epsilon}$ in terms of $\mathbf{v}$:
\begin{align}
    \mathbf{v} & = \cos(\phi_t) \boldsymbol{\epsilon} - \sin(\phi_t)\mathbf{x} 
    \\[2pt] & \coloneqq \alpha_t \boldsymbol{\epsilon} - \sigma_t \mathbf{x} \\[2pt] &= \alpha_t \boldsymbol{\epsilon} - \sigma_t \left(\frac{\mathbf{z}_t - \sigma_t \boldsymbol{\epsilon}}{\alpha_t}\right)
    \\[2pt] \mathbf{v} +\frac{\sigma_t \mathbf{z}_t}{\alpha_t} & = \boldsymbol{\epsilon} \left( \frac{\alpha_t^2}{\alpha_t}+ \frac{\sigma_t^2}{\alpha_t}\right)
    \\[2pt] \mathbf{v} +\frac{\sigma_t \mathbf{z}_t}{\alpha_t} & = \boldsymbol{\epsilon} \left( \frac{1 - \sigma_t^2 + \sigma_t^2}{\alpha_t}\right) \customtag{recall $\alpha^2 = 1- \sigma_t^2$}
    \\[2pt] \alpha_t\mathbf{v} + \sigma_t \mathbf{z}_t & = \boldsymbol{\epsilon} 
    \\[5pt] \implies \hat{\boldsymbol{\epsilon}}_{\boldsymbol{\theta}}(\mathbf{z}_t;t) & = \alpha_t \hat{\mathbf{v}}_{\boldsymbol{\theta}}(\mathbf{z}_t;t) + \sigma_t \mathbf{z}_t.
\end{align}
Now recalling that $\mathbf{s}_{\boldsymbol{\theta}}(\mathbf{z}_t;t) = -\hat{\boldsymbol{\epsilon}}_{\boldsymbol{\theta}}(\mathbf{z}_t;t)/\sigma_t$, and substituting the above into the score-based parameterization in Equation~\ref{eq: score_net_mu}, we get:
\begin{align}
    \boldsymbol{\mu}_{\boldsymbol{\theta}}(\mathbf{z}_t;s,t) & = \frac{\mathbf{z}_t}{\alpha_{t|s}} + \frac{\sigma^2_{t|s}}{\alpha_{t|s}} \cdot - \frac{1}{\sigma_t}\left(\alpha_t \hat{\mathbf{v}}_{\boldsymbol{\theta}}(\mathbf{z}_t;t) + \sigma_t \mathbf{z}_t\right) 
    \\[5pt] & = \frac{\mathbf{z}_t}{\alpha_{t|s}} - \frac{\sigma^2_{t|s}\mathbf{z}_t}{\alpha_{t|s}} - \frac{\sigma^2_{t|s}\alpha_t}{\alpha_{t|s}\sigma_t}\hat{\mathbf{v}}_{\boldsymbol{\theta}}(\mathbf{z}_t;t)
    \\[5pt] & = \frac{1 - \sigma^2_{t|s}}{\alpha_{t|s}}\mathbf{z}_t - \frac{\sigma^2_{t|s}\alpha_s}{\sigma_t}\hat{\mathbf{v}}_{\boldsymbol{\theta}}(\mathbf{z}_t;t). 
\end{align}
As we show next, there exists a close link between velocity prediction and Flow Matching~\citep{lipman2023flow,albergo2023building,liu2023flow}, which can be interpreted as a type of Gaussian diffusion.~\cite{kingma2023understanding} explore this relation under what they call $\mathbf{o}$-prediction (c.f. Appendix D.3 in~\cite{kingma2023understanding}).
\paragraph{Flow-based {\normalfont model} \mathversion{normal}{$\hat{\mathbf{u}}_{\boldsymbol{\theta}}(\mathbf{z}_t;t)$}{\normalfont:}}
\begin{align}
    \boldsymbol{\mu}_{\boldsymbol{\theta}}(\mathbf{z}_t;s,t) = \frac{\sigma_t - \sigma_{t|s}^2}{\alpha_{t|s}\sigma_t}\mathbf{z}_t - \frac{\sigma^2_{t|s}\alpha_s }{\sigma_{t}}
    \hat{\mathbf{u}}_{\boldsymbol{\theta}}(\mathbf{z}_t;t),
    \label{eq: flow_net_mu}
\end{align}
where $\hat{\mathbf{u}}_{\boldsymbol{\theta}}$ also represents a velocity field, but here the noise schedule of choice is a \textit{linear interpolation} between data and noise over time. In our context, this translates to each latent variable $\mathbf{z}_t$ being given by:
\begin{align}
    \mathbf{z}_t & = \alpha_t \mathbf{x} + \sigma_t \boldsymbol{\epsilon}
    \\[2pt] & 
    = (1 - t) \mathbf{x} + t \boldsymbol{\epsilon},
\end{align}
and the velocity field $\mathbf{u}$ is obtained by differentiating w.r.t. time $t$:
\begin{align}
    \mathbf{u} \coloneqq \dv{\mathbf{z}_t}{t} = \boldsymbol{\epsilon} - \mathbf{x}.
\end{align}
To derive our result in Equation~\ref{eq: flow_net_mu}, we first express $\boldsymbol{\epsilon}$ in terms of $\mathbf{u}$:
\begin{align}
    \boldsymbol{\epsilon} & = \mathbf{u} + \mathbf{x}
    \\[2pt] & = \mathbf{u} + \left( \frac{\mathbf{z}_t - t \boldsymbol{\epsilon}}{1-t} \right)
    \\[2pt] \boldsymbol{\epsilon}\left(\frac{1}{1-t}\right) & = \mathbf{u} + \frac{\mathbf{z}_t}{1-t}
    \\[2pt] \boldsymbol{\epsilon} & = (1-t)\mathbf{u} + \mathbf{z}_t
    \\[2pt] \implies \hat{\boldsymbol{\epsilon}}_{\boldsymbol{\theta}}(\mathbf{z}_t;t) & = \alpha_t\hat{\mathbf{u}}_{\boldsymbol{\theta}}(\mathbf{z}_t;t) + \mathbf{z}_t.
\end{align}
Now we simply substitute the above into the noise prediction parameterization in Equation~\ref{eq: noise_net_out} and simplify:
\begin{align}
    \boldsymbol{\mu}_{\boldsymbol{\theta}}(\mathbf{z}_t;s,t) & = \frac{1}{\alpha_{t|s}}\mathbf{z}_t - \frac{\sigma^2_{t|s} }{\alpha_{t|s}\sigma_{t}}
    (\alpha_t\hat{\mathbf{u}}_{\boldsymbol{\theta}}(\mathbf{z}_t;t) + \mathbf{z}_t)
    \\[5pt] & = \frac{\mathbf{z}_t\sigma_t}{\alpha_{t|s}\sigma_t} - \frac{\sigma_{t|s}^2\mathbf{z}_t}{\alpha_{t|s}\sigma_t} - \frac{\sigma^2_{t|s}\alpha_t}{\alpha_{t|s}\sigma_{t}}
    \hat{\mathbf{u}}_{\boldsymbol{\theta}}(\mathbf{z}_t;t)
    \\[5pt] & = \frac{\sigma_t - \sigma_{t|s}^2}{\alpha_{t|s}\sigma_t}\mathbf{z}_t - \frac{\sigma^2_{t|s}\alpha_s }{\sigma_{t}}
    \hat{\mathbf{u}}_{\boldsymbol{\theta}}(\mathbf{z}_t;t).
\end{align}
The $\mathbf{u}$-prediction parameterization is best known as Flow Matching (FM)~\citep{lipman2023flow}, has many other flavours~\citep{albergo2023building,liu2023flow}, and is reminiscent of velocity prediction~\citep{salimans2022progressive}. As shown above, Gaussian FM is practically equivalent to (continuous-time) diffusion modelling but uses a particular neural network parameterization and noise schedule.

To conclude, one other parameterization not covered here but certainly worth learning about is $\mathbf{F}$-prediction~\citep{karras2022elucidating}.

\paragraph{Practical Advantages.} Each aforementioned parameterization has its own strengths and weaknesses. For example, noise prediction $\hat{\boldsymbol{\epsilon}}_{\boldsymbol{\theta}}$ is stable at lower noise levels but can explode errors at high noise levels, as the \textit{implied} image prediction $\hat{\mathbf{x}}_{\boldsymbol{\theta}}$ approaches a division by zero:
\begin{align}
     &&\hat{\mathbf{x}}_{\boldsymbol{\theta}} = \frac{\mathbf{z}_t - \sigma_t \hat{\boldsymbol{\epsilon}}_{\boldsymbol{\theta}}}{\alpha_t}, && \mathrm{where} && \alpha_t \xrightarrow{t\to1} 0.
\end{align}
The same is true for the score-based parameterization. In fact, any reparameterization formula that involves division by one of the noise schedule parameters is subject to this problem, and these can simply be read off Table~\ref{tab: equiv_param}. Importantly, both of the velocity prediction versions (i.e. $\hat{\mathbf{v}}_{\boldsymbol{\theta}}$ and $\hat{\mathbf{u}}_{\boldsymbol{\theta}}$) overcome this issue, as the corresponding image prediction formulas do \textit{not} require any division by a noise schedule parameter. Direct image prediction overcomes the division by zero issue too but is subject to a more subtle problem: it provides a weak supervision signal at low noise levels. Collectively, these considerations make a compelling argument for adopting velocity-based parameterizations.
%
\begin{table}[t]
    \footnotesize
    \centering
    \begin{tabular}{l|ccc}
        \toprule
        \multirow{2}{*}{Loss} & Image Denoising & Noise Prediction & Velocity Prediction 
        \\[2pt]
        & $\|\mathbf{x} -\hat{\mathbf{x}}_{\boldsymbol{\theta}}(\mathbf{z}_t;t) \|_2^2$ & 
        $\|\boldsymbol{\epsilon} - \hat{\boldsymbol{\epsilon}}_{\boldsymbol{\theta}}(\mathbf{z}_t;t) \|_2^2$ & $\|\mathbf{v} -\hat{\mathbf{v}}_{\boldsymbol{\theta}}(\mathbf{z}_t;t) \|_2^2$ 
        \\
        \midrule 
        \\[-8pt]
        $\|\mathbf{x} -\hat{\mathbf{x}}_{\boldsymbol{\theta}}(\mathbf{z}_t;t) \|_2^2$ & 1 & $\sigma_t^2 / \alpha_t^2$ & $\sigma_t^2$
        \\[10pt]
        $\|\boldsymbol{\epsilon} - \hat{\boldsymbol{\epsilon}}_{\boldsymbol{\theta}}(\mathbf{z}_t;t) \|_2^2$ & $\alpha_t^2 / \sigma_t^2$ & 1 & $1/\alpha_t^2$  
        \\[5pt]
        $\|\mathbf{v} -\hat{\mathbf{v}}_{\boldsymbol{\theta}}(\mathbf{z}_t;t) \|_2^2$ & $\sigma^2_t\left(\frac{\alpha_t^2}{ \sigma_t^2} + 1\right)^2$ & $\alpha^2_t\left(\frac{\sigma_t^2}{\alpha_t^2} + 1\right)^2$ & 1  
        \\[5pt]
        \bottomrule
    \end{tabular}
    \caption{\textbf{Translating diffusion model loss parameterizations}. Each loss on the LHS column can be rewritten in terms of the other parameterizations weighted by a specific constant. For example, the image prediction loss can be written in terms of noise prediction weighted by $\sigma_t^2 /\alpha_t^2$, that is: $\|\mathbf{x} -\hat{\mathbf{x}}_{\boldsymbol{\theta}}(\mathbf{z}_t;t) \|_2^2 = \sigma_t^2 /\alpha_t^2 \|\boldsymbol{\epsilon} - \hat{\boldsymbol{\epsilon}}_{\boldsymbol{\theta}}(\mathbf{z}_t;t) \|_2^2$, whereas the $\mathbf{v}$-prediction~\citep{salimans2022progressive} loss can be written in terms of image prediction by: $\|\mathbf{v} -\hat{\mathbf{v}}_{\boldsymbol{\theta}}(\mathbf{z}_t;t) \|_2^2 = \sigma^2_t\left(\alpha_t^2 / \sigma_t^2 + 1\right)^2 \|\mathbf{x} -\hat{\mathbf{x}}_{\boldsymbol{\theta}}(\mathbf{z}_t;t) \|_2^2$.
    }
    \label{tab: equiv_losses}
\end{table}
\section{Translating Loss Parameterizations} Translating between different loss parameterizations is straightforward due to the linearity of the forward diffusion process $\mathbf{z}_t = \alpha_t \mathbf{x} + \sigma_t \boldsymbol{\epsilon}$. This will be particularly useful for analyzing diffusion loss objectives later on. For now, we provide the derivations of the most popular loss parameterizations and summarize the results in Table~\ref{tab: equiv_losses}. 

\paragraph{Image to Noise Prediction.} Firstly, we rewrite the image prediction $\hat{\mathbf{x}}_{\boldsymbol{\theta}}(\mathbf{z}_t;t)$ loss in terms of noise prediction $\hat{\boldsymbol{\epsilon}}_{\boldsymbol{\theta}}(\mathbf{z}_t;t)$ by:
\begin{align}
    \left\| \mathbf{x} - \hat{\mathbf{x}}_{\boldsymbol{\theta}}(\mathbf{z}_t;t) \right\|^2_2 & = \left\| \frac{\mathbf{z}_t - \sigma_t \boldsymbol{\epsilon}}{\alpha_t} - \frac{\mathbf{z}_t - \sigma_t \hat{\boldsymbol{\epsilon}}_{\boldsymbol{\theta}}(\mathbf{z}_t;t)}{\alpha_t} \right\|^2_2 \customtag{since $\mathbf{z}_t = \alpha_t \mathbf{x} + \sigma_t \boldsymbol{\epsilon}$}
    \\[5pt] & = \frac{\sigma_t^2}{\alpha_t^2}\left\| \boldsymbol{\epsilon} - \hat{\boldsymbol{\epsilon}}_{\boldsymbol{\theta}}(\mathbf{z}_t;t) \right\|^2_2. \customtag{cancel terms and factor}
\end{align}
\paragraph{Image to Velocity Prediction.} Similarly, in terms of velocity prediction, we first review the definition:
\begin{align}
    \mathbf{v} & \coloneqq \alpha_t \boldsymbol{\epsilon} - \sigma_t \mathbf{x} \customtag{by definition} 
    \\[5pt] & = \alpha_t \left(\frac{\mathbf{z}_t - \alpha_t\mathbf{x}}{\sigma_t}\right) - \sigma_t \mathbf{x} \customtag{substitute $\boldsymbol{\epsilon} = (\mathbf{z}_t - \alpha_t\mathbf{x}) / \sigma_t$},
\end{align}
since $\alpha_t^2 = 1 - \sigma_t^2$ in a variance preserving process we then get:
\begin{align}
    \alpha_t\mathbf{z}_t - \sigma_t \mathbf{v} & = (1-\sigma_t^2)\mathbf{x} + \sigma_t^2\mathbf{x}
    \\[5pt] \implies \mathbf{x} & = \alpha_t\mathbf{z}_t - \sigma_t \mathbf{v},
\end{align}
which we can now substitute into the image prediction loss, with a $\mathbf{v}$-prediction model $\hat{\mathbf{v}}_{\boldsymbol{\theta}}(\mathbf{z}_t;t)$:
\begin{align}
    \left\| \mathbf{x} - \hat{\mathbf{x}}_{\boldsymbol{\theta}}(\mathbf{z}_t;t) \right\|^2_2 & = \left\| \alpha_t\mathbf{z}_t - \sigma_t \mathbf{v} - \left( \alpha_t\mathbf{z}_t - \sigma_t \hat{\mathbf{v}}_{\boldsymbol{\theta}}(\mathbf{z}_t;t) \right) \right\|^2_2 
    \\[5pt] & = \left\| \sigma_t \hat{\mathbf{v}}_{\boldsymbol{\theta}}(\mathbf{z}_t;t) - \sigma_t \mathbf{v} \right\|^2_2 \customtag{$\alpha_t\mathbf{z}_t$ terms cancel}
    \\[5pt] & = \sigma_t^2\left\| \hat{\mathbf{v}}_{\boldsymbol{\theta}}(\mathbf{z}_t;t) - \mathbf{v} \right\|^2_2. \customtag{by factoring}
\end{align}
\paragraph{Noise to Image Prediction.} To rewrite the noise prediction loss in terms of image prediction we have:
\begin{align}
    \left\| \boldsymbol{\epsilon} - \hat{\boldsymbol{\epsilon}}_{\boldsymbol{\theta}}(\mathbf{z}_t;t) \right\|^2_2 & = \left\| \frac{\mathbf{z}_t - \alpha_t \mathbf{x}}{\sigma_t} - \frac{\mathbf{z}_t - \alpha_t \hat{\mathbf{x}}_{\boldsymbol{\theta}}(\mathbf{z}_t;t)}{\sigma_t} \right\|^2_2 \customtag{recall that $\mathbf{z}_t = \alpha_t \mathbf{x} + \sigma_t \boldsymbol{\epsilon}$}
    \\[5pt] & = \left\| \frac{\alpha_t}{\sigma_t} \left(\hat{\mathbf{x}}_{\boldsymbol{\theta}}(\mathbf{z}_t;t) - \mathbf{x} \right) \right\|^2_2 \customtag{cancel $\mathbf{z}_t$ terms and factor}
    \\[5pt] & = \mathrm{SNR}(t) \left\| \hat{\mathbf{x}}_{\boldsymbol{\theta}}(\mathbf{z}_t;t) - \mathbf{x} \right\|^2_2. \customtag{recall $\mathrm{SNR}(t) = \alpha^2_t/\sigma^2_t$}
\end{align}
\paragraph{Noise to Velocity Prediction.} To express noise prediction in terms of velocity prediction we proceed as:
\begin{align}
    \left\| \boldsymbol{\epsilon} - \hat{\boldsymbol{\epsilon}}_{\boldsymbol{\theta}}(\mathbf{z}_t;t) \right\|^2_2 & = \left\| \frac{\mathbf{v} + \sigma_t \mathbf{x}}{\alpha_t} - \frac{\hat{\mathbf{v}}_{\boldsymbol{\theta}}(\mathbf{z}_t;t) + \sigma_t \mathbf{x}}{\alpha_t} \right\|^2_2 \customtag{solving $\mathbf{v} = \alpha_t \boldsymbol{\epsilon} - \sigma_t \mathbf{x}$ for $\boldsymbol{\epsilon}$}
    \\[5pt] & = \left\| \frac{1}{\alpha_t} \left(\mathbf{v} -\hat{\mathbf{v}}_{\boldsymbol{\theta}}(\mathbf{z}_t;t) \right) \right\|^2_2 \customtag{cancel $\mathbf{x}$ terms and factor}
    \\[5pt] & =  \frac{1}{\alpha_t^2} \left\| \mathbf{v} 
 - \hat{\mathbf{v}}_{\boldsymbol{\theta}}(\mathbf{z}_t;t)\right\|^2_2.
\end{align}
\paragraph{Velocity to Image Prediction.} We can rewrite velocity prediction in terms of image prediction as follows:
\begin{align}
    &\left\| \mathbf{v} 
    - \hat{\mathbf{v}}_{\boldsymbol{\theta}}(\mathbf{z}_t;t)\right\|^2_2 = \left\| \alpha_t\boldsymbol{\epsilon} - \sigma_t \mathbf{x} - \left( \alpha_t\hat{\boldsymbol{\epsilon}}_{\boldsymbol{\theta}}(\mathbf{z}_t;t) - \sigma_t \hat{\mathbf{x}}_{\boldsymbol{\theta}}(\mathbf{z}_t;t) \right) \right\|^2_2
    \\[5pt] & = \left\| \alpha_t\left( \frac{\mathbf{z}_t - \alpha_t\mathbf{x}}{\sigma_t}\right) - \sigma_t \mathbf{x} - \alpha_t\left( \frac{\mathbf{z}_t - \alpha_t\hat{\mathbf{x}}_{\boldsymbol{\theta}}(\mathbf{z}_t;t)}{\sigma_t}\right) + \sigma_t \hat{\mathbf{x}}_{\boldsymbol{\theta}}(\mathbf{z}_t;t) \right\|^2_2
    \\[5pt] & = \left\| \frac{\alpha_t^2\hat{\mathbf{x}}_{\boldsymbol{\theta}}(\mathbf{z}_t;t)}{\sigma_t} + \sigma_t\hat{\mathbf{x}}_{\boldsymbol{\theta}}(\mathbf{z}_t;t) - \frac{\alpha_t^2\mathbf{x}}{\sigma_t} - \sigma_t\mathbf{x} \right\|^2_2 \customtag{cancel $\mathbf{z}_t$ terms and factor}
    \\[5pt] & = \left\| \left(\frac{\alpha_t^2}{\sigma_t} + \sigma_t\right) \left(\hat{\mathbf{x}}_{\boldsymbol{\theta}}(\mathbf{z}_t;t) - \mathbf{x} \right)\right\|^2_2
    \\[5pt] & = \sigma_t^2\left(\mathrm{SNR}(t) + 1\right)^2 \left\| \hat{\mathbf{x}}_{\boldsymbol{\theta}}(\mathbf{z}_t;t) - \mathbf{x} \right\|^2_2 \customtag{as $\mathrm{SNR}(t) = \alpha^2_t/\sigma^2_t$}.
\end{align}
\paragraph{Velocity to Noise Prediction.} We proceed similarly to the above for noise prediction, but now substituting image-related terms to get:
\begin{align}
    &\left\| \mathbf{v} 
 - \hat{\mathbf{v}}_{\boldsymbol{\theta}}(\mathbf{z}_t;t)\right\|^2_2 = \left\| \alpha_t\boldsymbol{\epsilon} - \sigma_t \mathbf{x} - \left( \alpha_t\hat{\boldsymbol{\epsilon}}_{\boldsymbol{\theta}}(\mathbf{z}_t;t) - \sigma_t \hat{\mathbf{x}}_{\boldsymbol{\theta}}(\mathbf{z}_t;t) \right) \right\|^2_2 
 \\[5pt] & = \left\| \alpha_t\boldsymbol{\epsilon} - \sigma_t \left( \frac{\mathbf{z}_t - \sigma_t\boldsymbol{\epsilon}}{\alpha_t}\right) - \alpha_t\hat{\boldsymbol{\epsilon}}_{\boldsymbol{\theta}}(\mathbf{z}_t;t) + \sigma_t \left( \frac{\mathbf{z}_t - \sigma_t\hat{\boldsymbol{\epsilon}}_{\boldsymbol{\theta}}(\mathbf{z}_t;t)}{\alpha_t}\right) \right\|^2_2
 \\[5pt] & = \left\| \left(\frac{\sigma_t^2}{\alpha_t} + \alpha_t\right) \left( \boldsymbol{\epsilon} - \hat{\boldsymbol{\epsilon}}_{\boldsymbol{\theta}}(\mathbf{z}_t;t) \right)\right\|^2_2 \customtag{cancel $\mathbf{z}_t$ terms and factor}
\\[5pt] & = \alpha^2_t\left(\frac{\sigma_t^2}{\alpha^2_t} + 1\right)^2 \left\| \boldsymbol{\epsilon} - \hat{\boldsymbol{\epsilon}}_{\boldsymbol{\theta}}(\mathbf{z}_t;t) \right\|^2_2.
\end{align}
\newpage
\section{Invariance to the Noise Schedule}
\label{subsubsec: Invariance to the Noise Schedule}
An important result established that the continuous-time VLB is invariant to the noise schedule of the forward diffusion process~\citep{kingma2021variational}. To explain this result, we begin by performing a change of variables; i.e. we transform the integral w.r.t time $t$ in the diffusion loss (Equation~\ref{eq: t_int}) into an integral w.r.t the signal-to-noise ratio. Since the signal-to-noise ratio function $\mathrm{SNR}(t)= \exp(-\gamma_{\boldsymbol{\eta}}(t))$ is \textit{monotonic}, it is invertible ($\mathrm{SNR}(t)$ is entirely non-increasing in time $t$ meaning: $\mathrm{SNR}(t) < \mathrm{SNR}(s)$, for any $t>s$; ref. Section~\ref{subsubsec: Noise Schedule}). Using this fact, we can re-express our loss in terms of a new variable $v \equiv \mathrm{SNR}(t)$, such that time $t$ is instead given by $t = \mathrm{SNR}^{-1}(v)$. Let $\mathbf{z}_v = \alpha_v \mathbf{x} + \sigma_v \boldsymbol{\epsilon}$ denote the latent variable $\mathbf{z}_v$ whose noise-schedule functions $\alpha_v$ and $\sigma_v$ correspond to $\alpha_t$ and $\sigma_t$ evaluated at $t = \mathrm{SNR}^{-1}(v)$. 

By applying the \textit{integration by substitution} formula 
\begin{align}
    \int_a^b f(g(t)) \cdot g'(t) \mathop{\mathrm{d} t} = \int_{g(a)}^{g(b)} f(v) \mathop{\mathrm{d} v},
\end{align}
we can express the diffusion loss in terms of our new variable $v$ as follows:
\begin{align}
    &\mathcal{L}_\infty(\mathbf{x}) = -\frac{1} {2}\mathbb{E}_{\boldsymbol{\epsilon} \sim \mathcal{N}(0,\mathbf{I}),t \sim \mathcal{U}(0,1)}\left[\mathrm{SNR}'(t)\left\| \mathbf{x} - \hat{\mathbf{x}}_{\boldsymbol{\theta}}\left( \mathbf{z}_t;t\right) \right\|^2_2 \right]
    \\[5pt] & = -\frac{1} {2}\mathbb{E}_{\boldsymbol{\epsilon} \sim \mathcal{N}(0,\mathbf{I})}\left[\int_0^1\left\| \mathbf{x} - \hat{\mathbf{x}}_{\boldsymbol{\theta}}\left( \sigma_t\left(\mathbf{x}\sqrt{\mathrm{SNR}(t)} + \boldsymbol{\epsilon}\right) ;t\right) \right\|^2_2 \cdot \mathrm{SNR}'(t)\mathop{\mathrm{d}t}\right] 
    \\[5pt] & = -\frac{1} {2}\mathbb{E}_{\boldsymbol{\epsilon} \sim \mathcal{N}(0,\mathbf{I})}\left[\int_{\mathrm{SNR}(0)}^{\mathrm{SNR}(1)}\left\| \mathbf{x} - \hat{\mathbf{x}}_{\boldsymbol{\theta}}\left( \mathbf{z}_v;v\right) \right\|^2_2 \mathop{\mathrm{d}v}\right] \customtag{$\mathrm{d}v = \mathrm{SNR}'(t) \mathop{\mathrm{d}t}$}
    \\[5pt] & = \frac{1} {2}\mathbb{E}_{\boldsymbol{\epsilon} \sim \mathcal{N}(0,\mathbf{I})}\left[\int_{\mathrm{SNR}(1)}^{\mathrm{SNR}(0)}\left\| \mathbf{x} - \hat{\mathbf{x}}_{\boldsymbol{\theta}}\left( \mathbf{z}_v;v\right) \right\|^2_2 \mathop{\mathrm{d}v}\right] \customtag{swap limits}
    \\[5pt] & = \frac{1} {2}\mathbb{E}_{\boldsymbol{\epsilon} \sim \mathcal{N}(0,\mathbf{I})}\left[\int_{\mathrm{SNR}_{\mathrm{min}}}^{\mathrm{SNR}_{\mathrm{max}}}\left\| \mathbf{x} - \hat{\mathbf{x}}_{\boldsymbol{\theta}}\left( \mathbf{z}_v;v\right) \right\|^2_2 \mathop{\mathrm{d}v}\right], \label{eq:snrminmax}
\end{align}
where $\mathrm{SNR}_{\mathrm{max}} = \mathrm{SNR}(0)$ denotes the \textit{highest} signal-to-noise ratio at time $t=0$ resulting in the least noisy latent $\mathbf{z}_0$ at the start of the diffusion process (i.e. essentially the same as $\mathbf{x}$). Conversely, $\mathrm{SNR}_{\mathrm{min}} = \mathrm{SNR}(1)$ denotes the \textit{lowest} signal-to-noise ratio resulting in the noisiest latent $\mathbf{z}_1$ at time $t=1$.

The above shows that the diffusion loss is determined by the endpoints $\mathrm{SNR}_{\mathrm{min}}$ and $\mathrm{SNR}_{\mathrm{max}}$, and is invariant to the shape of $\mathrm{SNR}(t)$ between $t=0$ and $t=1$. More precisely, the \textit{noise schedule} function $\exp(-\gamma_{\boldsymbol{\eta}}(t))$ which maps the time variable $t \in [0,1]$ to the signal-to-noise ratio $\mathrm{SNR}(t)$ does not influence the diffusion loss integral in Equation~\ref{eq:snrminmax}, except for at its endpoints $\mathrm{SNR}_{\mathrm{max}}$ and $\mathrm{SNR}_{\mathrm{min}}$. Therefore, given $v$, the shape of the noise schedule function $\exp(-\gamma_{\boldsymbol{\eta}}(t))$ does not affect the diffusion loss.

Another way to understand the above result is by realizing that to compute the diffusion loss integral, it suffices to evaluate the antiderivative $F$ of the squared-error term at the endpoints $\mathrm{SNR}_{\mathrm{min}}$ and $\mathrm{SNR}_{\mathrm{max}}$:
\begin{align}
    & \mathcal{L}_\infty(\mathbf{x}) =
    \\ & -\frac{1} {2}\mathbb{E}_{\boldsymbol{\epsilon} \sim \mathcal{N}(0,\mathbf{I})}\left[\int_0^1\left\| \mathbf{x} - \hat{\mathbf{x}}_{\boldsymbol{\theta}}\left( \sigma_t\left(\mathbf{x}\sqrt{\mathrm{SNR}(t)} + \boldsymbol{\epsilon}\right) ;t\right) \right\|^2_2 \cdot \mathrm{SNR}'(t)\mathop{\mathrm{d}t}\right] 
    \\[5pt] & = -\frac{1} {2}\mathbb{E}_{\boldsymbol{\epsilon} \sim \mathcal{N}(0,\mathbf{I})}\left[\int_0^1 F'(\mathrm{SNR}(t)) \cdot \mathrm{SNR}'(t)\mathop{\mathrm{d}t} \right] 
    \\[5pt] & = -\frac{1} {2}\mathbb{E}_{\boldsymbol{\epsilon} \sim \mathcal{N}(0,\mathbf{I})}\left[\int_0^1 (F \circ \mathrm{SNR})' (t) \mathop{\mathrm{d}t} \right] 
    \\[5pt] & = \frac{1} {2}\mathbb{E}_{\boldsymbol{\epsilon} \sim \mathcal{N}(0,\mathbf{I})}\left[ -\left(F(\mathrm{SNR}(1)) - F(\mathrm{SNR}(0))\right) \right]
    \\[5pt] & = \frac{1} {2}\mathbb{E}_{\boldsymbol{\epsilon} \sim \mathcal{N}(0,\mathbf{I})}\left[ F(\mathrm{SNR}_\mathrm{max}) - F(\mathrm{SNR}_\mathrm{min}) \right],
\end{align}
since every continuous function has an antiderivative. Furthermore, there are infinitely many antiderivaties of the mean-square-error term, each of which, $G$, differs from $F$ by a only constant $c$:
\begin{align}
    G(v) \coloneqq \int \left\| \mathbf{x} - \hat{\mathbf{x}}_{\boldsymbol{\theta}}\left( \mathbf{z}_v;v\right) \right\|^2_2 \mathop{\mathrm{d}v} = F(v) + c,
\end{align}
for all signal-to-noise ratio functions $v \equiv \mathrm{SNR}(t)$.

\section{Weighted Diffusion Loss}
\label{subsubsec: weighted diffusion loss}
The diffusion objectives used in practice can be understood as a weighted version of the diffusion loss:
\begin{align}
    \mathcal{L}_\infty(\mathbf{x}, w) & = \frac{1} {2}\int_{\mathrm{SNR}_{\mathrm{min}}}^{\mathrm{SNR}_{\mathrm{max}}} w(v) \mathbb{E}_{\boldsymbol{\epsilon} \sim \mathcal{N}(0,\mathbf{I})}\left[\left\| \mathbf{x} - \hat{\mathbf{x}}_{\boldsymbol{\theta}}\left( \mathbf{z}_v;v\right) \right\|^2_2\right] \mathop{\mathrm{d}v},
    \\[5pt] & = -\frac{1}{2} \mathbb{E}_{\boldsymbol{\epsilon} \sim \mathcal{N}(0,\mathbf{I})}\left[ \int_{0}^{1} w(\mathrm{SNR}(t)) \mathrm{SNR}'(t) \left\| \mathbf{x} - \hat{\mathbf{x}}_{\boldsymbol{\theta}}\left( \mathbf{z}_t;t\right) \right\|^2_2 \mathop{\mathrm{d}t}\right], \customtag{recall Eq.~\ref{eq: dv_loss}}
\end{align}
where $w(v) = w(\mathrm{SNR}(t))$ is a chosen weighting function of the noise schedule. In intuitive terms, the weighting function stipulates the relative importance of each noise level prescribed by the noise schedule. Ideally, we would like to be able to adjust the weighting function such that the model focuses on modelling perceptually important information and ignoring imperceptible bits. From a Fourier analysis perspective~\citep{rissanen2023generative,kreis2022tutorial}, by encouraging our model to focus on some noise levels more than others using a weighting function, we are implicitly specifying a preference for modelling low, mid, and/or high-frequency details at different levels. 

When $w(v) = 1$, the diffusion objective is equivalent to maximizing the variational lower bound in Section~\ref{subsubsec: Variational Lower Bound: Top-down HVAE}. As detailed later in Section~\ref{subsec: Understanding Diffusion Objectives}, the invariance to the noise schedule property outlined in Section~\ref{subsubsec: Invariance to the Noise Schedule} still holds for weighted diffusion objectives.

In terms of noise prediction, following Equation~\ref{eq: ct_diff_loss}, the weighted diffusion objective becomes:
\begin{equation}
    \mathcal{L}_\infty(\mathbf{x}, w) = \frac{1} {2}\mathbb{E}_{\boldsymbol{\epsilon} \sim \mathcal{N}(0,\mathbf{I})}\left[\int_0^1 w(\mathrm{SNR}(t)) \gamma'_{\boldsymbol{\eta}}(t) \left\| \boldsymbol{\epsilon} - \hat{\boldsymbol{\epsilon}}_{\boldsymbol{\theta}}(\mathbf{z}_t;t) \right\|^2_2 \mathop{\mathrm{d}t} \right], \label{eq: weighted_ct_loss}
\end{equation}
where $w(\mathrm{SNR}(t)) = w(\mathrm{exp}(-\gamma_{\boldsymbol{\eta}}(t)))$, as per the definition of the (learned) noise schedule in Section~\ref{subsubsec: Noise Schedule}.

It turns out that the main difference between most diffusion model objectives boils down to the \textit{implied} weighting function $w(\mathrm{SNR}(t))$ being used~\citep{kingma2021variational,kingma2023understanding}. For instance, \cite{ho2020denoising,song2019generative,song2020improved,nichol2021improved} choose to minimize a so-called \textit{simple} objective of the form:
\begin{align}
    \mathcal{L}_{\infty\text{-}\mathrm{simple}}(\mathbf{x}) & \coloneqq \mathbb{E}_{\boldsymbol{\epsilon} \sim \mathcal{N}(0,\mathbf{I}), t \sim \mathcal{U}(0,1)} \left[ \left\| \boldsymbol{\epsilon} - \hat{\boldsymbol{\epsilon}}_{\boldsymbol{\theta}}(\mathbf{z}_t;t) \right\|^2_2 \right],
\end{align}
or the analogous discrete-time version
\begin{align}
    \mathcal{L}_{T\text{-}\mathrm{simple}}(\mathbf{x}) & \coloneqq \mathbb{E}_{\boldsymbol{\epsilon} \sim \mathcal{N}(0,\mathbf{I}), i \sim U\{1,T\}} \left[ \left\| \boldsymbol{\epsilon} - \hat{\boldsymbol{\epsilon}}_{\boldsymbol{\theta}}(\mathbf{z}_{t{(i)}};t{(i)}) \right\|^2_2 \right],
\end{align}
where $t(i) = i/T$ for $T$ . Contrasting the above with Equation~\ref{eq: weighted_ct_loss}, we can deduce that the $\mathcal{L}_{\infty\text{-}\mathrm{simple}}(\mathbf{x})$ objective above implies the following weighting function:
\begin{align}
    \mathcal{L}_\infty(\mathbf{x}, w) & = \frac{1} {2}\mathbb{E}_{\boldsymbol{\epsilon} \sim \mathcal{N}(0,\mathbf{I}),t \sim \mathcal{U}(0,1)}\left[ w(\mathrm{SNR}(t)) \gamma'_{\boldsymbol{\eta}}(t) \left\| \boldsymbol{\epsilon} - \hat{\boldsymbol{\epsilon}}_{\boldsymbol{\theta}}(\mathbf{z}_t;t) \right\|^2_2 \right]
    \\[5pt] & = \frac{1}{2} \mathbb{E}_{\boldsymbol{\epsilon} \sim \mathcal{N}(0,\mathbf{I}),t \sim \mathcal{U}(0,1)}\bigg[ \frac{1}{\gamma'_{\boldsymbol{\eta}}(t)} \gamma'_{\boldsymbol{\eta}}(t) \left\| \boldsymbol{\epsilon} - \hat{\boldsymbol{\epsilon}}_{\boldsymbol{\theta}}(\mathbf{z}_t;t) \right\|^2_2 \bigg] 
    \\[5pt] & = \frac{1}{2}\mathcal{L}_{\infty\text{-}\mathrm{simple}}(\mathbf{x}) \implies  w(\mathrm{SNR}(t)) = \frac{1}{\gamma'_{\boldsymbol{\eta}}(t)}.
\end{align}
It is worth restating that -- in contrast to VDMs -- the noise schedule specification in most commonly used diffusion models is fixed rather than learned from data, i.e. there are no learnable parameters $\boldsymbol{\eta}$.

Moreover, notice that~\cite{ho2020denoising}'s popular noise prediction objective is an implicitly defined \textit{weighted} objective in image space, where the weighting is a function of the signal-to-noise ratio:
\begin{align}
    \mathcal{L}_{\infty\text{-}\mathrm{simple}}(\mathbf{x}) & = \mathbb{E}_{\boldsymbol{\epsilon} \sim \mathcal{N}(0,\mathbf{I}), t \sim \mathcal{U}(0,1)} \left[ \left\| \boldsymbol{\epsilon} - \hat{\boldsymbol{\epsilon}}_{\boldsymbol{\theta}}(\mathbf{z}_t;t) \right\|^2_2 \right]
    \\[5pt] & = \mathbb{E}_{\boldsymbol{\epsilon} \sim \mathcal{N}(0,\mathbf{I}), t \sim \mathcal{U}(0,1)} \left[ \left\| \frac{\mathbf{z}_t - \alpha_t\mathbf{x}}{\sigma_t} - \frac{\mathbf{z}_t - \alpha_t\hat{\mathbf{x}}_{\boldsymbol{\theta}}(\mathbf{z}_t;t)}{\sigma_t} \right\|^2_2 \right]
    \\[5pt] & = \mathbb{E}_{\boldsymbol{\epsilon} \sim \mathcal{N}(0,\mathbf{I}), t \sim \mathcal{U}(0,1)} \left[ \frac{\alpha_t^2}{\sigma_t^2} \left\| \mathbf{x} - \hat{\mathbf{x}}_{\boldsymbol{\theta}}(\mathbf{z}_t;t) \right\|^2_2 \right]
    \\[5pt] & = \mathbb{E}_{\boldsymbol{\epsilon} \sim \mathcal{N}(0,\mathbf{I}), t \sim \mathcal{U}(0,1)} \left[ w(\mathrm{SNR}(t)) \left\| \mathbf{x} - \hat{\mathbf{x}}_{\boldsymbol{\theta}}(\mathbf{z}_t;t) \right\|^2_2 \right], 
\end{align}
recalling that $\mathbf{z}_t = \alpha_t \mathbf{x} + \sigma_t \boldsymbol{\epsilon}, \ \boldsymbol{\epsilon} \sim \mathcal{N}(0,\mathbf{I})$ by definition (ref. Section \ref{subsec: Gaussian Diffusion Process: Forward Time}). In this case, the implied weighting function of the noise schedule (in image space) is the identity: $w(\mathrm{SNR}(t)) = \mathrm{SNR}(t)$.
\section{Noise Schedule Density}
\label{subsubsec: noise schedule density}
To remain consistent with~\cite{kingma2023understanding}, let $\lambda = \log(\alpha_\lambda^2 / \sigma^2_\lambda) $ denote the logarithm of the signal-to-noise ratio function $\mathrm{SNR}(t)$, where $\alpha_\lambda^2 =  \mathrm{sigmoid}(\lambda_t)$ and $\sigma_\lambda^2 =  \mathrm{sigmoid}(-\lambda_t)$, for a timestep $t$. Let $f_\lambda : [0,1] \rightarrow \mathbb{R}$ denote the \textit{noise schedule} function, which maps from time $t \in [0,1]$ to the log-SNR $\lambda$, which we may explicitly denote by $\lambda_t$. Like before, the noise schedule function is monotonic thus invertible: $t = f^{-1}_\lambda(\lambda)$, and its endpoints are $\lambda_\mathrm{max} \coloneqq f_\lambda(0)$ and $\lambda_\mathrm{min} \coloneqq f_\lambda(1)$.

We can perform a \textit{change of variables} to define a probability density over noise levels:
\begin{align}
    p(\lambda) &= p_T(f^{-1}_\lambda(\lambda)) \left| \dv{f_\lambda^{-1}(\lambda)}{\lambda}(\lambda) \right| 
    \\[5pt] &= 1 \cdot \left| \dv{t}{\lambda}(\lambda) \right| \label{eq:pdfu} 
    \\[5pt] &= -\dv{t}{\lambda}(\lambda), \customtag{as $f_\lambda$ is monotonic} \customlabel{eq: plambda_density}
\end{align}
where $p_T = \mathcal{U}(0, 1)$ is a (continuous) uniform distribution over time, which we sample from during training $t \sim p_T$ to compute the log-SNR $\lambda = f_\lambda(t)$. In intuitive terms, the density $p(\lambda)$ describes the relative importance that the model assigns to different noise levels. Note that it can sometimes be beneficial to use different noise schedules for training and sampling~\citep{karras2022elucidating}. Since $f_\lambda$ is strictly monotonically decreasing in time and thus has negative slope, we can simplify the absolute value in Equation~\ref{eq:pdfu} with a negative sign to ensure the density $p(\lambda)$ remains positive.

Nothing that $\mathrm{SNR}(t) = e^\lambda$, the weighted diffusion objective can be trivially expressed in terms of $\lambda$ as:
\begin{align}
    \mathcal{L}_\infty(\mathbf{x}) & = -\frac{1} {2}\mathbb{E}_{\boldsymbol{\epsilon} \sim \mathcal{N}(0,\mathbf{I}),t \sim \mathcal{U}(0,1)}\left[\mathrm{SNR}'(t)\left\| \mathbf{x} - \hat{\mathbf{x}}_{\boldsymbol{\theta}}\left( \mathbf{z}_t;t\right) \right\|^2_2 \right]
    \\[5pt] & = -\frac{1} {2}\mathbb{E}_{\boldsymbol{\epsilon} \sim \mathcal{N}(0,\mathbf{I}),t \sim \mathcal{U}(0,1)}\left[e^{\lambda} \dv{\lambda}{t}\left\| \mathbf{x} - \hat{\mathbf{x}}_{\boldsymbol{\theta}}\left( \mathbf{z}_t;t\right) \right\|^2_2 \right] \customtag{chain rule}
    \\[5pt] & = -\frac{1}{2}\mathbb{E}_{\boldsymbol{\epsilon} \sim \mathcal{N}(0,\mathbf{I}), t \sim \mathcal{U}(0,1)} \left[ e^{\lambda} \dv{\lambda}{t} \left\| \frac{\mathbf{z}_t - \sigma_t\boldsymbol{\epsilon}}{\alpha_t} - \frac{\mathbf{z}_t - \sigma_t\hat{\boldsymbol{\epsilon}}_{\boldsymbol{\theta}}(\mathbf{z}_t;t)}{\alpha_t} \right\|^2_2 \right]
    \\[5pt] & = -\frac{1} {2}\mathbb{E}_{\boldsymbol{\epsilon} \sim \mathcal{N}(0,\mathbf{I}),t \sim \mathcal{U}(0,1)}\left[ e^{\lambda} \dv{\lambda}{t} \frac{\sigma_t^2}{\alpha_t^2}\left\| \boldsymbol{\epsilon} - \hat{\boldsymbol{\epsilon}}_{\boldsymbol{\theta}}(\mathbf{z}_t; \lambda_t) \right\|^2_2 \right]  
    \\[5pt] & = -\frac{1} {2}\mathbb{E}_{\boldsymbol{\epsilon} \sim \mathcal{N}(0,\mathbf{I}),t \sim \mathcal{U}(0,1)}\left[ \dv{\lambda}{t} \left\| \boldsymbol{\epsilon} - \hat{\boldsymbol{\epsilon}}_{\boldsymbol{\theta}}(\mathbf{z}_t; \lambda_t) \right\|^2_2 \right]. \customtag{since $e^\lambda = \alpha_t^2/\sigma_t^2$} \customlabel{eq: lambda_loss}
\end{align}
For complete clarity, the negative sign in front comes from the fact that $\lambda_t = -\gamma_{\boldsymbol{\eta}}(t)$ in the previous parameterization; so the negative sign in front of the original denoising objective in Equation~\ref{eq: dv_loss} no longer cancels out with the $-\gamma'_{\boldsymbol{\eta}}(t)$ term from the noise-prediction derivation in Equation~\ref{eq: negamma}.

As in Section~\ref{subsubsec: Invariance to the Noise Schedule}, we can perform a change of variables to transform our integral w.r.t. to time $t$ into an integral w.r.t. our new variable $\lambda$ -- the \textit{logarithm} of the signal-to-noise ratio:
\begin{align}
    &\mathcal{L}_\infty(\mathbf{x}) = -\frac{1} {2}\mathbb{E}_{\boldsymbol{\epsilon} \sim \mathcal{N}(0,\mathbf{I}),t \sim \mathcal{U}(0,1)}\left[ \dv{\lambda}{t} \left\| \boldsymbol{\epsilon} - \hat{\boldsymbol{\epsilon}}_{\boldsymbol{\theta}}(\mathbf{z}_t; \lambda_t) \right\|^2_2 \right]
    \\[5pt] & = -\frac{1} {2}\mathbb{E}_{\boldsymbol{\epsilon} \sim \mathcal{N}(0,\mathbf{I})}\left[\int_0^1\left\| \boldsymbol{\epsilon} - \hat{\boldsymbol{\epsilon}}_{\boldsymbol{\theta}}\left( \sigma_t\left(\mathbf{x}\sqrt{\mathrm{exp}(\lambda_t)} + \boldsymbol{\epsilon}\right) ;t\right) \right\|^2_2 \cdot \dv{\lambda}{t}\mathop{\mathrm{d}t}\right] 
    \\[5pt] & = -\frac{1} {2}\mathbb{E}_{\boldsymbol{\epsilon} \sim \mathcal{N}(0,\mathbf{I})}\left[\int_{f_\lambda(0)}^{f_\lambda(1)}\left\| \boldsymbol{\epsilon} - \hat{\boldsymbol{\epsilon}}_{\boldsymbol{\theta}}\left( \mathbf{z}_\lambda;\lambda\right) \right\|^2_2 \mathop{\mathrm{d}\lambda}\right]
    \\[5pt] & = \frac{1} {2}\mathbb{E}_{\boldsymbol{\epsilon} \sim \mathcal{N}(0,\mathbf{I})}\left[\int_{f_\lambda(1)}^{f_\lambda(0)}\left\| \boldsymbol{\epsilon} - \hat{\boldsymbol{\epsilon}}_{\boldsymbol{\theta}}\left( \mathbf{z}_\lambda;\lambda\right) \right\|^2_2 \mathop{\mathrm{d}\lambda}\right] \customtag{swap limits}
    \\[5pt] & = \frac{1}{2}\mathbb{E}_{\boldsymbol{\epsilon} \sim \mathcal{N}(0,\mathbf{I})}\left[\int_{\lambda_{\mathrm{min}}}^{\lambda_{\mathrm{max}}}\left\| \boldsymbol{\epsilon} - \hat{\boldsymbol{\epsilon}}_{\boldsymbol{\theta}}\left( \mathbf{z}_\lambda;\lambda\right) \right\|^2_2 \mathop{\mathrm{d}\lambda}\right]. \label{eq:logsnrminmax}
\end{align}

The weighted version of the objective is then simply
\begin{align}
    \mathcal{L}_w(\mathbf{x}) = \frac{1}{2}\mathbb{E}_{\boldsymbol{\epsilon} \sim \mathcal{N}(0,\mathbf{I})}\left[\int_{\lambda_{\mathrm{min}}}^{\lambda_{\mathrm{max}}} w(\lambda) \left\| \boldsymbol{\epsilon} - \hat{\boldsymbol{\epsilon}}_{\boldsymbol{\theta}}\left( \mathbf{z}_\lambda;\lambda\right) \right\|^2_2 \mathop{\mathrm{d}\lambda}\right], \label{eq: weighted_loss_lambda}
\end{align}
which once again shows that the diffusion loss integral does \textit{not} depend directly on the \textit{noise schedule} function $f_\lambda$ except for at its endpoints $\lambda_{\mathrm{min}}, \lambda_{\mathrm{max}}$; and through the choice of weighting function $w(\lambda)$. In other words, given the value of $\lambda$, the value of $t = f_\lambda^{-1}(\lambda)$ is simply irrelevant for evaluating the integral.

Therefore, the only meaningful difference between diffusion objectives is the choice of weighting function used~\citep{kingma2023understanding}.

\section{Importance Sampling Distribution}
Although the invariance to the noise schedule still holds under different weighting functions $w(\lambda)$ in Equation~\ref{eq: weighted_loss_lambda}, it does \textit{not} hold for the Monte Carlo estimator we use during training (e.g. Equation~\ref{eq: lambda_loss}), which is based on random samples from our distribution over the time variable $t \sim \mathcal{U}(0,1)$, and Gaussian noise distribution $\boldsymbol{\epsilon} \sim \mathcal{N}(0,\mathbf{I})$. Indeed, the choice of noise schedule affects the \textit{variance} of the Monte Carlo estimator of the diffusion loss. To demonstrate this fact, we first briefly review Importance Sampling (IS); which is a set of Monte Carlo methods used to estimate expectations under a \textit{target} distribution $p$ using a weighted average of samples from an \textit{importance} distribution $q$ of our choosing. 

Let $p(x)$ be a probability density for a random variable $X$, and $f(X)$ be some function we would like to compute the expectation of $\mu = \mathbb{E}_p\left[f(X)\right]$. The basic probability result of IS stipulates that whenever sampling from some target distribution $p(x)$ directly is inefficient or impossible (e.g. we only know $p(x)$ up to a normalizing constant), we can choose any density $q(x)$ to compute $\mu$:   
\begin{align}
    \mu = \int f(x)p(x) \mathop{\mathrm{d}x} = \int f(x) p(x) \frac{q(x)}{q(x)} \mathop{\mathrm{d}x} = \mathbb{E}_{q}\left[ \frac{p(X)}{q(X)}f(X)\right],
\end{align}
as long as $q(x) > 0$ whenever $f(x)p(x) \neq 0$. Concretely, we can estimate $\mu$ using samples from $q$:
\begin{align}
    &&\widehat{\mu} = \frac{1}{N} \sum_{i=1}^N \frac{p(X_i)}{q(X_i)}f(X_i), &&X_1,\dots,X_N \mathop{\sim}\limits^{\mathrm{iid}} q,&&
\end{align}
where, by the weak law of large numbers, $\widehat{\mu} \xrightarrow{P} \mu$ when $N \rightarrow \infty$.

Now, observe that it is possible to rewrite the weighted diffusion objective above (i.e. Equation~\ref{eq: lambda_loss}) such that the noise schedule density $p(\lambda)$ is revealed to be an \textit{importance sampling} distribution:
\begin{align}
    \mathcal{L}_w(\mathbf{x}) & = \frac{1} {2}\mathbb{E}_{\boldsymbol{\epsilon} \sim \mathcal{N}(0,\mathbf{I}),t \sim \mathcal{U}(0,1)}\left[ w(\lambda_t) \cdot -\dv{\lambda}{t} \cdot \left\| \boldsymbol{\epsilon} - \hat{\boldsymbol{\epsilon}}_{\boldsymbol{\theta}}(\mathbf{z}_t; \lambda) \right\|^2_2 \right] 
    \\[5pt] & = -\int_0^1 \left(\frac{1}{2}\int \left\| \boldsymbol{\epsilon} - \hat{\boldsymbol{\epsilon}}_{\boldsymbol{\theta}}(\mathbf{z}_t; \lambda) \right\|^2_2 p(\boldsymbol{\epsilon})\mathop{\mathrm{d}\boldsymbol{\epsilon}}\right) w(\lambda) \dv{\lambda}{t} \mathop{\mathrm{d}t}
    \\[5pt] & \eqqcolon -\int_0^1 h(t; \mathbf{x}) w(\lambda)\dv{\lambda}{t} \mathop{\mathrm{d}t} \customtag{define $h(\cdot)$ for brevity}
    \\[5pt] & = \int_{f_\lambda(1)}^{f_\lambda(0)} h(\lambda; \mathbf{x}) w(\lambda)\mathop{\mathrm{d}\lambda} \customtag{change-of-variables}
    \\[5pt] & = \int_{f_\lambda(1)}^{f_\lambda(0)} h(\lambda; \mathbf{x}) w(\lambda) \frac{p(\lambda)}{p(\lambda)} \mathop{\mathrm{d}\lambda} \customtag{introduce IS distribution}
    \\[5pt] & = \mathbb{E}_{\lambda \sim p(\lambda)}\left[ \frac{w(\lambda)}{p(\lambda)} h(\lambda;\mathbf{x})\right]
    \\[5pt] & = \mathbb{E}_{\lambda \sim p(\lambda)}\left[ \frac{w(\lambda)}{p(\lambda)} \mathbb{E}_{\boldsymbol{\epsilon} \sim \mathcal{N}(0,\mathbf{I})} \left[\frac{1}{2}\left\| \boldsymbol{\epsilon} - \hat{\boldsymbol{\epsilon}}_{\boldsymbol{\theta}}(\mathbf{z}_\lambda; \lambda) \right\|^2_2\right]\right]
    \\[5pt] & = \frac{1}{2}\mathbb{E}_{\boldsymbol{\epsilon} \sim \mathcal{N}(0,\mathbf{I}),\lambda \sim p(\lambda)}\left[ \frac{w(\lambda)}{p(\lambda)} \left\| \boldsymbol{\epsilon} - \hat{\boldsymbol{\epsilon}}_{\boldsymbol{\theta}}(\mathbf{z}_\lambda; \lambda) \right\|^2_2\right]. \label{eq: is_loss}
\end{align}
It is clear then that different choices of the noise schedule affect the variance of the Monte Carlo estimator of the diffusion loss because the noise schedule density $p(\lambda)$ acts as an importance sampling distribution. Importantly, judicious choices of the importance distribution can substantially increase the efficiency of Monte Carlo algorithms for numerically evaluating integrals.

A natural question to ask at this stage is how one may select $p(\lambda)$, such that a variance reduction is obtained. Variance reduction is obtained if and only if the difference between the variance of the original estimator $h(\lambda; \mathbf{x})$ and the importance sampling estimator $\hat{h}(\lambda; \mathbf{x}) \coloneqq h(\lambda;\mathbf{x})w(\lambda) / p(\lambda)$ is strictly positive. Formally, the following expression should evaluate to a value greater than $0$:
\begin{align}
    &\mathbb{V}_w(h(\lambda; \mathbf{x})) - \mathbb{V}_p(\hat{h}(\lambda; \mathbf{x})) = \nonumber
    \\[5pt] & = \int h^2(\lambda;\mathbf{x})w(\lambda)\mathop{\mathrm{d}\lambda} - \left(\int h(\lambda;\mathbf{x})w(\lambda) \mathop{\mathrm{d}\lambda}\right)^2
    \\[5pt] & \qquad - \left(\int \left(\frac{h(\lambda;\mathbf{x})w(\lambda)}{p(\lambda)}\right)^2 p(\lambda)\mathop{\mathrm{d}\lambda} - \left(\int \frac{h(\lambda;\mathbf{x})w(\lambda)}{p(\lambda)} p(\lambda) \mathop{\mathrm{d}\lambda}\right)^2\right)
    \\[5pt] & = \int h^2(\lambda;\mathbf{x})w(\lambda)\mathop{\mathrm{d}\lambda} - \left(\int h(\lambda;\mathbf{x})w(\lambda) \mathop{\mathrm{d}\lambda}\right)^2 
    \\[5pt] & \qquad - \int \hat{h}^2(\lambda; \mathbf{x}) p(\lambda)\mathop{\mathrm{d}\lambda} + \left(\int h(\lambda;\mathbf{x})w(\lambda) \mathop{\mathrm{d}\lambda}\right)^2 
    \\[5pt] & = \int h^2(\lambda;\mathbf{x})w(\lambda) -\frac{h^2(\lambda;\mathbf{x})w^2(\lambda)}{p(\lambda)} \mathop{\mathrm{d}\lambda}
    \customtag{substitute out $\hat{h}$}
    \\[5pt] & = \mathbb{E}_{w}\left[\left(1 -\frac{w(\lambda)}{p(\lambda)} \right)  h^2(\lambda;\mathbf{x}) \right],
\end{align}
revealing a concise expression that may be useful for practical evaluation. It is a well-known result that the optimal IS distribution is of the form $p^*(\lambda) \propto \lvert h(\lambda);\mathbf{x} \rvert w(\lambda)$, since it minimizes the variance of the IS estimator~\citep{wasserman2004all}. However, this result is mostly of theoretical interest rather then practical, as it requires knowledge of the integral we are aiming to estimate in the first place.

Furthermore, our current setting is somewhat different from the type of problem one would typically attack with importance sampling, as here we get to choose both distributions involved:
\begin{enumerate}[(i)]
    \item The weighting function $w(\lambda)$ acts as the \textit{target} distribution. It stipulates the relative importance of each noise level and ensures the model is focusing on perceptually important information. However, $w(\lambda)$ may not be a valid probability density as the most commonly used (implied) weighting functions do not integrate to 1 over their support.
    \item The \textit{importance} distribution is the noise schedule density $p(\lambda)$, which specifies the noise schedule of the Gaussian diffusion process.
\end{enumerate}
This means we technically ought to retune the noise schedule for different choices of weighting function. To avoid this,~\cite{kingma2023understanding} propose an adaptive noise schedule where:
\begin{align}
    p(\lambda) \propto \mathbb{E}_{\mathbf{x} \sim \mathcal{D}, \boldsymbol{\epsilon} \sim \mathcal{N}(0, \mathbf{I})} \left[ w(\lambda) \left\| \boldsymbol{\epsilon} - \hat{\boldsymbol{\epsilon}}_{\boldsymbol{\theta}}(\mathbf{z}_\lambda; \lambda) \right\|^2_2 \right],
\end{align}
thereby ensuring that the magnitude of the loss (Equation~\ref{eq: is_loss}) is approximately invariant to $\lambda$, and spread evenly across time $t$. This approach is often found to speed up optimization significantly.~\cite{song2021maximum,vahdat2021score} have also explored variance reduction techniques for diffusion models but from a score-based perspective.

\paragraph{Alternative Adaptive Weighting Methods.} \cite{karras2024analyzing} introduce a learning-based adaptation mechanism that balances gradient magnitudes, thereby changing the effective weighting of noise levels during training. Relatedly,~\cite{dieleman2022continuous} propose a technique they call \textit{time warping}, which uses a learnable function to convert loss values into an adaptive noise-level distribution that balances model capacity.~\cite{mueller2023continuous} extend time warping to heterogenous data, learning different noise level distributions for different data types. All the above mechanisms, among others~\citep{santos2023using,sabour2024align}, represent different ways to adapt the noise level weighting, and this topic remains a fertile ground for future work.
%
\section{ELBO with Data Augmentation}
\label{subsubsec: elbo with data aug}
In this section, we dissect the main result presented by~\cite{kingma2023understanding}; that when the weighting function of the diffusion loss is monotonic, the resulting objective is equivalent to the ELBO under simple data augmentation using Gaussian additive noise. We provide an instructive derivation of this result, discuss its implications, and discuss an extension to the setting of non-monotonic weighting functions.

The general goal is to inspect the behaviour of the weighted diffusion objective across time $t$, and manipulate the expression such that we end up with an expectation under a valid probability distribution specified by the weighting function $w(\lambda)$. This then allows us to examine the integrand and reveal that it corresponds to the expected negative ELBO of noise-perturbed data.

To that end, let $q(\mathbf{z}_{t:1} \mid \mathbf{x}) \coloneqq q(\mathbf{z}_{t},\mathbf{z}_{t + \mathop{\mathrm{d}t}},\dots,\mathbf{z}_{1} \mid \mathbf{x})$ denote the joint distribution of the posterior (forward process) for a subset of timesteps: $\{t,t + \mathop{\mathrm{d}t},\dots,1\}$, where $t > 0$ and $\mathop{\mathrm{d}t}$ denotes an infinitesimal change in time. Analogously, let $p(\mathbf{z}_{t:1})$ denote the prior (generative model) for the same subset of timesteps. 

The KL divergence of the joint posterior $q(\mathbf{z}_{t:1} \mid \mathbf{x})$ from the joint prior $p(\mathbf{z}_{t:1})$ is given by
\begin{align}
    \mathcal{L}(t;\mathbf{x}) &\coloneqq D_{\mathrm{KL}}(q(\mathbf{z}_{t:1} \mid \mathbf{x}) \parallel p(\mathbf{z}_{t:1}))
    \\[5pt] & = \frac{1} {2}\mathbb{E}_{\boldsymbol{\epsilon} \sim \mathcal{N}(0,\mathbf{I})}\left[-\int_{f_\lambda(t)}^{f_\lambda(1)} \left\| \boldsymbol{\epsilon} - \hat{\boldsymbol{\epsilon}}_{\boldsymbol{\theta}}(\mathbf{z}_\lambda; \lambda) \right\|^2_2 \mathop{\mathrm{d}\lambda} \right]. 
    \customtag{from Eq.~\ref{eq:logsnrminmax}}
\end{align}
Next, we rearrange and differentiate under the integral sign w.r.t. time $t$ to give:
\begin{align}
    \dv{\mathcal{L}(t;\mathbf{x})}{t} & = \dv{}{t} \left(\int_{f_\lambda(t)}^{f_\lambda(1)}-\frac{1}{2}\mathbb{E}_{\boldsymbol{\epsilon} \sim \mathcal{N}(0,\mathbf{I})}\left[\left\| \boldsymbol{\epsilon} - \hat{\boldsymbol{\epsilon}}_{\boldsymbol{\theta}}(\mathbf{z}_\lambda; \lambda) \right\|^2_2 \right] \mathop{\mathrm{d}\lambda} \right)
    \\[5pt] & \eqqcolon \dv{}{t} \left(\int_{f_\lambda(t)}^{f_\lambda(1)}h(\lambda;\mathbf{x}) \mathop{\mathrm{d}\lambda} \right)
    \\[5pt] & = \dv{}{t} \Big[F(f_\lambda(1)) - F(f_\lambda(t)) \Big]\customtag{$F$ is an antiderivative of $h$}
    \\[5pt] & = 0 - F'(f_\lambda(t)) \cdot f'_\lambda(t) \customtag{chain rule}
    \\[5pt] & = -F'(\lambda) \cdot \dv{\lambda}{t} \customtag{recall $\lambda = f_\lambda(t)$}
    \\[5pt] & = \frac{1}{2}\dv{\lambda}{t} \mathbb{E}_{\boldsymbol{\epsilon} \sim \mathcal{N}(0,\mathbf{I})}\left[\left\| \boldsymbol{\epsilon} - \hat{\boldsymbol{\epsilon}}_{\boldsymbol{\theta}}(\mathbf{z}_\lambda; \lambda) \right\|^2_2 \right], \customtag{since $F'(\lambda) = h(\lambda;\mathbf{x})$}
\end{align}
which allows us to rewrite the weighted diffusion objective by substituting in the above result:
\begin{align}
    \mathcal{L}_w(\mathbf{x}) & = -\frac{1} {2}\mathbb{E}_{\boldsymbol{\epsilon} \sim \mathcal{N}(0,\mathbf{I}),t \sim \mathcal{U}(0,1)}\left[ w(\lambda_t) \cdot \dv{\lambda}{t} \cdot \left\| \boldsymbol{\epsilon} - \hat{\boldsymbol{\epsilon}}_{\boldsymbol{\theta}}(\mathbf{z}_t; \lambda) \right\|^2_2 \right] 
    \\[5pt] & = \mathbb{E}_{t \sim \mathcal{U}(0,1)}\left[ w(\lambda_t) \cdot -\frac{1} {2} \dv{\lambda}{t} \mathbb{E}_{\boldsymbol{\epsilon} \sim \mathcal{N}(0,\mathbf{I})}\left[\left\| \boldsymbol{\epsilon} - \hat{\boldsymbol{\epsilon}}_{\boldsymbol{\theta}}(\mathbf{z}_t; \lambda) \right\|^2_2 \right] \right]
    \\[5pt] & = \mathbb{E}_{t \sim \mathcal{U}(0,1)}\left[-\dv{\mathcal{L}(t;\mathbf{x})}{t} w(\lambda_t) \right]. \customtag{time derivative}
\end{align}
After some simple manipulation, we can see that the resulting expression is an expectation of the time derivative of the joint KL divergence $\mathcal{L}(t;\mathbf{x})$, weighted by the weighting function $w(\lambda_t)$. This result is not particularly interesting or surprising by itself, but it enables the next step; using integration by parts to turn the above expression into an expectation under a valid probability distribution specified by the weighting function. Recall that the formula for integration by parts is given by:
\begin{align}
   \int_a^b u(t) v'(t) \mathop{\mathrm{d}t} = u(b) v(b) - u(a) v(a) - \int_a^b u'(t) v(t) \mathop{\mathrm{d}t}.
 \end{align}
Setting $u(t) = w(\lambda_t)$ and $v'(t) = \mathrm{d}/\mathrm{d}t \ \mathcal{L}(t;\mathbf{x})$ then gives:
\begin{align}
    \mathcal{L}_w(\mathbf{x}) & = \int_0^1 -\dv{\mathcal{L}(t;\mathbf{x})}{t} w(\lambda_t) \mathop{\mathrm{d}t}
    \\[5pt] & = -\left(w(\lambda_1)\mathcal{L}(1;\mathbf{x})\mathop{-} w(\lambda_0)\mathcal{L}(0;\mathbf{x}) - \int_0^1 \dv{w(\lambda_t)}{t}\mathcal{L}(t;\mathbf{x})  \mathop{\mathrm{d}t} \right)
    \\[5pt] & = \int_0^1 \dv{w(\lambda_t)}{t}\mathcal{L}(t;\mathbf{x})  \mathop{\mathrm{d}t} \mathop{+} w(\lambda_0)\mathcal{L}(0;\mathbf{x})\mathop{-} w(\lambda_1)\mathcal{L}(1;\mathbf{x}) 
    \\[5pt] & = \int_0^1 \dv{w(\lambda_t)}{t}\mathcal{L}(t;\mathbf{x})  \mathop{\mathrm{d}t} \mathop{+} c, \customtag{absorb constants into $c$} \customlabel{eq: time_derivative_c}
\end{align}
where $c$ is a small constant for two simple reasons: 
\begin{enumerate}[(i)]
    \item $w(\lambda_0)\mathcal{L}(0;\mathbf{x}) = w(\lambda_{\mathrm{max}})D_{\mathrm{KL}}(q(\mathbf{z}_{0:1} \mid \mathbf{x}) \parallel p(\mathbf{z}_{0:1}))$ is small due to the weighting function acting on $\lambda_{\mathrm{max}}$ always being very small by construction~\citep{kingma2023understanding};
    \item $\mathop{-} w(\lambda_1)\mathcal{L}(1;\mathbf{x}) = \mathop{-} w(\lambda_{\mathrm{min}})D_{\mathrm{KL}}(q(\mathbf{z}_{1} \mid \mathbf{x}) \parallel p(\mathbf{z}_{1}))$ includes the KL between the posterior of the noisiest latent $\mathbf{z}_1$ and the prior, which is both independent of the parameters $\boldsymbol{\theta}$ of the model $\hat{\boldsymbol{\epsilon}}_{\boldsymbol{\theta}}(\mathbf{z}_t; \lambda)$, and very close to $0$ for a well-specified forward diffusion process.
\end{enumerate}

The astute reader may notice that the derivative term $\mathrm{d}/\mathrm{d}t \ w(\lambda_t)$ in Equation~\ref{eq: time_derivative_c} is a valid probability density function (PDF) specified by the weighting function, so long as $w(\lambda_t)$ is monotonically increasing w.r.t. time $t$, and $w(\lambda_{t=1}) = 1$. The proof is straightforward: by the Fundamental Theorem of Calculus, the PDF $f(x)$ of a random variable $X$ is obtained by differentiating the cumulative distribution function (CDF) $F(x)$, that is: $f(x) = \mathrm{d}/\mathrm{d}x \ F(x)$, where $F : \mathbb{R} \rightarrow [0,1]$, $\lim_{x\rightarrow-\infty}F(x)=0$ and $\lim_{x\rightarrow\infty}F(x)=1$. 

Therefore, in our context, $w$ is a valid CDF if it satisfies the following three standard conditions:
\begin{flalign}
    \qquad \mathrm{(i)} \ \ & w : \mathbb{R} \to [0, 1] 
    \qquad &\\[3pt] \mathrm{(ii)} \ \ & t > t - \mathrm{d}t \implies w(\lambda_t) \geq w(\lambda_{t - \mathrm{d}t}), \ \forall t \in [0,1] 
    \qquad &\\[3pt] \mathrm{(iii)} \ \ & \lim\limits_{t \to 0} w(\lambda_t) = 0, \ \mathrm{and} \ \lim\limits_{t \to 1} w(\lambda_t) = 1.
\end{flalign}
In words, $w$ must: (i) map the real line to $[0,1]$, (ii) be non-decreasing w.r.t. time $t$, and (iii) be normalized w.r.t. time $t$.

If the above conditions hold, we can define a valid probability distribution $p_w(t)$ specified by the weighting function:
\begin{align}
    &&p_w(t) \coloneqq \dv{w(\lambda_t)}{t}, && \mathrm{where} && w(\lambda_t) = \int_0^{\lambda_t} p_w(t) \mathop{\mathrm{d}t},&&
\end{align}
with support on the range $[0,1]$, thus $\int_0^1 p_w(t) \mathop{\mathrm{d}t} = 1$.

This then permits us to rewrite the diffusion loss as an expectation under $p_w(t)$ by substituting:
\begin{align}
    \mathcal{L}_w(\mathbf{x}) & = \int_0^1 \dv{w(\lambda_t)}{t}\mathcal{L}(t;\mathbf{x})  \mathop{\mathrm{d}t} \mathop{+} c \customtag{from Eq.~\ref{eq: time_derivative_c}}
    \\[5pt] & = \mathbb{E}_{t \sim p_w(t)}\left[\mathcal{L}(t;\mathbf{x})\right] \mathop{+} c. \label{eq: pwt_expected_loss}
\end{align}

The final move is to show that the joint KL divergence $\mathcal{L}(t;\mathbf{x})$ for any subset of timesteps $\{t, t + \mathop{\mathrm{d}t}, \dots, 1\}$ decomposes into the expected negative ELBO of noisy data $\mathbf{z}_t \sim q(\mathbf{z}_t \mid \mathbf{x})$ at any timestep $t$:
\begin{align}
    &\mathcal{L}(t;\mathbf{x}) = D_{\mathrm{KL}}(q(\mathbf{z}_{t:1} \mid \mathbf{x}) \parallel p(\mathbf{z}_{t:1}))
    \\[5pt] & = \int q(\mathbf{z}_{t:1} \mid \mathbf{x}) \log \frac{q(\mathbf{z}_{t:1} \mid \mathbf{x})}{p(\mathbf{z}_{t:1})}\mathop{\mathrm{d}\mathbf{z}_{t:1}}
    \\[5pt] & = \int q(\mathbf{z}_{t} \mid \mathbf{x})q(\mathbf{z}_{t + \mathrm{d}t:1} \mid \mathbf{x}) \log \frac{q(\mathbf{z}_{t} \mid  \mathbf{x})q(\mathbf{z}_{t + \mathrm{d}t:1} \mid \mathbf{x})}{p(\mathbf{z}_{t} \mid \mathbf{z}_{t+\mathrm{d}t})p(\mathbf{z}_{t+\mathrm{d}t:1})}\mathop{\mathrm{d}\mathbf{z}_{t:1}} 
    \\[5pt] & = \mathbb{E}_{q(\mathbf{z}_{t} \mid  \mathbf{x})} \left[\mathbb{E}_{q(\mathbf{z}_{t + \mathrm{d}t:1} \mid \mathbf{x})} \left[ \log \frac{q(\mathbf{z}_{t + \mathrm{d}t:1} \mid \mathbf{x})}{p(\mathbf{z}_{t+\mathrm{d}t:1})} - \log p(\mathbf{z}_{t} \mid \mathbf{z}_{t+\mathrm{d}t}) \right]\right] 
    \nonumber
    \\[5pt] & \qquad + \mathbb{E}_{q(\mathbf{z}_{t} \mid \mathbf{x})} \left[ \log q(\mathbf{z}_{t} \mid \mathbf{x}) \right] 
    \customtag{constant entropy term $\mathcal{H}(\cdot)$}
    \\[5pt] & = \mathbb{E}_{q(\mathbf{z}_{t} \mid  \mathbf{x})} \Big[\mathbb{E}_{q(\mathbf{z}_{t + \mathrm{d}t} \mid \mathbf{x})} \left[-\log p(\mathbf{z}_{t} \mid \mathbf{z}_{t+\mathrm{d}t})\right] 
    \nonumber
    \\[5pt] & \qquad
    + D_\mathrm{KL}(q(\mathbf{z}_{t + \mathrm{d}t:1} \mid \mathbf{x}) \parallel p(\mathbf{z}_{t+\mathrm{d}t:1})) \Big] - \mathcal{H}(q(\mathbf{z}_{t} \mid \mathbf{x})) 
    \\[5pt] & = \mathbb{E}_{q(\mathbf{z}_{t} \mid  \mathbf{x})} \left[ -\mathrm{ELBO}_t (\mathbf{z}_t)\right] - \mathcal{H}(q(\mathbf{z}_{t} \mid \mathbf{x})).
\end{align}
As shown, factoring the joint distributions into infinitesimal transitions between $\mathbf{z}_t$ and $\mathbf{z}_{t + \mathrm{d}t}$ reveals an expected variational free energy term (negative ELBO), which is an upper bound on the negative log-likelihood of noisy data: $-\mathrm{ELBO}_t(\mathbf{z}_t) \geq -\log p(\mathbf{z}_t)$, where $\mathbf{z}_t \sim q(\mathbf{z}_t \mid \mathbf{x})$ for any timestep $t$. The entropy term $\mathcal{H}(q(\mathbf{z}_{t} \mid \mathbf{x}))$ is constant since our forward process is fixed, i.e. it is a Gaussian diffusion.

Finally, substituting the above result into the (expected) weighted diffusion loss gives
\begin{align}
    \mathcal{L}_w(\mathbf{x}) & = \mathbb{E}_{p_w(t)} \left[\mathcal{L}(t;\mathbf{x})\right] + c \customtag{from Equation~\ref{eq: pwt_expected_loss}}
    \\[5pt] & = \mathbb{E}_{p_w(t)} \left[\mathbb{E}_{q(\mathbf{z}_{t} \mid \mathbf{x})} \left[ -\mathrm{ELBO}_t (\mathbf{z}_t)\right] - \mathcal{H}(q(\mathbf{z}_{t} \mid \mathbf{x}))\right] + c \customtag{substitute}
    \\[5pt] & = -\mathbb{E}_{p_w(t),q(\mathbf{z}_{t} \mid  \mathbf{x})} \left[ \mathrm{ELBO}_t (\mathbf{z}_t)\right] + c \customtag{absorb entropy constant}
    \\[5pt] & \geq -\mathbb{E}_{p_w(t),q(\mathbf{z}_{t} \mid  \mathbf{x})} \left[ \log p(\mathbf{z}_t) \right] + c, \customtag{noisy data log-likelihood}
\end{align}
which proves that when the weighting function $w(\lambda_t)$ is monotonically increasing w.r.t. time $t$, diffusion objectives are equivalent to the ELBO under simple data augmentation using Gaussian additive noise. To be clear, the Gaussian additive noise comes from the fact that the forward diffusion specification is linear Gaussian, and as such, each $\mathbf{z}_t \sim q(\mathbf{z}_t \mid \mathbf{x})$ is simply a noisy version of the data $\mathbf{x}$. The distribution $p_w(t)$ acts as a sort of data augmentation kernel, specifying the importance of different noise levels. It is worth noting that this type of data augmentation setup resembles distribution augmentation (DistAug) and distribution smoothing methods~\citep{meng2020improved,jun2020distribution}, which have previously been shown to improve the sample quality of autoregressive generative models.

Now going back to Equation~\ref{eq: time_derivative_c}, we see that the diffusion loss is a weighted integral of ELBOs:
\begin{align}
    \mathcal{L}_w(\mathbf{x}) & = \int_0^1 \mathcal{L}(t;\mathbf{x}) \dv{w(\lambda_t)}{t} \mathop{\mathrm{d}t} \mathop{+} c 
    \\[5pt] & = \int_0^1 \mathbb{E}_{q(\mathbf{z}_t \mid \mathbf{x})}\left[-\mathrm{ELBO}_t(\mathbf{z}_t)\right]  \mathop{\mathrm{d}w(\lambda_t)} \mathop{+} c, \label{eq: weighted_integral_elbo}
\end{align}
since $\mathcal{L}(t;\mathbf{x})$ equates to the expected negative ELBO for noise-perturbed data $\mathbf{z}_t \sim q(\mathbf{z}_t \mid \mathbf{x})$ as explained above, and the $\mathop{\mathrm{d}w(\lambda_t)}$ term simply weights the ELBO at each noise level.

\subsection{Non-monotonic Weighting Functions}
Several works have observed impressive synthesis results when using non-monotonic weighting functions~\citep{nichol2021improved,karras2022elucidating,choi2022perception,hang2023efficient}. What are the theoretical implications of using such weighting functions? 

Looking again at Equation~\ref{eq: weighted_integral_elbo}, we observe that regardless of the weighting function, diffusion objectives boil down to a weighted integral of ELBOs. However, if the weighting function is non-monotonic (thus $w$ is not a valid CDF) then the derivative term $\mathrm{d}/\mathrm{d}t \ w(\lambda_t)$ will be negative for some points in time, meaning we end up \textit{minimizing} the ELBO at those noise levels rather than maximizing it! This is somewhat inconvenient in light of the practical success of non-monotonic weighting functions and seems to reaffirm the widespread belief that maximum likelihood may not be the appropriate objective for generating high-quality samples. One sensible explanation for this success is that the non-monotonic weighting functions sacrifice some modes of the likelihood in exchange for better perceptual synthesis. Indeed, the majority of bits in images are allocated to imperceptible details and can, therefore, be largely ignored if what we care about is perceptual quality. This aspect was briefly touched upon by~\cite{ho2020denoising}; they found that although their diffusion models were not competitive with the state-of-the-art likelihood-based models in terms of lossless codelengths, the samples were of high quality, nonetheless.

With that said, \cite{kingma2023understanding} showed that contrary to popular belief, likelihood maximization (i.e. maximizing the ELBO) and high-quality image synthesis are not mutually exclusive in diffusion models. They were able to achieve state-of-the-art FID~\citep{heusel2017gans}/Inception~\citep{salimans2016improved} scores on the high-resolution ImageNet benchmark using \textit{monotonic} weighting functions and some practical/architectural improvements proposed by~\cite{pmlr-v202-hoogeboom23a}. For a more detailed characterisation of commonly used weighting functions and whether they are monotonic we encourage the reader to refer to \cite{kingma2023understanding} and \cite{hoogeboom2024simpler}. To conclude, as we have learned, optimizing the weighted diffusion loss with a monotonic weighting function is equivalent to maximizing the ELBO under simple data augmentation using Gaussian additive noise.
\chapter{Discussion \& Outlook}
\label{sec: discussion_outlook}
Despite the growing interest in diffusion models, gaining a deep understanding of the model class remained somewhat elusive for the uninitiated in non-equilibrium statistical physics. With that in mind, we have synthesized a holistic view of diffusion models using only directed graphical modelling and variational inference principles, which is simple and imposes relatively fewer prerequisites on the reader.

\paragraph{The Top-down Perspective.} We began by reviewing hierarchical latent variable models and established a unifying graphical modelling-based perspective on their connection with diffusion models. We showed that diffusion models share a specific \textit{top-down} latent variable hierarchy structure with ladder networks~\citep{valpola2015neural} and top-down inference HVAEs~\citep{sonderby2016ladder}, which explains why they share optimization objectives.
Although introducing additional latent variables significantly improves the flexibility of both the inference and generative models, it comes with additional challenges. We highlighted the difficulties with using purely \textit{bottom-up} inference procedures in deep latent variable hierarchies, including \textit{posterior collapse} for instance, whereby the posterior distribution (of the top-most layer, say) may collapse to a standard Gaussian prior, failing to learn meaningful representations and deactivating latent variables.~\citet{burda2015importance,sohl2015deep} point to the asymmetry between the generative and inference models in bottom-up HVAEs as a source of difficulty in training the inference model efficiently, as there is no way to express each term in the VLB as an expectation under a distribution over a single variable.~\citet{luo2022understanding,bishop2023} echo similar efficiency-based arguments against optimizing the VLB under bottom-up inference in hierarchical latent variable models.

We showed that efficiency-based perspectives paint an incomplete picture; the main reason one ought to avoid bottom-up inference in hierarchical latent variable models is the lack of direct \textit{feedback} from the generative model. We showed that since the purpose of the inference model is to perform \textit{Bayesian inference} at any given layer in the hierarchy, interleaving feedback from each transition in the generative model into each respective transition in the inference model makes both procedures more accurate. Although this may not benefit diffusion models to the same extent as HVAEs, as the inference model of the former is fixed, the top-down hierarchy structure is nonetheless ubiquitous. 

In diffusion models the Markovian transitions between latent states $q(\mathbf{z}_t \mid \mathbf{z}_{t-1})$ are chosen to be linear Gaussian with isotropic covariances, thus the top-down posterior $q(\mathbf{z}_{t-1} \mid \mathbf{z}_{t}, \mathbf{x})$ is tractable through Gaussian conjugacy and the KL divergence terms in the VLB simplify significantly to squared-error terms (Section~\ref{subsubsec: deriving dkl}). Since the Gaussian diffusion process can be defined directly in terms of the conditionals $q(\mathbf{z}_t \mid \mathbf{x})$ (Section~\ref{subsec: Gaussian Diffusion Process: Forward Time}), it is possible to: (i) train any level of the latent variable hierarchy independently; (ii) share the same denoising model across the whole hierarchy. These are critical advantages over ladder networks and top-down HVAEs, as they both induce hierarchical dependencies between the latent states, preventing the training of individual layers independently. This advantage is particularly salient for infinitely deep latent variable hierarchies induced by diffusion as detailed in Section~\ref{sec: variational diffusion models}. 

\textit{In summary, the top-down perspective offers an intuitive understanding of diffusion models as a specific instantiation of ladder networks and hierarchical VAEs with top-down inference models.}

\paragraph{Addressing the W[H]ole Problem with VAEs.} We highlight that a major problem with (hierarchical) VAEs, which is \textit{not} present in diffusion models, is the \textit{hole problem}. The hole problem refers to the mismatch between the aggregate posterior and the prior. As shown in Figure~\ref{fig: hole}, there can be regions with high probability density under the prior which have low probability density under the aggregate posterior. This affects the quality of generated samples, as the decoder may receive latents sampled from regions not covered by the training data. Furthermore, the higher the dimensionality of our data, the less likely it is that our finite dataset covers the entirety of the input space. The manifold hypothesis posits that high-dimensional datasets lie along a much lower-dimensional latent manifold. However, since providing latent variable identifiability guarantees is challenging for most problems~\citep{hyvarinen2024identifiability}, in practice, we often resort to unfalsifiable assumptions about both the functional form and dimensionality of the latent space.

\textit{Diffusion models cleverly circumvent both of the aforementioned issues} by: (i) defining the aggregate posterior to be equal to the prior by construction; and (ii) sacrificing the ability to learn \textit{representations} by fixing the posterior distribution to follow a pre-determined noise schedule. Point (i) ensures a smooth transition between the prior $p(\mathbf{z}_T)$ and the model $p(\mathbf{x} \mid \mathbf{z}_1)$, thereby avoiding drift at inference time. Point (ii) entails defining the latent variables $\mathbf{z}_{1:T}$ as simple noisy versions of the input rather than learnable representations. The added noise can be interpreted as a kind of data augmentation technique, which helps smooth out the data density landscape and connect distant modes of the underlying data distribution.

\paragraph{An Inductive Bias for Perceptual Quality.} As explained in Section~\ref{subsec: Gaussian Diffusion Process: Forward Time}, the noise schedule is specified by parameters $\alpha_t$, $\sigma^2_t$ and in combination with a weighting function $w(\alpha^2_t/\sigma^2_t)$, stipulates the relative importance of each noise level in the diffusion objective (Section~\ref{subsubsec: weighted diffusion loss}). \textit{It turns out that the main difference between most diffusion model objectives boils down to the \text{implied} weighting function being used}~\citep{kingma2023understanding}. If the primary goal is high-quality synthesis, it suffices to adjust the noise schedule and weighting function such that the model focuses on perceptually important information. Indeed, the majority of bits in images are allocated to imperceptible details which can in principle be ignored. Moreover, by encouraging the model to focus on some noise levels more than others, we are implicitly prescribing a preference for modelling low, mid, and/or high-frequency details at different noise levels. Relatedly,~\citet{ho2020denoising} found that although their diffusion models were not competitive with state-of-the-art likelihood-based models in terms of lossless codelengths (i.e. compression ability), the samples generated were of high quality nonetheless. \textit{This demonstrates that diffusion models possess excellent inductive biases for image data.}

\paragraph{Why Diffusion Works.} We argue that the success of diffusion models can be partly attributed to an additional reduction in \textit{degrees-of-freedom} compared to top-down HVAEs. In VAEs, several simplifying assumptions are needed to make the inference problem tractable and scalable: (i) amortized posterior; (ii) mean-field variational family; (iii) parametric assumptions on $p(\mathbf{x}, \mathbf{z})$; (iv) a Monte Carlo estimator of the VLB. Under the top-down perspective established in Section~\ref{sec: variational diffusion models}, it is easy to see that diffusion models constitute yet another simplifying assumption by fixing the inference distribution. This transforms the learning problem from minimizing the \textit{reverse} KL divergence to improve our posterior approximation $q$ where the prior $p$ may be fixed: 
\begin{align}
    \argmin_{q \in \mathcal{Q}}D_{\mathrm{KL}}( q(\mathbf{z}_{1:T} \mid \mathbf{x}) \parallel p(\mathbf{z}_{1:T})),    
\end{align}
to minimizing the \textit{forward} KL where the posterior $q$ is fixed: 
\begin{align}
    &\argmin_{p \in \mathcal{P}} D_{\mathrm{KL}}( q(\mathbf{z}_{1:T} \mid \mathbf{x}) \parallel p(\mathbf{z}_{1:T}))
    \\[-10pt] &\qquad = \argmin_{p \in \mathcal{P}} \mathbb{E}_{q(\mathbf{z}_{1:T} \mid \mathbf{x})} \left[-\log p(\mathbf{z}_{1:T}) \right] - \overbrace{\mathcal{H}(q(\mathbf{z}_{1:T}\mid \mathbf{x})),}^{\mathrm{constant}}
    \label{eq: erm}
\end{align}
which amounts to a \textit{supervised learning} problem under noise-augmented data, optimized via maximum likelihood. Unsupervised learning tries to represent \textit{all} the information about $p(\mathbf{x})$, which includes imperceptible details and complicates the learning problem. Conversely, supervised learning is effective at filtering out unnecessary information for the task at hand, which in combination with carefully weighted diffusion objectives, helps explain how/why diffusion models are capable of high-quality image synthesis that better aligns with human perception. Generally speaking, having a learnable posterior $q$ can destabilize HLVM training by creating asynchronous learning dynamics with the prior/model $p$, often leading to posterior collapse due to a constantly shifting target. Diffusion models avoid this by simply holding $q$ fixed, thereby providing a fixed target for $p$ to match during training. Relatedly, the scalability of simple $L2$-like diffusion objectives also plays a role in its current practical success compared to other HLVM objectives.

In summary, when one's goal is merely high-quality synthesis, it is generally advantageous to sacrifice representation learning ability by fixing the posterior $q$ and leveraging the tried-and-tested machinery of supervised learning to train a good generative model. \textit{Interestingly, this supervised learning rationale was the motivation behind ladder networks~\citep{valpola2015neural}, and we argue that it provides an intuitive perspective on the success of diffusion models.}

\paragraph{Is Maximum Likelihood the Right Objective?} In Section~\ref{subsec: Understanding Diffusion Objectives}, we provided a thorough explanation of the various weighted diffusion objectives in the literature. By analyzing weighted diffusion objectives, it is possible to show that they are equivalent to the ELBO under simple data augmentation using Gaussian additive noise~\citep{kingma2023understanding}, so long as the weighting function $w(\lambda_t)$ is \textit{monotonically} increasing w.r.t. time $t$. In Section~\ref{subsec: Understanding Diffusion Objectives} we provide an instructive derivation of this result and show that it holds if $w$ is a valid CDF. Multiple recent works report impressive image synthesis results using \textit{non-monotonic} weighting functions~\citep{nichol2021improved, karras2022elucidating}, which somewhat peculiarly implies that the ELBO is being \textit{minimized} at certain noise levels. This seems to reaffirm the widespread belief that maximum likelihood may \textit{not} be the right objective to use for high-quality image synthesis. However,~\citet{kingma2023understanding} suggest that likelihood maximization needs \textit{not} be intrinsically at odds with high-quality image synthesis, as they were able to achieve state-of-the-art FID scores on the high-resolution ImageNet benchmark by optimizing the ELBO under simple data augmentation.

\newpage 
In light of these results, we further stress that studying the connection between weighted diffusion objectives and maximum likelihood (i.e. the ELBO) is particularly important. This is because we know from Shannon's \textit{source coding theorem} that the average codelength of the optimal compression scheme is the entropy of the data $\mathbb{\mathcal{H}}(X) = \mathbb{E}[-\log p(\mathbf{x})]$, and as long as we are minimizing codelengths given by the information content $-\log p_{\boldsymbol{\theta}}(\mathbf{x})$ defined by a probabilistic model $p_{\boldsymbol{\theta}}$ (e.g. by maximizing likelihood w.r.t. $\boldsymbol{\theta}$), then the resulting average codelength $\mathbb{E}[-\log p_{\boldsymbol{\theta}}(\mathbf{x})]$ approaches the entropy of the true data distribution. \textit{This is the fundamental goal of both generative modelling and compression, a goal with which maximum likelihood learning is well aligned.}

\paragraph{Closing Remarks.} The success of diffusion models is arguably as much a product of collective engineering effort and scale as it is a product of algorithmic and theoretical insight. Nonetheless, identifying analogies between model classes undoubtedly aids in understanding, and recognizing the unique properties of specific models helps refine our intuitions about what may or may not work in the future. Fertile ground for future work includes: (i) enabling diffusion models to learn meaningful representations, perhaps with inspiration from predecessors like ladder networks~\citep{valpola2015neural} and denoising autoencoders~\citep{vincent2008extracting}; (ii) expanding the family of forward diffusion processes beyond linear Gaussian transitions with isotropic covariances, as this should tighten the VLB; and (iii) studying the identifiability guarantees provided by deterministic diffusion modelling (e.g. probability flow ODEs~\citep{song2021scorebased}) for causal representation learning.

\begin{acknowledgements}
We would like to thank Charles Jones, Rajat Rasal and Avinash Kori for fruitful discussions and valuable feedback. The authors were supported by funding from the ERC under the EU’s Horizon 2020 research and innovation programme (Project MIRA, grant No. 757173), and acknowledge support from EPSRC for the Causality in Healthcare AI Hub (grant number EP/Y028856/1).
\end{acknowledgements}

\appendix
\chapter{Notation \& Extras}
\label{App}

\section{Notation}
\begin{longtable}{rp{0.58\textwidth}c}
        \textsc{Symbol} & \textsc{Description} & \textsc{Section} 
        \\
        \midrule
        \\[-10pt]
        $\mathbf{x}$ & Observed datapoint, e.g. input image & \S\ref{sec: introduction} 
        \\[5pt]
        $t$ & Time index variable $t \in \{1,2,\dots,T\}$, or $t \in [0,1]$ for continuous-time & \S\ref{subsec: Hierarchical VAE}
        \\[20pt]
        $\mathbf{z}_t$ & Latent variable at time $t$
        & \S\ref{subsec: Hierarchical VAE} 
        \\[5pt]
        $\mathbf{z}_{1:T}$ & Finite set of latent variables representing $\mathbf{z}_1, \mathbf{z}_2,\dots,\mathbf{z}_T$ & \S\ref{subsec: Hierarchical VAE} 
        \\[20pt]
        $\mathbf{z}_{0:1}$ & Set of latent variables in continuous-time from $t=0$ to $t=1$ & \S\ref{subsec: Gaussian Diffusion Process: Forward Time} 
        \\[20pt]
        $\alpha_t$ & Noise schedule coefficient $\alpha_t \in (0, 1)$ & \S\ref{subsec: Gaussian Diffusion Process: Forward Time} 
        \\[5pt]
        $\sigma^2_t$ & Noise schedule variance $\sigma^2_t \in (0, 1)$ & \S\ref{subsec: Gaussian Diffusion Process: Forward Time} 
        \\[5pt]
        $\boldsymbol{\epsilon}_t$ & Isotropic random noise, $\boldsymbol{\epsilon}_t \sim \mathcal{N}(0,\mathbf{I})$ & \S\ref{sec: introduction}
        \\[5pt]
        $\mathrm{SNR}(t)$ & Signal-to-noise ratio (SNR) function at time $t$, defined as $\alpha^2_t / \sigma^2_t$ & \S\ref{subsubsec: Noise Schedule}
        \\[20pt]
        $q(\mathbf{z}_t \mid \mathbf{x})$ & Latent variable distribution given $\mathbf{x}$ & \S\ref{subsec: Gaussian Diffusion Process: Forward Time} 
        \\[5pt]
        $q(\mathbf{z}_t \mid \mathbf{z}_s)$ & Transition distribution from time $s$ to time $t$, where $s < t$ & \S\ref{subsubsec: lgt} 
        \\[20pt]
        $\alpha_{t|s}$ & Transition coefficient from time $s$ to $t$ & \S\ref{subsubsec: lgt} 
        \\[5pt]
        $\sigma^2_{t|s}$ & Variance of transition distribution & \S\ref{subsubsec: lgt} 
        \\[5pt]
        $q(\mathbf{z}_s \mid \mathbf{z}_t, \mathbf{x})$ & Top-down posterior distribution at time $s$ & \S\ref{subsec: Top-down Inference Model}
        \\[5pt]
        $\boldsymbol{\mu}_Q(\mathbf{z}_t, \mathbf{x};s,t)$ & Mean of top-down posterior distribution at time $s$; $\boldsymbol{\mu}_Q$ for short & \S\ref{subsubsec: qzs} 
        \\[20pt]
        $\sigma^2_Q(s,t)$ & Variance of top-down posterior distribution; $\sigma^2_Q$ for short & \S\ref{subsubsec: qzs} 
        \\[20pt]
        $p(\mathbf{z}_s \mid \mathbf{z}_t)$ & Generative transition distribution defined as $q(\mathbf{z}_s \mid \mathbf{z}_t, \mathbf{x}=\hat{\mathbf{x}}_{\boldsymbol{\theta}}(\mathbf{z}_t, t))$ & \S\ref{subsec: Discrete-time Generative Model} 
        \\[20pt]
        $p(\mathbf{x} \mid \mathbf{z}_0)$ & Observation likelihood (e.g. input image), analogous to $p(\mathbf{x} \mid \mathbf{z}_1)$ in discrete-time & \S\ref{subsec: Discrete-time Generative Model}
        \\[20pt]
        $\boldsymbol{\phi}$ & Variational parameters related to $q_{\boldsymbol{\phi}}$ & \S\ref{sec: introduction} 
        \\[5pt]
        $\boldsymbol{\theta}$ & Model parameters pertaining to $p_{\boldsymbol{\theta}}$ & \S\ref{sec: introduction}
        \\[5pt]
        $\hat{\mathbf{x}}_{\boldsymbol{\theta}}(\mathbf{z}_t, t)$ & Denoising model mapping any $\mathbf{z}_t$ to $\mathbf{x}$ & \S\ref{subsubsec: Deriving p} 
        \\[5pt] $\hat{\boldsymbol{\epsilon}}_{\boldsymbol{\theta}}(\mathbf{z}_t, t)$ & Noise prediction model, which approximates $\nabla_{\mathbf{z}_t} \log q(\mathbf{z}_t)$ & \S\ref{subsubsec: Deriving p} 
        \\[20pt]
        $\hat{\mathbf{s}}_{\boldsymbol{\theta}}(\mathbf{z}_t, t)$ & Score prediction model, equivalent to $-\hat{\boldsymbol{\epsilon}}_{\boldsymbol{\theta}}(\mathbf{z}_t, t) / \sigma_t$ & \S\ref{subsubsec: Deriving p} 
        \\[20pt]
        $\boldsymbol{\mu}_{\boldsymbol{\theta}}(\mathbf{z}_t;s,t)$ & Predicted posterior mean at time $s < t$
        & \S\ref{subsubsec: Deriving p} 
        \\[5pt]
        $\mathrm{VLB}(\mathbf{x})$ & Single-datapoint variational lower bound; also denoted as $\mathrm{ELBO}(\mathbf{x})$ & \S\ref{sec: introduction}
        \\[20pt]
        $\mathcal{L}_T(\mathbf{x})$ & Discrete-time diffusion loss & \S\ref{subsubsec: Variational Lower Bound: Top-down HVAE} 
        \\[5pt]
        $\mathcal{L}_{\infty}(\mathbf{x})$ & Continuous-time diffusion loss & \S\ref{subsubsec: on infinite depth}
        \\[5pt]
        $\mathcal{L}_w(\mathbf{x})$ & Weighted diffusion loss; also $\mathcal{L}_{\infty}(\mathbf{x}, w)$ & \S\ref{subsubsec: weighted diffusion loss}
        \\[5pt]
        $\boldsymbol{\gamma}_{\boldsymbol{\eta}}(t)$ & Neural network with parameters $\boldsymbol{\eta}$ for learning the noise schedule & \S\ref{subsubsec: Noise Schedule} \\[20pt]
        $w(\cdot)$ & Noise level weighting function & \S\ref{subsubsec: weighted diffusion loss} 
        \\[5pt]
        $\lambda$ & Logarithm of $\mathrm{SNR}(t)$; also $\lambda_t$ & \S\ref{subsubsec: noise schedule density} 
        \\[5pt]
        $f_\lambda(t)$ & Noise schedule function, mapping $t$ to $\lambda$ & \S\ref{subsubsec: noise schedule density} 
        \\[5pt]
        $\lambda_{\mathrm{min}}$ & Lowest log SNR given by $f_\lambda(t=1)$ & \S\ref{subsubsec: noise schedule density} 
        \\[5pt]
        $\lambda_{\mathrm{max}}$ & Highest log SNR given by $f_\lambda(t=0)$ & \S\ref{subsubsec: noise schedule density} 
        \\[5pt]
        $p(\lambda)$ & Density over noise levels & \S\ref{subsubsec: noise schedule density} 
        \\[5pt]
        $\mathcal{L}(t;\mathbf{x})$ & Joint KL divergence up to time $t$ & \S\ref{subsubsec: elbo with data aug} 
        \\[5pt]
        $p_w(t)$ & Augmentation kernel specified by $w(\cdot)$ & \S\ref{subsubsec: elbo with data aug}
\end{longtable}
\renewcommand{\thefootnote}{}
\vspace{-4pt}
\footnote{To remain consistent with prior work and avoid notational clutter, we may use the same symbols to denote random variables and their outcomes whenever our intentions can be clearly understood from context.}
\renewcommand{\thefootnote}{\arabic{footnote}}

\newpage
\section{Learning the Noise Schedule}
\label{subsubsec: Noise Schedule}
Perturbing data with multiple noise scales and choosing an appropriate \textit{noise schedule} is instrumental to the success of diffusion models. The noise schedule of the forward process is typically pre-specified and has no learnable parameters, however, VDMs learn the noise schedule via the parameterization:
\begin{align}
    \sigma_t^2 = \mathrm{sigmoid}\left(\gamma_{\boldsymbol{\eta}}(t)\right),
\end{align}
where $\gamma_{\boldsymbol{\eta}}(t)$ is a \textit{monotonic} neural network comprised of linear layers with weights $\boldsymbol{\eta}$ restricted to be positive. A monotonic function is a function defined on a subset of the real numbers which is either entirely non-increasing or entirely non-decreasing. As explained later, the noise schedule can be conveniently parameterized in terms of the signal-to-noise ratio. The signal-to-noise ratio (SNR) is defined as $\mathrm{SNR}(t) = \alpha_t^2 / \sigma_t^{2}$, and since $\mathbf{z}_t$ grow noisier over time we have that: $\mathrm{SNR}(t) < \mathrm{SNR}(s)$ for any $t > s$. 

For now, we provide some straightforward derivations of the expressions for $\alpha_t^2$ and $\mathrm{SNR}(t)$ as a function of $\gamma_{\boldsymbol{\eta}}(t)$. Recall that in a variance-preserving diffusion process $\alpha_t^2 = 1 - \sigma_t^2$, therefore:
\begin{align}
     \alpha_t^2 &= 1- \sigma_t^2 
     \\[5pt] & = 1 - \mathrm{sigmoid}\left(\gamma_{\boldsymbol{\eta}}(t)\right) 
     \\[5pt] \implies \alpha_t^2 & = \mathrm{sigmoid}\left(-\gamma_{\boldsymbol{\eta}}(t)\right),
\end{align}
as for an input $x \in \mathbb{R}$ the following holds 
\begin{align}
     1 - \mathrm{sigmoid}\left(x\right) &= 1 - \frac{1}{1 + e^{-x}} 
     \\[5pt] & = \frac{1 + e^{-x}}{1 + e^{-x}} - \frac{1}{1 + e^{-x}} 
     \\[5pt] & = \frac{e^{-x}}{1 + e^{-x}} \cdot \frac{e^{x}}{e^{x}} 
     \\[5pt] &
     = \mathrm{sigmoid}\left(-x\right).
\end{align}
To derive $\mathrm{SNR}(t)$ as a function of $\gamma_{\boldsymbol{\eta}}(t)$, we simply substitute in the above equations and simplify:
\begin{align}
    \mathrm{SNR}(t) &= \frac{\alpha_t^2}{\sigma_t^2} = \frac{\mathrm{sigmoid}\left(-\gamma_{\boldsymbol{\eta}}(t)\right)}{\mathrm{sigmoid}\left(\gamma_{\boldsymbol{\eta}}(t)\right)} \customtag{by definition}
    \\[5pt] &= \frac{(1+e^{\gamma_{\boldsymbol{\eta}}(t)})^{-1}}{(1+e^{-\gamma_{\boldsymbol{\eta}}(t)})^{-1}} 
    \\[5pt] &
    = \frac{1+e^{-\gamma_{\boldsymbol{\eta}}(t)}}{1+e^{\gamma_{\boldsymbol{\eta}}(t)}} 
    \\[5pt] &
    = \frac{\frac{e^{\gamma_{\boldsymbol{\eta}}(t)}}{e^{\gamma_{\boldsymbol{\eta}}(t)}}+\frac{1}{e^{\gamma_{\boldsymbol{\eta}}(t)}
    }}{1+e^{\gamma_{\boldsymbol{\eta}}(t)}} \cdot \frac{e^{\gamma_{\boldsymbol{\eta}}(t)}}{e^{\gamma_{\boldsymbol{\eta}}(t)}} 
    \\[5pt] & = \frac{e^{\gamma_{\boldsymbol{\eta}}(t)}+1}{e^{\gamma_{\boldsymbol{\eta}}(t)}(1+e^{\gamma_{\boldsymbol{\eta}}(t)})}
    \\[5pt] &
    = \frac{1}{e^{\gamma_{\boldsymbol{\eta}}(t)}}, \label{eq: snrt}
\end{align}
which is equivalently expressed as $\mathrm{SNR}(t) = \mathrm{exp}(-
\gamma_{\boldsymbol{\eta}}(t))$.
\section{Numerically Stable Primitives} 
\label{app: Simplified Expressions for the Generative Transitions}
The closed-form expressions for the mean and variance of $p(\mathbf{z}_s \mid \mathbf{z}_t)$ can be further simplified to include more numerically stable functions like $\mathrm{expm1}(\cdot) = \exp(\cdot) - 1$, which are available in standard numerical packages. The resulting simplified expressions -- which we derive in detail next -- enable more numerically stable implementations as highlighted by~\cite{kingma2021variational}.

Recall from Section~\ref{subsubsec: Noise Schedule} that the noise schedule parameters are given by: $\sigma_t^2 = \mathrm{sigmoid}(\gamma_{\boldsymbol{\eta}}(t))$, and $\alpha_t^2 = \mathrm{sigmoid}(-\gamma_{\boldsymbol{\eta}}(t))$, for any $t$. For brevity, let $s$ and $t$ be shorthand notation for $\gamma_{\boldsymbol{\eta}}(s)$ and $\gamma_{\boldsymbol{\eta}}(t)$ respectively. The posterior variance simplifies to:
\begin{align}
    \sigma_Q^2(s,t) & = \frac{\sigma_{t|s}^2\sigma_s^2}{\sigma^2_t} = \frac{\sigma_s^2\left(\sigma_{t}^2 - \frac{\alpha_t^2}{\alpha_s^2}\sigma_s^2\right)}{\sigma^2_t} \\[5pt] & = \frac{\frac{1}{1+e^{-s}} \cdot \left(\frac{1}{1+e^{-t}} - \frac{(1+e^{t})^{-1}}{(1+e^s)^{-1}}\cdot\frac{1}{1+e^{-s}}\right)}{\frac{1}{1+e^{-t}}} 
    \customtag{cancel denominator}
    \\[5pt] & = \left(1 + e^{-t}\right) \cdot \frac{1}{1+e^{-s}} \cdot \left(\frac{1}{1+e^{-t}} - \frac{1+e^s}{1+e^t}\cdot\frac{1}{1+e^{-s}}\right) 
    \customtag{distribute $1 + e^{-t}$} \customlabel{eq: distribute}
    \\[5pt] & = \frac{1}{1+e^{-s}} \cdot \left(1 - \frac{1+e^s}{1+e^t}\cdot\frac{1 + e^{-t}}{1+e^{-s}}\right)
    \\[5pt] & = \frac{1}{1+e^{-s}} \cdot \left(1 - \frac{e^s\left(1+e^{-s}\right)} {1+e^t}\cdot\frac{e^{-t}\left(1+e^{t}\right)}{1+e^{-s}}\right) 
    \customtag{cancel common factors}
    \\[5pt] & = \frac{1}{1+e^{-s}} \cdot \left(1 -e^{s-t}\right) \label{eq: 1me}
    \\[5pt] & = \sigma_s^2 \cdot \left( -\mathrm{expm1}\left(\gamma_{\boldsymbol{\eta}}(s) -\gamma_{\boldsymbol{\eta}}(t)\right)\right).
    \customtag{$\mathrm{expm1}(\cdot) = \exp(\cdot) - 1$}
\end{align}
The posterior mean -- under a noise-prediction model $\hat{\boldsymbol{\epsilon}}_{\boldsymbol{\theta}}(\mathbf{z}_t;t)$ -- simplifies in a similar fashion to:
\begin{align}
    &\boldsymbol{\mu}_{\boldsymbol{\theta}}(\mathbf{z}_t;s,t) = \frac{1}{\alpha_{t|s}}\mathbf{z}_t - \frac{\sigma^2_{t|s}}{\alpha_{t|s}\sigma_{t}}\hat{\boldsymbol{\epsilon}}_{\boldsymbol{\theta}}(\mathbf{z}_t;t)
    \\[5pt] & = \frac{\alpha_s}{\alpha_t} \left( \mathbf{z}_t - \frac{\sigma^2_{t|s}}{\sigma_t}\hat{\boldsymbol{\epsilon}}_{\boldsymbol{\theta}}(\mathbf{z}_t;t)\right)
    \\[5pt] & = \frac{\alpha_s}{\alpha_t} \left( \mathbf{z}_t - \frac{\sigma_t^2 - \frac{\alpha_t^2}{\alpha_s^2}\sigma_s^2}{\sigma_t}\hat{\boldsymbol{\epsilon}}_{\boldsymbol{\theta}}(\mathbf{z}_t;t)\right) 
    \customtag{substituting $\sigma_{t|s}^2 = \sigma_t^2 - \alpha_{t|s}^2\sigma_s^2$}
    \\[5pt] & = \frac{\alpha_s}{\alpha_t} \left( \mathbf{z}_t - \frac{\frac{1}{1+e^{-t}} - \frac{1+e^s}{1+e^t} \cdot \frac{1}{1+e^{-s}}}{\sqrt{\frac{1}{1+e^{-t}}}}\hat{\boldsymbol{\epsilon}}_{\boldsymbol{\theta}}(\mathbf{z}_t;t)\right)
    \\[5pt] & = \frac{\alpha_s}{\alpha_t} \Bigg( \mathbf{z}_t - (1+e^{-t}) \cdot \sqrt{\frac{1}{1+e^{-t}}} 
    \\[5pt] & \qquad\qquad\qquad \cdot  \left(\frac{1}{1+e^{-t}} - \frac{1+e^s}{1+e^t} \cdot \frac{1}{1+e^{-s}}\right)\hat{\boldsymbol{\epsilon}}_{\boldsymbol{\theta}}(\mathbf{z}_t;t)\Bigg) \label{eq: big_eq}
    \\[5pt] & = \frac{\alpha_s}{\alpha_t} \left( \mathbf{z}_t - \sigma_t \left(1 - e^{s-t}\right)\hat{\boldsymbol{\epsilon}}_{\boldsymbol{\theta}}(\mathbf{z}_t;t)\right)
    \\[5pt] & = \frac{\alpha_s}{\alpha_t} \left( \mathbf{z}_t + \sigma_t \mathrm{expm1}\left(\gamma_{\boldsymbol{\eta}}(s) -\gamma_{\boldsymbol{\eta}}(t) \right)\hat{\boldsymbol{\epsilon}}_{\boldsymbol{\theta}}(\mathbf{z}_t;t)\right),
\end{align}
where Equation~\ref{eq: big_eq} simplifies significantly via the same logical steps in Equations~\ref{eq: distribute}-\ref{eq: 1me} above.
\subsection{Numerically Stable Loss Estimator}
\label{app: Numerically stable Estimator}
The estimator of the discrete-time diffusion loss can be made more numerically stable in practice by re-expressing the constant term inside the expectation using more numerically stable primitives. Specifically:
\begin{align}
    \frac{\mathrm{SNR}(s)}{\mathrm{SNR}(t)} - 1 & = \frac{\alpha_s^2}{\sigma_s^2} \olddiv \frac{\alpha_t^2}{\sigma_t^2} - 1
    \\[5pt] & = 
    \frac{\alpha_s^2\sigma_t^2}{\alpha_t^2\sigma_s^2} - 1
    \\[5pt] & = 
    \frac{\mathrm{sigmoid}(-\gamma_{\boldsymbol{\eta}}(s))\cdot \mathrm{sigmoid}(\gamma_{\boldsymbol{\eta}}(t))}{\mathrm{sigmoid}(-\gamma_{\boldsymbol{\eta}}(t))\cdot \mathrm{sigmoid}(\gamma_{\boldsymbol{\eta}}(s))} - 1,
\end{align}
letting $s$ and $t$ denote $\gamma_{\boldsymbol{\eta}}(s)$ and $\gamma_{\boldsymbol{\eta}}(t)$ for brevity we have:
\begin{align}
    \frac{\frac{1}{1+e^s} \cdot \frac{1}{1+e^{-t}}}{\frac{1}{1+e^t} \cdot \frac{1}{1+e^{-s}}} - 1 & = \frac{\left(1 + e^t\right)\left(1 + e^{-s}\right)}{\left(1 + e^{s}\right)\left(1 + e^{-t}\right)} - 1
    \\[5pt] & = \frac{e^t\left(1 + e^{-t}\right)e^{-s}\left(1 + e^{s}\right)}{\left(1 + e^{s}\right)\left(1 + e^{-t}\right)} - 1
    \\[5pt] & = e^te^{-s} - 1  
    \\[5pt] & = \mathrm{expm1}\left(\gamma_{\boldsymbol{\eta}}(t) - \gamma_{\boldsymbol{\eta}}(s) \right). \label{eq: e_constant}
\end{align}
Substituting the above back into the (noise-prediction-based) diffusion loss estimator gives:
\begin{align}
    \mathcal{L}_T(\mathbf{x}) &= \frac{T}{2}\mathbb{E}_{\boldsymbol{\epsilon} \sim \mathcal{N}(0,\mathbf{I}),i \sim U{\{1,T\}}} \Big[
    \\ & \qquad \qquad \mathrm{expm1}\left(\gamma_{\boldsymbol{\eta}}(t) - \gamma_{\boldsymbol{\eta}}(s) \right)\left\| \boldsymbol{\epsilon} - \hat{\boldsymbol{\epsilon}}_{\boldsymbol{\theta}}(\mathbf{z}_t;t) \right\|^2_2 \Big],
\end{align}
which is the final form of the objective we wanted to show.
\subsection{Dealing with Edge Effects}
\label{app: Dealing with Edge Effects}
There is an edge effect at diffusion time $t = 0 $, possibly causing numerical issues~\citep{sohl2015deep,song2021maximum}, which we can avoid by setting the likelihood term to:
\begin{align}
    p(\mathbf{x} \mid \mathbf{z}_1) = \frac{q(\mathbf{z}_1 \mid \mathbf{x}) p(\mathbf{x})}{p(\mathbf{z}_1)},
\end{align}
and removing it from the variational lower bound. In discrete-time, this looks like:
\begin{align}
    \mathrm{VLB} & = \mathbb{E}_{q(\mathbf{z}_{1:T}, \mathbf{x})} \left[\log \frac{p(\mathbf{x} \mid \mathbf{z}_1)}{q(\mathbf{z}_1 \mid \mathbf{x})} + \log p(\mathbf{z}_T) + \sum_{t=2}^T \log \frac{p(\mathbf{z}_{t-1} \mid \mathbf{z}_t)}{q(\mathbf{z}_{t} \mid \mathbf{z}_{t-1})} \right]
    \\[5pt] &= \cancel{\mathbb{E}_{q(\mathbf{z}_{1}, \mathbf{x})} \left[ \log \frac{q(\mathbf{z}_1 \mid \mathbf{x}) p(\mathbf{x})}{q(\mathbf{z}_1 \mid \mathbf{x})p(\mathbf{z}_1)} \right]} 
    \\[5pt] & \qquad\qquad
    + \mathbb{E}_{q(\mathbf{z}_{1:T}, \mathbf{x})} \left[ \log p(\mathbf{z}_T) + \sum_{t=2}^T \log \frac{p(\mathbf{z}_{t-1} \mid \mathbf{z}_t)}{q(\mathbf{z}_{t} \mid \mathbf{z}_{t-1})} \right].
\end{align}
The left-hand side (LHS) term above cancels out as the SNR $ \to \infty $ (i.e. $\alpha_1 \to 1$ and $\sigma_1 \to 0$) since the least noisy latent variable $ \mathbf{z}_1 = \alpha_1 \mathbf{x} + \sigma_1 \boldsymbol{\epsilon}$ approaches $ \mathbf{x} $, meaning $ p(\mathbf{z}_1) \approx p(\mathbf{x})$. Note that $p$ in $ p(\mathbf{z}_1) $ and $ p(\mathbf{x}) $ above refers to the (tractable) prior distribution of choice.

For continuous-time diffusion where $T \to \infty$ and $t \in [0,1]$, we have that learning a model $ p(\mathbf{z}_{0}) $ is practically equivalent to learning a model $ p(\mathbf{x}) $ since $ \mathbf{z}_0 $ (the least noisy latent variable) is almost identical to $ \mathbf{x} $ in the limit given large enough log-SNR $ \lambda_{\text{max}} = \log(\alpha_0^2/\sigma_0^2)$. However, if one chooses to learn the noise schedule rather than fixing it, the $p(\mathbf{x} \mid \mathbf{z}_0)$ term may need to be incorporated back into the VLB objective, representing a final discrete step from latent space to image space. This manifests as some variation of a decoding step in both VDMs~\citep{kingma2021variational} and score-based diffusion models~\citep{song2021maximum}.
\section{Equivalence of Diffusion Specifications} 
\cite{kingma2021variational} elaborate on the equivalence of diffusion noise-schedule specifications using the following straightforward example. Firstly, the change of variables we used implies that $\sigma_v$ is given by:
\begin{align}
    v = \frac{\alpha^2_v}{\sigma^2_v} &\implies \sqrt{v} = \frac{\alpha_v}{\sigma_v} \implies \sigma_v = \frac{\alpha_v}{\sqrt{v}},
\end{align}
therefore, $\mathbf{z}_v$ can be equivalently expressed as
\begin{align}
    \mathbf{z}_v & = \alpha_v \mathbf{x} + \sigma_v \boldsymbol{\epsilon} 
    = \alpha_v \mathbf{x} + \frac{\alpha_v}{\sqrt{v}} \boldsymbol{\epsilon} 
    = \alpha_v \left( \mathbf{x} + \frac{\boldsymbol{\epsilon}}{\sqrt{v}}\right),
    \label{eq: equivv}
\end{align}
which holds for any diffusion specification (forward process) by definition. Now, consider two distinct diffusion specifications denoted as $\left\{\alpha^A_v, \sigma^A_v, \tilde{\mathbf{x}}^A_{\boldsymbol{\theta}}\right\}$ and $\left\{\alpha^B_v, \sigma^B_v, \tilde{\mathbf{x}}^B_{\boldsymbol{\theta}}\right\}$. Due to Equation~\ref{eq: equivv}, any two diffusion specifications produce equivalent latents, up to element-wise rescaling:
\begin{align}
    \mathbf{z}^A_v &= \frac{\alpha_v^A}{\alpha_v^B} \mathbf{z}^B_v
    \\[5pt] \alpha^A_v \left( \mathbf{x} + \frac{\boldsymbol{\epsilon}}{\sqrt{v}}\right) &= \frac{\alpha_v^A}{\alpha_v^B} \alpha^B_v \left( \mathbf{x} + \frac{\boldsymbol{\epsilon}}{\sqrt{v}}\right).
\end{align}
This implies that we can denoise from any latent $\mathbf{z}^B_v$ using a model $\tilde{\mathbf{x}}_{\boldsymbol{\theta}}^A$ trained under a different noise specification, by trivially rescaling the latent $\mathbf{z}^B_v$ such that it'd be equivalent to denoising from $\mathbf{z}^A_v$:
\begin{align}
    \tilde{\mathbf{x}}_{\boldsymbol{\theta}}^B\left(\mathbf{z}^B_v, v \right) \equiv \tilde{\mathbf{x}}_{\boldsymbol{\theta}}^A\left(\frac{\alpha_v^A}{\alpha_v^B} \mathbf{z}^B_v, v \right).
\end{align}
Furthermore, when two diffusion specifications have equal $\mathrm{SNR}_{\mathrm{min}}$ and $\mathrm{SNR}_{\mathrm{max}}$, then the marginal distributions $p^A(\mathbf{x})$ and $p^B(\mathbf{x})$ defined by the two generative models are equal:
\begin{align}
    \tilde{\mathbf{x}}_{\boldsymbol{\theta}}^B\left(\mathbf{z}^B_v, v \right) \equiv \tilde{\mathbf{x}}_{\boldsymbol{\theta}}^A\left(\frac{\alpha_v^A}{\alpha_v^B} \mathbf{z}^B_v, v \right) \implies p^A(\mathbf{x}) = p^B(\mathbf{x}),
\end{align}
and both specifications yield identical diffusion loss in continuous time: $\mathcal{L}^A_\infty(\mathbf{x}) = \mathcal{L}^B_\infty(\mathbf{x})$, due to Equation~\ref{eq:snrminmax}. Importantly, this does \textit{not} mean that training under different noise specifications will result in the same model. To be clear, the $\tilde{\mathbf{x}}_{\boldsymbol{\theta}}^B$ model is fully determined by the $\tilde{\mathbf{x}}_{\boldsymbol{\theta}}^A$ model and the rescaling operation $\alpha_v^A/\alpha_v^B$. Furthermore, this invariance to the noise schedule does not hold for the Monte Carlo estimator of the diffusion loss, as the noise schedule affects the \textit{variance} of the estimator and therefore affects optimization efficiency.

\backmatter  

\printbibliography

\begin{thebibliography}{73}
\providecommand{\natexlab}[1]{#1}
\providecommand{\url}[1]{\texttt{#1}}
\expandafter\ifx\csname urlstyle\endcsname\relax
  \providecommand{\doi}[1]{doi: #1}\else
  \providecommand{\doi}{doi: \begingroup \urlstyle{rm}\Url}\fi

\bibitem[Albergo and Vanden-Eijnden(2023)]{albergo2023building}
Michael~Samuel Albergo and Eric Vanden-Eijnden.
\newblock Building normalizing flows with stochastic interpolants.
\newblock In \emph{The Eleventh International Conference on Learning Representations}, 2023.

\bibitem[Anderson(1982)]{anderson1982reverse}
Brian~DO Anderson.
\newblock Reverse-time diffusion equation models.
\newblock \emph{Stochastic Processes and their Applications}, 12\penalty0 (3):\penalty0 313--326, 1982.

\bibitem[Bishop and Bishop(2023)]{bishop2023}
Christopher~M Bishop and Hugh Bishop.
\newblock \emph{Deep Learning: Foundations and Concepts}.
\newblock Springer, 2023.

\bibitem[Blei et~al.(2017)Blei, Kucukelbir, and McAuliffe]{blei2017variational}
David~M Blei, Alp Kucukelbir, and Jon~D McAuliffe.
\newblock Variational inference: A review for statisticians.
\newblock \emph{Journal of the American statistical Association}, 112\penalty0 (518):\penalty0 859--877, 2017.

\bibitem[Burda et~al.(2015)Burda, Grosse, and Salakhutdinov]{burda2015importance}
Yuri Burda, Roger Grosse, and Ruslan Salakhutdinov.
\newblock Importance weighted autoencoders.
\newblock \emph{arXiv preprint arXiv:1509.00519}, 2015.

\bibitem[Child(2020)]{child2020very}
Rewon Child.
\newblock Very deep vaes generalize autoregressive models and can outperform them on images.
\newblock In \emph{International Conference on Learning Representations}, 2020.

\bibitem[Choi et~al.(2022)Choi, Lee, Shin, Kim, Kim, and Yoon]{choi2022perception}
Jooyoung Choi, Jungbeom Lee, Chaehun Shin, Sungwon Kim, Hyunwoo Kim, and Sungroh Yoon.
\newblock Perception prioritized training of diffusion models.
\newblock In \emph{Proceedings of the IEEE/CVF Conference on Computer Vision and Pattern Recognition}, pages 11472--11481, 2022.

\bibitem[De~Sousa~Ribeiro et~al.(2023)De~Sousa~Ribeiro, Xia, Monteiro, Pawlowski, and Glocker]{pmlr-v202-de-sousa-ribeiro23a}
Fabio De~Sousa~Ribeiro, Tian Xia, Miguel Monteiro, Nick Pawlowski, and Ben Glocker.
\newblock High fidelity image counterfactuals with probabilistic causal models.
\newblock In \emph{Proceedings of the 40th International Conference on Machine Learning}, pages 7390--7425. PMLR, 2023.

\bibitem[Dhariwal and Nichol(2021)]{dhariwal2021diffusion}
Prafulla Dhariwal and Alexander Nichol.
\newblock Diffusion models beat gans on image synthesis.
\newblock \emph{Advances in neural information processing systems}, 34:\penalty0 8780--8794, 2021.

\bibitem[Dieleman et~al.(2022)Dieleman, Sartran, Roshannai, Savinov, Ganin, Richemond, Doucet, Strudel, Dyer, Durkan, et~al.]{dieleman2022continuous}
Sander Dieleman, Laurent Sartran, Arman Roshannai, Nikolay Savinov, Yaroslav Ganin, Pierre~H Richemond, Arnaud Doucet, Robin Strudel, Chris Dyer, Conor Durkan, et~al.
\newblock Continuous diffusion for categorical data.
\newblock \emph{arXiv preprint arXiv:2211.15089}, 2022.

\bibitem[Du et~al.(2023)Du, Durkan, Strudel, Tenenbaum, Dieleman, Fergus, Sohl-Dickstein, Doucet, and Grathwohl]{du2023reduce}
Yilun Du, Conor Durkan, Robin Strudel, Joshua~B Tenenbaum, Sander Dieleman, Rob Fergus, Jascha Sohl-Dickstein, Arnaud Doucet, and Will~Sussman Grathwohl.
\newblock Reduce, reuse, recycle: Compositional generation with energy-based diffusion models and mcmc.
\newblock In \emph{International conference on machine learning}, pages 8489--8510. PMLR, 2023.

\bibitem[Hang et~al.(2023)Hang, Gu, Li, Bao, Chen, Hu, Geng, and Guo]{hang2023efficient}
Tiankai Hang, Shuyang Gu, Chen Li, Jianmin Bao, Dong Chen, Han Hu, Xin Geng, and Baining Guo.
\newblock Efficient diffusion training via min-snr weighting strategy.
\newblock \emph{arXiv preprint arXiv:2303.09556}, 2023.

\bibitem[Heusel et~al.(2017)Heusel, Ramsauer, Unterthiner, Nessler, and Hochreiter]{heusel2017gans}
Martin Heusel, Hubert Ramsauer, Thomas Unterthiner, Bernhard Nessler, and Sepp Hochreiter.
\newblock Gans trained by a two time-scale update rule converge to a local nash equilibrium.
\newblock \emph{Advances in neural information processing systems}, 30, 2017.

\bibitem[Ho et~al.(2020)Ho, Jain, and Abbeel]{ho2020denoising}
Jonathan Ho, Ajay Jain, and Pieter Abbeel.
\newblock Denoising diffusion probabilistic models.
\newblock \emph{Advances in neural information processing systems}, 33:\penalty0 6840--6851, 2020.

\bibitem[Ho et~al.(2022)Ho, Saharia, Chan, Fleet, Norouzi, and Salimans]{ho2022cascaded}
Jonathan Ho, Chitwan Saharia, William Chan, David~J Fleet, Mohammad Norouzi, and Tim Salimans.
\newblock Cascaded diffusion models for high fidelity image generation.
\newblock \emph{The Journal of Machine Learning Research}, 23\penalty0 (1):\penalty0 2249--2281, 2022.

\bibitem[Hoffman and Johnson(2016)]{hoffman2016elbo}
Matthew~D Hoffman and Matthew~J Johnson.
\newblock Elbo surgery: yet another way to carve up the variational evidence lower bound.
\newblock In \emph{Workshop in Advances in Approximate Bayesian Inference, NIPS}, volume~1, 2016.

\bibitem[Hoffman et~al.(2013)Hoffman, Blei, Wang, and Paisley]{hoffman2013stochastic}
Matthew~D Hoffman, David~M Blei, Chong Wang, and John Paisley.
\newblock Stochastic variational inference.
\newblock \emph{Journal of Machine Learning Research}, 2013.

\bibitem[Hoogeboom et~al.(2022)Hoogeboom, Satorras, Vignac, and Welling]{hoogeboom2022equivariant}
Emiel Hoogeboom, V{\i}ctor~Garcia Satorras, Cl{\'e}ment Vignac, and Max Welling.
\newblock Equivariant diffusion for molecule generation in 3d.
\newblock In \emph{International conference on machine learning}, pages 8867--8887. PMLR, 2022.

\bibitem[Hoogeboom et~al.(2023)Hoogeboom, Heek, and Salimans]{pmlr-v202-hoogeboom23a}
Emiel Hoogeboom, Jonathan Heek, and Tim Salimans.
\newblock simple diffusion: End-to-end diffusion for high resolution images.
\newblock In \emph{Proceedings of the 40th International Conference on Machine Learning}, volume 202 of \emph{Proceedings of Machine Learning Research}, pages 13213--13232. PMLR, 2023.

\bibitem[Hoogeboom et~al.(2024)Hoogeboom, Mensink, Heek, Lamerigts, Gao, and Salimans]{hoogeboom2024simpler}
Emiel Hoogeboom, Thomas Mensink, Jonathan Heek, Kay Lamerigts, Ruiqi Gao, and Tim Salimans.
\newblock Simpler diffusion (sid2): 1.5 fid on imagenet512 with pixel-space diffusion.
\newblock \emph{arXiv preprint arXiv:2410.19324}, 2024.

\bibitem[Huang et~al.(2021)Huang, Lim, and Courville]{huang2021variational}
Chin-Wei Huang, Jae~Hyun Lim, and Aaron~C Courville.
\newblock A variational perspective on diffusion-based generative models and score matching.
\newblock \emph{Advances in Neural Information Processing Systems}, 34:\penalty0 22863--22876, 2021.

\bibitem[Hyv{\"a}rinen and Dayan(2005)]{hyvarinen2005estimation}
Aapo Hyv{\"a}rinen and Peter Dayan.
\newblock Estimation of non-normalized statistical models by score matching.
\newblock \emph{Journal of Machine Learning Research}, 6\penalty0 (4), 2005.

\bibitem[Hyv{\"a}rinen et~al.(2024)Hyv{\"a}rinen, Khemakhem, and Monti]{hyvarinen2024identifiability}
Aapo Hyv{\"a}rinen, Ilyes Khemakhem, and Ricardo Monti.
\newblock Identifiability of latent-variable and structural-equation models: from linear to nonlinear.
\newblock \emph{Annals of the Institute of Statistical Mathematics}, 76\penalty0 (1):\penalty0 1--33, 2024.

\bibitem[Jordan et~al.(1999)Jordan, Ghahramani, Jaakkola, and Saul]{jordan1999introduction}
Michael~I Jordan, Zoubin Ghahramani, Tommi~S Jaakkola, and Lawrence~K Saul.
\newblock An introduction to variational methods for graphical models.
\newblock \emph{Machine learning}, 37:\penalty0 183--233, 1999.

\bibitem[Jun et~al.(2020)Jun, Child, Chen, Schulman, Ramesh, Radford, and Sutskever]{jun2020distribution}
Heewoo Jun, Rewon Child, Mark Chen, John Schulman, Aditya Ramesh, Alec Radford, and Ilya Sutskever.
\newblock Distribution augmentation for generative modeling.
\newblock In \emph{International Conference on Machine Learning}, pages 5006--5019. PMLR, 2020.

\bibitem[Karras et~al.(2022)Karras, Aittala, Aila, and Laine]{karras2022elucidating}
Tero Karras, Miika Aittala, Timo Aila, and Samuli Laine.
\newblock Elucidating the design space of diffusion-based generative models.
\newblock \emph{Advances in Neural Information Processing Systems}, 35:\penalty0 26565--26577, 2022.

\bibitem[Karras et~al.(2024)Karras, Aittala, Lehtinen, Hellsten, Aila, and Laine]{karras2024analyzing}
Tero Karras, Miika Aittala, Jaakko Lehtinen, Janne Hellsten, Timo Aila, and Samuli Laine.
\newblock Analyzing and improving the training dynamics of diffusion models.
\newblock In \emph{Proceedings of the IEEE/CVF Conference on Computer Vision and Pattern Recognition}, pages 24174--24184, 2024.

\bibitem[Kingma et~al.(2021)Kingma, Salimans, Poole, and Ho]{kingma2021variational}
Diederik Kingma, Tim Salimans, Ben Poole, and Jonathan Ho.
\newblock Variational diffusion models.
\newblock \emph{Advances in neural information processing systems}, 34:\penalty0 21696--21707, 2021.

\bibitem[Kingma and Gao(2023)]{kingma2023understanding}
Diederik~P Kingma and Ruiqi Gao.
\newblock Understanding the diffusion objective as a weighted integral of elbos.
\newblock \emph{arXiv preprint arXiv:2303.00848}, 2023.

\bibitem[Kingma and Welling(2013)]{kingma2013auto}
Diederik~P Kingma and Max Welling.
\newblock Auto-encoding variational bayes.
\newblock \emph{arXiv preprint arXiv:1312.6114}, 2013.

\bibitem[Kingma et~al.(2019)Kingma, Welling, et~al.]{kingma2019introduction}
Diederik~P Kingma, Max Welling, et~al.
\newblock An introduction to variational autoencoders.
\newblock \emph{Foundations and Trends{\textregistered} in Machine Learning}, 12\penalty0 (4):\penalty0 307--392, 2019.

\bibitem[Kingma et~al.(2016)Kingma, Salimans, Jozefowicz, Chen, Sutskever, and Welling]{kingma2016improved}
Durk~P Kingma, Tim Salimans, Rafal Jozefowicz, Xi~Chen, Ilya Sutskever, and Max Welling.
\newblock Improved variational inference with inverse autoregressive flow.
\newblock \emph{Advances in neural information processing systems}, 29, 2016.

\bibitem[Kreis et~al.(2022)Kreis, Gao, and Vahdat]{kreis2022tutorial}
Karsten Kreis, Ruiqi Gao, and Arash Vahdat.
\newblock Tutorial on denoising diffusion-based generative modeling: Foundations and applications, 2022.
\newblock Tutorial presented at CVPR 2022.

\bibitem[Lipman et~al.(2023)Lipman, Chen, Ben-Hamu, Nickel, and Le]{lipman2023flow}
Yaron Lipman, Ricky T.~Q. Chen, Heli Ben-Hamu, Maximilian Nickel, and Matthew Le.
\newblock Flow matching for generative modeling.
\newblock In \emph{The Eleventh International Conference on Learning Representations}, 2023.

\bibitem[Liu et~al.(2023)Liu, Gong, and qiang liu]{liu2023flow}
Xingchao Liu, Chengyue Gong, and qiang liu.
\newblock Flow straight and fast: Learning to generate and transfer data with rectified flow.
\newblock In \emph{The Eleventh International Conference on Learning Representations}, 2023.

\bibitem[Luo(2022)]{luo2022understanding}
Calvin Luo.
\newblock Understanding diffusion models: A unified perspective.
\newblock \emph{arXiv preprint arXiv:2208.11970}, 2022.

\bibitem[Maal{\o}e et~al.(2016)Maal{\o}e, S{\o}nderby, S{\o}nderby, and Winther]{maaloe2016auxiliary}
Lars Maal{\o}e, Casper~Kaae S{\o}nderby, S{\o}ren~Kaae S{\o}nderby, and Ole Winther.
\newblock Auxiliary deep generative models.
\newblock In \emph{International conference on machine learning}, pages 1445--1453. PMLR, 2016.

\bibitem[Maal{\o}e et~al.(2019)Maal{\o}e, Fraccaro, Li{\'e}vin, and Winther]{maaloe2019biva}
Lars Maal{\o}e, Marco Fraccaro, Valentin Li{\'e}vin, and Ole Winther.
\newblock Biva: A very deep hierarchy of latent variables for generative modeling.
\newblock \emph{Advances in neural information processing systems}, 32, 2019.

\bibitem[Makhzani et~al.(2015)Makhzani, Shlens, Jaitly, Goodfellow, and Frey]{makhzani2015adversarial}
Alireza Makhzani, Jonathon Shlens, Navdeep Jaitly, Ian Goodfellow, and Brendan Frey.
\newblock Adversarial autoencoders.
\newblock \emph{arXiv preprint arXiv:1511.05644}, 2015.

\bibitem[Meng et~al.(2020)Meng, Song, Song, Zhao, and Ermon]{meng2020improved}
Chenlin Meng, Jiaming Song, Yang Song, Shengjia Zhao, and Stefano Ermon.
\newblock Improved autoregressive modeling with distribution smoothing.
\newblock In \emph{International Conference on Learning Representations}, 2020.

\bibitem[Monteiro et~al.(2022)Monteiro, Ribeiro, Pawlowski, Castro, and Glocker]{monteiro2022measuring}
Miguel Monteiro, Fabio De~Sousa Ribeiro, Nick Pawlowski, Daniel~C Castro, and Ben Glocker.
\newblock Measuring axiomatic soundness of counterfactual image models.
\newblock In \emph{The Eleventh International Conference on Learning Representations}, 2022.

\bibitem[Mueller et~al.(2023)Mueller, Gruber, and Fok]{mueller2023continuous}
Markus Mueller, Kathrin Gruber, and Dennis Fok.
\newblock Continuous diffusion for mixed-type tabular data.
\newblock \emph{arXiv preprint arXiv:2312.10431}, 2023.

\bibitem[Nichol and Dhariwal(2021)]{nichol2021improved}
Alexander~Quinn Nichol and Prafulla Dhariwal.
\newblock Improved denoising diffusion probabilistic models.
\newblock In \emph{International Conference on Machine Learning}, pages 8162--8171. PMLR, 2021.

\bibitem[Nichol et~al.(2022)Nichol, Dhariwal, Ramesh, Shyam, Mishkin, Mcgrew, Sutskever, and Chen]{nichol2022glide}
Alexander~Quinn Nichol, Prafulla Dhariwal, Aditya Ramesh, Pranav Shyam, Pamela Mishkin, Bob Mcgrew, Ilya Sutskever, and Mark Chen.
\newblock Glide: Towards photorealistic image generation and editing with text-guided diffusion models.
\newblock In \emph{International Conference on Machine Learning}, pages 16784--16804. PMLR, 2022.

\bibitem[Ranganath et~al.(2016)Ranganath, Tran, and Blei]{ranganath2016hierarchical}
Rajesh Ranganath, Dustin Tran, and David Blei.
\newblock Hierarchical variational models.
\newblock In \emph{International conference on machine learning}, pages 324--333. PMLR, 2016.

\bibitem[Rasmus et~al.(2015)Rasmus, Berglund, Honkala, Valpola, and Raiko]{rasmus2015semi}
Antti Rasmus, Mathias Berglund, Mikko Honkala, Harri Valpola, and Tapani Raiko.
\newblock Semi-supervised learning with ladder networks.
\newblock \emph{Advances in neural information processing systems}, 28, 2015.

\bibitem[Rezende(2018)]{rezendeblog}
Danilo~J. Rezende.
\newblock Short notes on divergence measures.
\newblock \url{https://danilorezende.com/wp-content/uploads/2018/07/divergences.pdf}, 2018.
\newblock Accessed: 03-09-2024.

\bibitem[Rezende and Viola(2018)]{rezende2018taming}
Danilo~Jimenez Rezende and Fabio Viola.
\newblock Taming vaes.
\newblock \emph{arXiv preprint arXiv:1810.00597}, 2018.

\bibitem[Rezende et~al.(2014)Rezende, Mohamed, and Wierstra]{rezende2014stochastic}
Danilo~Jimenez Rezende, Shakir Mohamed, and Daan Wierstra.
\newblock Stochastic backpropagation and approximate inference in deep generative models.
\newblock In \emph{International conference on machine learning}, pages 1278--1286. PMLR, 2014.

\bibitem[Rissanen et~al.(2023)Rissanen, Heinonen, and Solin]{rissanen2023generative}
Severi Rissanen, Markus Heinonen, and Arno Solin.
\newblock Generative modelling with inverse heat dissipation.
\newblock In \emph{The Eleventh International Conference on Learning Representations}, 2023.

\bibitem[Rombach et~al.(2022)Rombach, Blattmann, Lorenz, Esser, and Ommer]{rombach2022high}
Robin Rombach, Andreas Blattmann, Dominik Lorenz, Patrick Esser, and Bj{\"o}rn Ommer.
\newblock High-resolution image synthesis with latent diffusion models.
\newblock In \emph{Proceedings of the IEEE/CVF conference on computer vision and pattern recognition}, pages 10684--10695, 2022.

\bibitem[Sabour et~al.(2024)Sabour, Fidler, and Kreis]{sabour2024align}
Amirmojtaba Sabour, Sanja Fidler, and Karsten Kreis.
\newblock Align your steps: Optimizing sampling schedules in diffusion models.
\newblock \emph{arXiv preprint arXiv:2404.14507}, 2024.

\bibitem[Saharia et~al.(2022)Saharia, Chan, Saxena, Li, Whang, Denton, Ghasemipour, Gontijo~Lopes, Karagol~Ayan, Salimans, et~al.]{saharia2022photorealistic}
Chitwan Saharia, William Chan, Saurabh Saxena, Lala Li, Jay Whang, Emily~L Denton, Kamyar Ghasemipour, Raphael Gontijo~Lopes, Burcu Karagol~Ayan, Tim Salimans, et~al.
\newblock Photorealistic text-to-image diffusion models with deep language understanding.
\newblock \emph{Advances in Neural Information Processing Systems}, 35:\penalty0 36479--36494, 2022.

\bibitem[Salimans and Ho(2021)]{salimans2021should}
Tim Salimans and Jonathan Ho.
\newblock Should {EBM}s model the energy or the score?
\newblock In \emph{Energy Based Models Workshop - ICLR 2021}, 2021.

\bibitem[Salimans and Ho(2022)]{salimans2022progressive}
Tim Salimans and Jonathan Ho.
\newblock Progressive distillation for fast sampling of diffusion models.
\newblock In \emph{International Conference on Learning Representations}, 2022.

\bibitem[Salimans et~al.(2015)Salimans, Kingma, and Welling]{salimans2015markov}
Tim Salimans, Diederik Kingma, and Max Welling.
\newblock Markov chain monte carlo and variational inference: Bridging the gap.
\newblock In \emph{International conference on machine learning}, pages 1218--1226. PMLR, 2015.

\bibitem[Salimans et~al.(2016)Salimans, Goodfellow, Zaremba, Cheung, Radford, and Chen]{salimans2016improved}
Tim Salimans, Ian Goodfellow, Wojciech Zaremba, Vicki Cheung, Alec Radford, and Xi~Chen.
\newblock Improved techniques for training gans.
\newblock \emph{Advances in neural information processing systems}, 29, 2016.

\bibitem[Santos and Lin(2023)]{santos2023using}
Javier~E Santos and Yen~Ting Lin.
\newblock Using ornstein-uhlenbeck process to understand denoising diffusion probabilistic model and its noise schedules.
\newblock \emph{arXiv preprint arXiv:2311.17673}, 2023.

\bibitem[Shu and Ermon(2022)]{shu2022bit}
Rui Shu and Stefano Ermon.
\newblock Bit prioritization in variational autoencoders via progressive coding.
\newblock In \emph{International Conference on Machine Learning}, pages 20141--20155. PMLR, 2022.

\bibitem[Sohl-Dickstein et~al.(2015)Sohl-Dickstein, Weiss, Maheswaranathan, and Ganguli]{sohl2015deep}
Jascha Sohl-Dickstein, Eric Weiss, Niru Maheswaranathan, and Surya Ganguli.
\newblock Deep unsupervised learning using nonequilibrium thermodynamics.
\newblock In \emph{International conference on machine learning}, pages 2256--2265. PMLR, 2015.

\bibitem[S{\o}nderby et~al.(2016)S{\o}nderby, Raiko, Maal{\o}e, S{\o}nderby, and Winther]{sonderby2016ladder}
Casper~Kaae S{\o}nderby, Tapani Raiko, Lars Maal{\o}e, S{\o}ren~Kaae S{\o}nderby, and Ole Winther.
\newblock Ladder variational autoencoders.
\newblock \emph{Advances in neural information processing systems}, 29, 2016.

\bibitem[Song and Ermon(2019)]{song2019generative}
Yang Song and Stefano Ermon.
\newblock Generative modeling by estimating gradients of the data distribution.
\newblock \emph{Advances in neural information processing systems}, 32, 2019.

\bibitem[Song and Ermon(2020)]{song2020improved}
Yang Song and Stefano Ermon.
\newblock Improved techniques for training score-based generative models.
\newblock \emph{Advances in neural information processing systems}, 33:\penalty0 12438--12448, 2020.

\bibitem[Song and Kingma(2021)]{song2021train}
Yang Song and Diederik~P Kingma.
\newblock How to train your energy-based models.
\newblock \emph{arXiv preprint arXiv:2101.03288}, 2021.

\bibitem[Song et~al.(2021{\natexlab{a}})Song, Durkan, Murray, and Ermon]{song2021maximum}
Yang Song, Conor Durkan, Iain Murray, and Stefano Ermon.
\newblock Maximum likelihood training of score-based diffusion models.
\newblock \emph{Advances in Neural Information Processing Systems}, 34:\penalty0 1415--1428, 2021{\natexlab{a}}.

\bibitem[Song et~al.(2021{\natexlab{b}})Song, Sohl-Dickstein, Kingma, Kumar, Ermon, and Poole]{song2021scorebased}
Yang Song, Jascha Sohl-Dickstein, Diederik~P Kingma, Abhishek Kumar, Stefano Ermon, and Ben Poole.
\newblock Score-based generative modeling through stochastic differential equations.
\newblock In \emph{International Conference on Learning Representations}, 2021{\natexlab{b}}.

\bibitem[Tomczak and Welling(2018)]{tomczak2018vae}
Jakub Tomczak and Max Welling.
\newblock Vae with a vampprior.
\newblock In \emph{International Conference on Artificial Intelligence and Statistics}, pages 1214--1223. PMLR, 2018.

\bibitem[Vahdat and Kautz(2020)]{vahdat2020nvae}
Arash Vahdat and Jan Kautz.
\newblock Nvae: A deep hierarchical variational autoencoder.
\newblock \emph{Advances in Neural Information Processing Systems}, 33:\penalty0 19667--19679, 2020.

\bibitem[Vahdat et~al.(2021)Vahdat, Kreis, and Kautz]{vahdat2021score}
Arash Vahdat, Karsten Kreis, and Jan Kautz.
\newblock Score-based generative modeling in latent space.
\newblock \emph{Advances in Neural Information Processing Systems}, 34:\penalty0 11287--11302, 2021.

\bibitem[Valpola(2015)]{valpola2015neural}
Harri Valpola.
\newblock From neural pca to deep unsupervised learning.
\newblock In \emph{Advances in independent component analysis and learning machines}, pages 143--171. Elsevier, 2015.

\bibitem[Vincent(2011)]{vincent2011connection}
Pascal Vincent.
\newblock A connection between score matching and denoising autoencoders.
\newblock \emph{Neural computation}, 23\penalty0 (7):\penalty0 1661--1674, 2011.

\bibitem[Vincent et~al.(2008)Vincent, Larochelle, Bengio, and Manzagol]{vincent2008extracting}
Pascal Vincent, Hugo Larochelle, Yoshua Bengio, and Pierre-Antoine Manzagol.
\newblock Extracting and composing robust features with denoising autoencoders.
\newblock In \emph{Proceedings of the 25th international conference on Machine learning}, pages 1096--1103, 2008.

\bibitem[Wasserman(2004)]{wasserman2004all}
Larry Wasserman.
\newblock \emph{All of statistics: a concise course in statistical inference}, volume~26.
\newblock Springer, 2004.

\end{thebibliography}

\end{document}